\documentclass[lettersize,journal]{IEEEtran}
\usepackage{amsmath,amsfonts}
\usepackage{algorithmic}
\usepackage{algorithm}
\usepackage{array}
\usepackage{multirow}

\usepackage[caption=false,font=footnotesize,labelfont=sf,textfont=sf]{subfig}
\usepackage{textcomp}
\usepackage{stfloats}
\usepackage{url}
\usepackage{verbatim}
\usepackage{graphicx}
\usepackage{cite}
\DeclareUnicodeCharacter{2212}{-}

\begin{document}

\title{A High Resolution Multi-exposure Stereoscopic Image \& Video Database of Natural Scenes}

\author{Rohit Choudhary $^{1}$,~Mansi~Sharma $^{1}$, and ~Aditya Wadaskar $^{1}$
\thanks{$^1$Department of Electrical Engineering,
Indian Institute of Technology Madras, Tamil Nadu, 600036, India
(e-mail: ee20s002@smail.iitm.ac.in, mansisharmaiitd@gmail.com, mansisharma@ee.iitm.ac.in, ay.wadaskar@gmail.com)}
}

\maketitle

\begin{abstract}
Immersive displays such as VR headsets, AR glasses, Multiview displays, Free point televisions have emerged as a new class of display technologies in recent years, offering a better visual experience and viewer engagement as compared to conventional displays. With the evolution of 3D video and display technologies, the consumer market for High Dynamic Range (HDR) cameras and displays is quickly growing. The lack of appropriate experimental data is a critical hindrance for the development of primary research efforts in the field of 3D HDR video technology. Also, the unavailability of sufficient real-world multi-exposure experimental dataset is a major bottleneck for HDR imaging research, thereby limiting the quality of experience (QoE) for the viewers. In this paper, we introduce a diversified stereoscopic multi-exposure dataset captured within the campus of Indian Institute of Technology Madras, which is home to a diverse flora and fauna. The dataset is captured using ZED stereoscopic camera and provides intricate scenes of outdoor locations such as gardens, roadside views, festival venues, buildings and indoor locations such as academic and residential areas. The proposed dataset accommodates wide depth range, complex depth structure, complicate object movement, illumination variations, rich color dynamics, texture discrepancy in addition to significant randomness introduced by moving camera and background motion. The proposed dataset is made publicly available to the research community. Furthermore, the procedure for capturing, aligning and calibrating multi-exposure stereo videos and images is described in detail. Finally, we have discussed the progress, challenges, potential use cases and future research opportunities with respect to HDR imaging, depth estimation, consistent tone mapping and 3D HDR coding.
\end{abstract}


\section{Introduction}
\label{sec:intro}

The vast research in immersive visual media aims to provide the viewer's realistic experience. Immersive technology creates a sense of immersion by blurring the lines between the actual, virtual, and simulated worlds \cite{Lee2013PresenceIV}. The ISO/IEC, JCT-3V, ITU-T, MEPG/VCEG industry and research groups are focusing their efforts on building markets for immersive HDR/WCG video, UltraHD, light field displays, 3D TVs, omnidirectional 360$^{\circ}$ video, Free viewpoint television (FTV) and Mobile 3D apps \cite{Assuno20183DBOOK}. Extracting 3D depth information, 3D content development and optimization, quality evaluations of virtual objects, and digital hologram patterns are all examples of immersive media approaches. With the release of the AVC (Advanced Video Coding), HEVC (High Efficiency Video Coding) and following advances in coding systems such as MVC (Multiview Video Coding), 3D-HEVC, MV-HEVC, JPEG Pleno and NGVC (Next-Generation Video Coding technology), immersive media technologies are attaining momentum. As several of these technologies are still in their early stages, the lack of publicly available 3D HDR datasets is a significant roadblock to further research and development. 

With the rise in demand for realistic and immersive multimedia, high dynamic range (HDR) imaging has recently received much attention from academia and industry. The ultimate aim is to provide the end-users a close to natural quality of experience (QoE). Typically natural scenes have a very high range of luminance. The dynamic range of an image is determined as the ratio of the maximum luminance to the minimum luminance. In real scenes, the dynamic range of irradiance may exceed up to 100,000,000:1. Depending upon certain circumstances, the human eye can detect dynamic ranges varying from 10,000:1 to 1,000,000:1, whereas a conventional display can exhibit a low dynamic range of 100:1 to 300:1 \cite{LE2020102971}.

Low dynamic range (LDR) images represent luminance in 256 steps, \textit{i.e.}, from 0 to 255 using 8-bits. As a result, pixel values outside this range are replaced with 0 or 255, resulting in loss of details. HDR imaging significantly reduces this contrast loss by allocating additional bits to indicate luminance, hence expanding the luminance dynamic range close to Human Visual System (HVS) perceived gamut. Many algorithms are there to construct HDR images from multiple LDR images. With the help of floating-point capabilities on graphics cards, the visual industries are fast transitioning to an HDR rendering pipeline. Traditional display devices are not designed to provide the required luminance range for a full HDR experience. Also, they fail to display floating-point data \cite{DoHDRDisplaysSupportLDR2007}. Consequently, to prepare HDR images for display on regular screens, dynamic range reduction (\textit{i.e.}, tone mapping) is done. During tone mapping most of the color and brightness information is preserved. Additionally, fusion of SDR images at varying exposure levels gives an output SDR image that contains more details than input images. Unlike tone mapping, this approach does not need the creation of an HDR irradiance map at all. 

\begin{figure}[t!]
\centering
  \includegraphics[height=35mm,width=75mm]{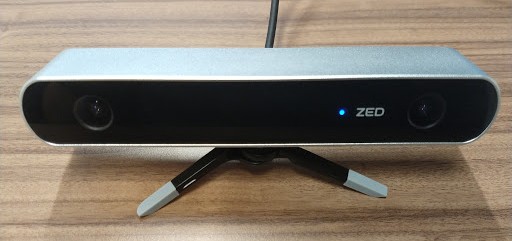}
  \caption{The ZED Camera is used for scene acquisition.}
  \label{fig:zed}
\end{figure}

Despite a surge in scientific interest for HDR imaging, its industrial use is still limited, majorly due to the lack of standard HDR image/video coding schemes \cite{Korshunov2015HDRJPEG}. The lack of publicly available HDR databases hinder the research growth in HDR computational photography, HDR video compression, HDR quality assessment, HDR image coding, and HDR video coding.

While HDR technology adds a more realistic gamut of light and colour, 3D technologies adds more realism to the viewing experience by introducing the depth. 3D reconstruction is essential in applications like smart robotics, virtual reality, augmented reality and autonomous driving. The process of creating 3D videos is usually divided into three categories. First, 3D content can be collected using two or more synchronised cameras. Second, 3D video can be produced from 2D videos using post-production video processing techniques \cite{sharma2017unclaibrated}. Third, depth information obtained by a sensor can be used to supplement the video output. Stereo vision technology utilize two different cameras separated by a baseline distance, to capture images of the same scenes from two different viewpoints which enables the viewer to perceive scene depth. Standardizing stereo and multi-view video are popular commercialization initiatives in 3D technology. In general, a standard video compression method is critical for transmitting and storing stereoscopic and multi-view data \cite{Vetro2011MVC}.

The latest commercialization step in 3D technology includes the usage of glass-free automultiscopic displays, which delivers a significantly more realistic 3D experience to numerous viewers. It is achieved by converting stereoscopic 3D content into high quality multi-view 3D content using stereo to multi-view video conversion methods \cite{Kellnhofer20173DTV, Sharma2014Uncalibrated3DTV}. However, as compared to the HDR immersive video content, 3D databases (mostly stereo image and video) are still available \cite{Song20142DplusDepth, Goldmann2010StereoVideoDataset, Goldmann_impactof, Moorthy2013SubjectiveEO, CHEN20131143}. Multi-view RGB video plus depth datasets such as \cite{Lai2011MVDdataset, Yang2021MVDRobotics} are also available. As a result, there is a lot of research going on in the fields of 3D video compression \cite{Muller20133DVideo, Sullivan2013StdExtensionHEVC}, 3D visual attention modelling \cite{ChagnonForget2016EnhancedVM, BanitalebiDehkordi2018Benchmark3E} and quality assessment \cite{Hewage2009QualityEvaluation, shao2013QA}. 

\begin{table}[t!]
\centering
\caption{Relevant features of the ZED Camera}

\begin{tabular}{|c|c|}
\hline
\textbf{Dimensions} & $175 \times 30 \times 33 mm$ \\
\hline

\multirow{4}{*}
 & HD2K: $2208 \times 1242$ (15 FPS) \\ 
\textbf{Video mode \&} & HD1080:$ 1920 \times 1080$ (30, 15 FPS)  \\ 
\textbf{output resolution} & HD720: $1280 \times 720$ (60, 30, 15 FPS)   \\ 
 & WVGA: $672 \times 376$ (100, 60, 30, 15 FPS)  \\ \hline
 
\multirow{3}{*}{\textbf{Depth}} 
 & Depth range: $0.5m$ to $20m$ \\ 
 & Format: $32$ bits  \\ 
 & Baseline: $120 mm$ \\ \hline
 
\multirow{2}{*}
{$\textbf{Lens}$} 
 & Field of View: $110^{\circ}$ \\
 & $f/2.0$ aperture \\ \hline
 
\multirow{3}{*}{\textbf{Sensors}} 
 & Sensor size: $1/3$ \\
 & Pixel Size: $2\mu$ pixels \\
 & Format: $16:9$ \\ \hline
 
\multirow{3}{*}{\textbf{Camera controls}} 
 & Adjust: Contrast, Frame rate, Gamma, \\
 & Brightness, Resolution, Saturation, \\
 & Sharpness, Exposure \& White Balance \\ \hline
 
\multirow{2}{*}{\textbf{Connectivity}} 
 & USB $3.0$ $(5 V/380 mA)$ \\
 & $0^{\circ}C$ to $+45^{\circ}C$ \\ \hline
 
\multirow{4}{*}{} 
 &  Linux or Windows \\
 \textbf{SDK System} &  Dual$-$core $2.3$ GHz \\
 \textbf{requirements} & $4$ GB RAM \\
 & Nvidia GPU \\ \hline

\end{tabular}
\label{tab: FeaturesZED}
\end{table}

With the introduction of High Efficiency Video Coding (HEVC) standard and  subsequent introduction of coding techniques, such as 3D-HEVC and MV-HEVC, advanced immersive media technologies are gaining momentum \cite{Hannuksela2015MVHEVC}. A relatively advanced and less explored technology is 3D HDR video, which combines both 3D and HDR video technologies. To the best of our knowledge, we only know of one stereo HDR video database, which was publicly introduced by Amin Banitalebi Dehkordi \cite{3DSHDR2017}. The majority of the work in generating 3D HDR video revolves around merging various views of the same scene taken at different time instances to create synthetic stereo HDR images. Methods such as \cite{Rufenacht2011stereoscopichigh} generated HDR images using low-dynamic images acquired by two inexpensive LDR cameras for the same scene. Making an advancement to this approach \cite{SELMANOVIC2014216} used one LDR camera and an HDR camera to generate stereo HDR data. Amin Banitalebi-Dehkordi \cite{3DSHDR2017} used two commercially available side-by-side HDR cameras to produce HDR stereo data. 

Encouraged by SHDR video database released by \cite{3DSHDR2017}, we introduce a large-scale public database of multi-exposure stereo images and videos featuring complex natural scenes. The database comprises of diverse flora and fauna, water streams, irregular reflective surfaces, dynamic background giving an impressive composition which is rich in texture, illumination variations, colors and details. It's worth noting that we address the problems associated with HDR reconstruction using multi-exposure images or frames, stereo depth processing for 3D creation and 3D HDR reconstruction. Also relevant research areas for interested readers. As the research area of 3D HDR video becomes more popular, reliable benchmark datasets of natural scenes are required for further study. The proposed dataset is available at \cite{me-exposure-data}.

A shorter conference version to lay the foundation of this dataset is published at IEEE IC3D 2019 \cite{Wadaskar2019IC3D}. In this journal extension, we elaborate on the features of the proposed dataset with an extensive description of the intricate details of the captured scenes. Section \ref{sec:creation_database} discusses the process of creation of the proposed database in detail. It includes specification, configuration and calibration of the ZED camera, followed by the description of database capturing process, format and rectification. Section \ref{sec:database} gives an elaborate explanation of the intricacy and diversity of the database. Section \ref{sec:ChallengesImmersive} gives insight into the progress, use cases, challenges and potential research opportunities in the domain of HDR image/video reconstruction, depth estimation for 3D content generation, tone mapping and dynamic range compression. As a result, extensive analysis of hindrance and future research prospects in the field of 3D-HDR are discussed. Section \ref{sec:Conc} presents the conclusion and discuss the implications of the proposed database for research and development IN 3D HDR technologies.

\begin{table}[t!]
\centering
\caption{ZED Camera calibration parameters}

\begin{tabular}{|c|c|c|} 
\hline
\multicolumn{3}{|c|}{\textbf{L/R Sensor Parameters (2K)}} \\
\hline
\textbf{Parameter} & \textbf{Left} & \textbf{Right}\\ 
\hline
Focal length $f_x$  & $1400.64$ & $1396.97$ \\
\hline
Focal length $f_y$  & $1400.64$ & $1396.97$ \\
\hline
Principal point $c_x$  & $1078.91$ & $1072.43$ \\
\hline
$K_1$  & $-0.17233$ & $-0.17312$ \\
\hline
$K_2$  & $1400.64$ & $1396.97$ \\
\hline
$K_3$  & $0$ & $0$ \\
\hline
$p_1$  & $0.002236491$ & $0.00240052$ \\
\hline
$p_2$  & $-0.000661606$ & $-0.000573855$ \\
\hline
\multicolumn{3}{|c|}{\textbf{Stereo Parameters (2K)}} \\
\hline
\textbf{Parameter} & \multicolumn{2}{|c|}{\textbf{value}} \\
\hline
baseline & \multicolumn{2}{|c|}{$119.996$} \\
\hline
Translational $T_Y$ & \multicolumn{2}{|c|}{$-0.00121791$} \\
\hline
Translational $T_Z$ & \multicolumn{2}{|c|}{$0.00502657$} \\
\hline
Principal $C_V$ & \multicolumn{2}{|c|}{$0.00840017$} \\
\hline
Rotational $R_X$ & \multicolumn{2}{|c|}{$0.000188929$} \\
\hline
Rotational $R_Z$ & \multicolumn{2}{|c|}{$0.000412625$} \\
\hline

\end{tabular}
\label{tableZEDHDR}
\end{table}

\section{Creation of Multi-exposure Stereoscopic Image \& Video Database}
\label{sec:creation_database}
The multi-exposure stereo image \& video database has been captured using ZED stereoscopic camera \cite{StereoLabs}. In the next sections, ZED camera specifications, capturing setup, and the procedure for acquiring stereoscopic multi-exposure images and videos are described.

\subsection{Stereo camera specification \& configuration } 
The ZED camera has synchronized dual sensors mounted on a single frame to capture the left and right views as shown in Figure \ref{fig:zed}. Each sensor has a 4M pixels resolution, with large $2\mu$ sized pixels. The sensors are spaced by $12$ cm horizontally and have a ground height of roughly $3.5-4$ cm. The lenses are wide angle, having a field of vision of 90 degrees horizontally, 60 degrees vertically, and 110 degrees diagonally, with a $f/2.0$ aperture. To capture the scenes, the camera is put on a small tripod and placed on a sturdy surface (ground level or higher). The depth range of the camera is $0.5 m$ to $20 m$. The ZED SDK software support is used to capture simultaneous, frame-aligned left \& right views of images \& videos. Software controls also aid in varying exposure, resolution, frame rate, aspect ratio and brightness settings. Important features of ZED camera are mentioned in Table \ref{tab: FeaturesZED}.

\subsection{Stereo camera calibration}
The intrinsic and extrinsic parameters of the cameras are determined through the calibration procedure. The focal lengths ($f_x$, $f_y$), principal point coordinates ($c_x$, $c_y$), radial ($K_1$, $K_2$, $K_3$), and tangential ($p_1$, $p_2$) distortion factors are among the intrinsic parameters. Intrinsic parameters are often employed to acquire images free of distortions induced by lenses and camera structure, and to obtain three-dimensional representations of a scene. Extrinsic parameters, on the other hand, link real-world reference systems to the camera, specifying the device's location and orientation in the real-world coordinate system (using translation and rotation vector).

In addition to intrinsic and extrinsic calibrations, stereo calibration permits obtaining information that links the coordinates of the two cameras (left and right camera) in space. The ZED camera has been factory calibrated, and external calibration is not needed as such. However, to ensure highly accurate stereo and sensor calibration, we performed re-calibration using the in-built ZED calibration tool included in the SDK. The camera parameters obtained on calibration have been summarised in Table \ref{tableZEDHDR}. The table contains parameters for 2K captures only. 

\subsection{Stereo rectification \& depth estimation }
Stereo rectification is the method of correcting two stereo images of the same scene such that they appear to have been captured by two cameras with row-aligned image planes. Image rectification is obtained through ZED SDK. Rectified images simplify the estimation of stereo disparity which is a fundamental procedure before estimating the corresponding depth map. The depth information is computed by ZED camera using triangulation from the geometric model of non-distorted rectified cameras \cite{ortiz2018depth}.

\subsection{Capturing procedure \& data format specification}
The multi-exposure stereoscopic image database contains $39$ natural scenes. Each scene has been captured under $3-4$ different exposure settings achieved through the SDK controls. All images are acquired in 2K (full HD) resolution. The left and right views, each with a resolution of $2208 \times 1242$, are combined into a single output image of resolution $4416 \times 1242$. The camera is kept fixed between the successive multi-exposure captures. Hence, the database stereo images are frame-level aligned.

Similarly, our database of eighteen 3D HDR videos contain short fixed-frame captures of natural scenes with slight to medium, partially traceable object motion at varying depth ranges. All videos are captured in $1920 \times 1080$ for each view, at a frame rate of $30$ fps. Video datasets has been recorded with  ZED SDK. The SDK stores video clips in the Stereolabs SVO format \cite{StereoLabs}, with extra metadata such as timestamps and sensor data. Each scene has been captured under $3-4$ different exposures sequentially using the software controls, while ensuring perfect frame alignment between successive captures. However, since multiple exposures are captured sequentially, there is slight variation in the motion of object's in the acquired scene. For example, a scene with trees swaying due to the breeze, flowing water, etc. The complexity of captured scenes opens new challenges and opportunities to the researchers.

\section{Stereo data characteristics} 
\label{sec:database}

Most of IIT Madras campus is a protected forest carved out of the Guindy National Park. It is home to about 300 floral species unique to India's tropical dry evergreen regions. The campus has a large numbers of Blackbuck, spotted deer, bonnet macaque, and other rare wildlife species, that contribute significantly to the campus's biodiversity. The proposed stereoscopic database assists in the understanding of campus scenes for several computer vision and 3D tasks such as depth estimation, 2D/3D HDR reconstruction, development of tone mapping algorithms, etc. The database comprises of 39 scenes captured at varying exposures. In addition, 18 video scenes have also been provided. We attempted to capture scenes with a wide depth range, depth structure, camera motion, object motion, illumination and scene complexity. While capturing the scenes, great care was taken to prevent violating the 3D window \cite{Smolic20113DVProcessing}. The 3D window violation occurs when moving objects in the scene are clipped and overlap with the captured image border, causing retinal revelry. During post-processing, any discrepancies in the proposed database are rectified. 

The database captures a wide range of natural outdoor scenes and a few indoor scenes, providing researchers with fresh opportunities and problems to work on. Outdoor scenes include garden, roadside view, campus festival venue, buildings and other architectures. Indoor scenes are captured inside academic and residential areas. The images and videos chosen for this database have a strong depth bracket and offers rich variations of color, texture, and illuminations over the dynamic range. Occlusions and texture-less regions further adds to the complexity of the captured scenes. Our stereoscopic dataset can be categorized based on following characteristics:

{\textbullet \hspace{1 mm} \textbf{Rare Flora and Fauna}: Forest scenes with great details, rich shades of green, rigorous movement of trees in the wind, changing shadow pattern, fascinating interplay of sunlight, altering sunbeams, silhouette, and occlusion due to vegetation are included in the dataset. The data is rich and full of intricate details because of features like silhouetted leaves against the sky, sunlight flashing through foliage, or swinging banyan roots in the wind.}

{\textbullet \hspace{1 mm} \textbf{Sky-scapes}: The dataset also features rich sky-scapes with brightly illuminated sky, vivid cloud patterns and saturated evening sky. Conflict arises in capturing a scene with sky-scapes as details of the sky can be accurately recorded with very short exposures, while the tree shoots/leaves require a very high exposure to capture their details accurately. It is impossible to get features of the sky with long exposures as it makes the sky seem extraordinarily brilliant. As a result of the broad depth bracket and complicated lighting setup, such scenarios are perfect for enhancing dynamic range.}

\begin{figure*}[h!]
{\subfloat[$Banyan1$]{\includegraphics[height=1.1in,width=1.60in]{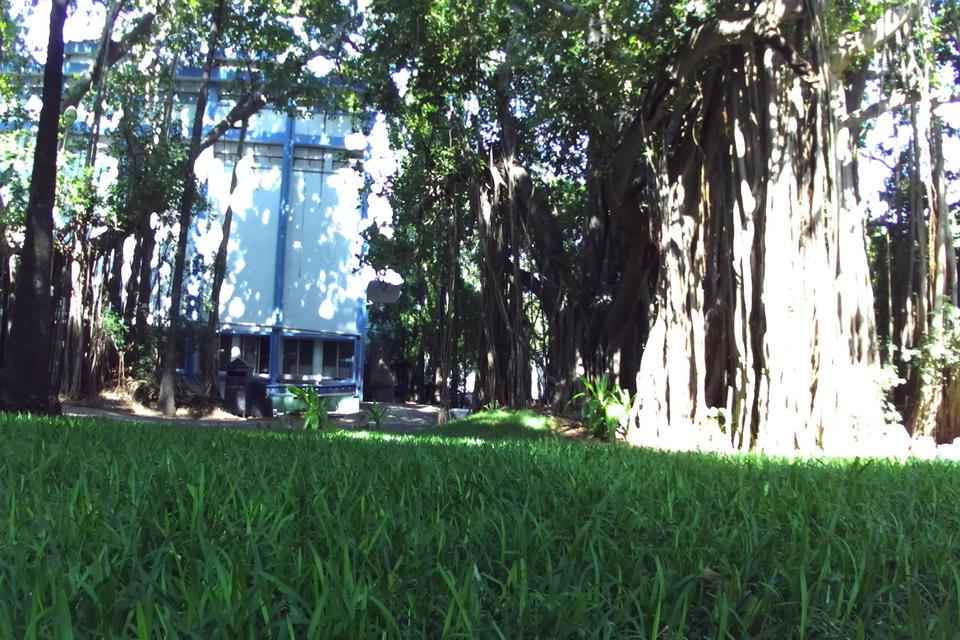}}}\hfill
{\subfloat[$BanyanAvenue$]{\includegraphics[height=1.1in,width=1.60in]{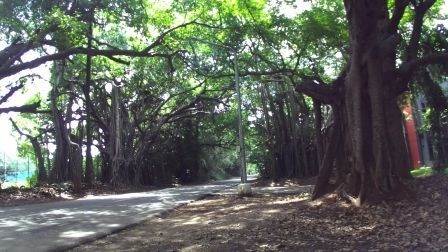}}}\hfill
{\subfloat[$Canopy$]{\includegraphics[height=1.1in,width=1.60in]{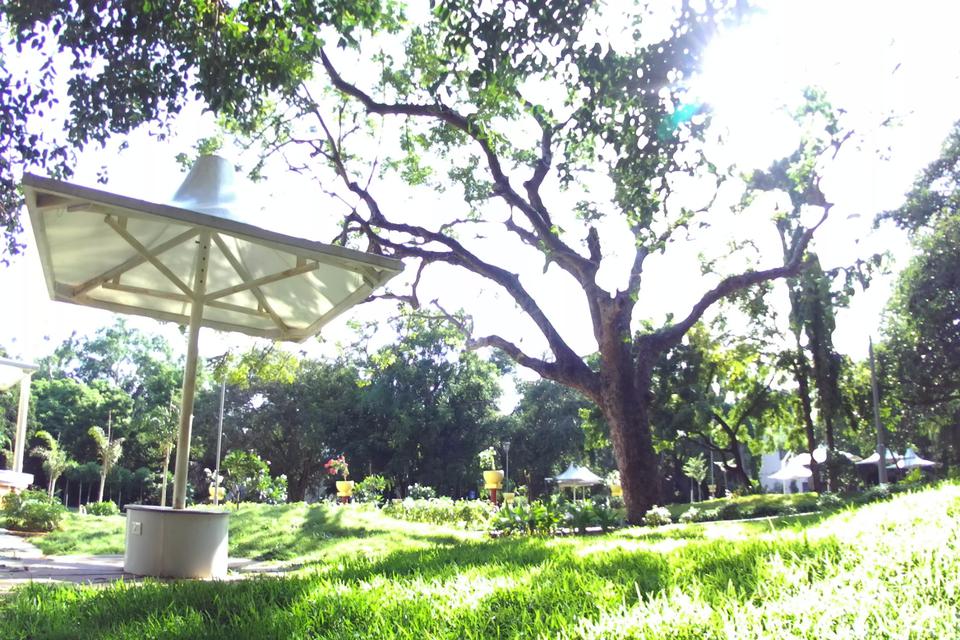}}}\hfill
{\subfloat[$Pergola$]{\includegraphics[height=1.1in,width=1.60in]{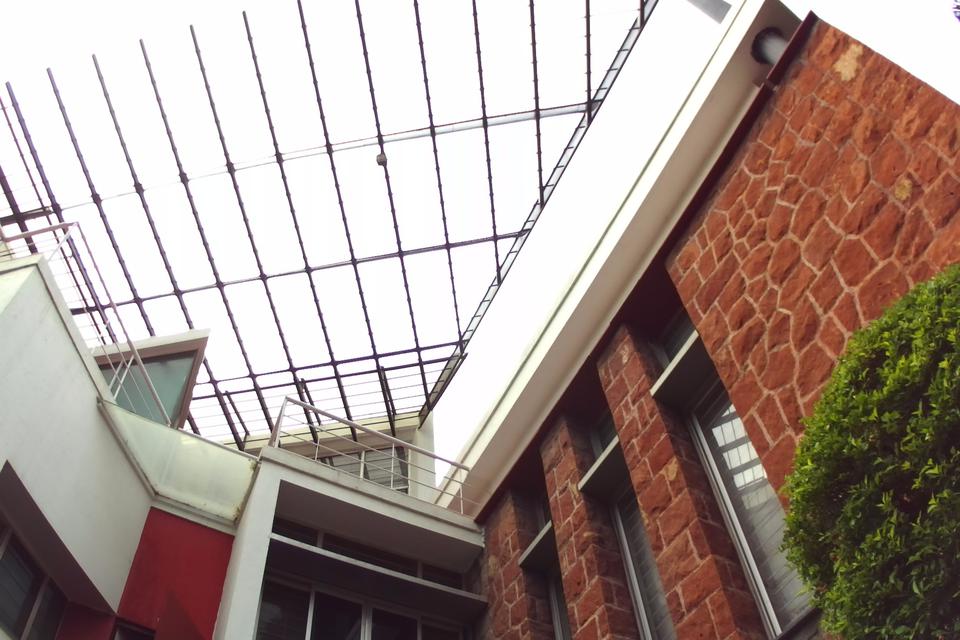}}}\hfill

{\subfloat[$GCLong$]{\includegraphics[height=1.1in,width=1.60in]{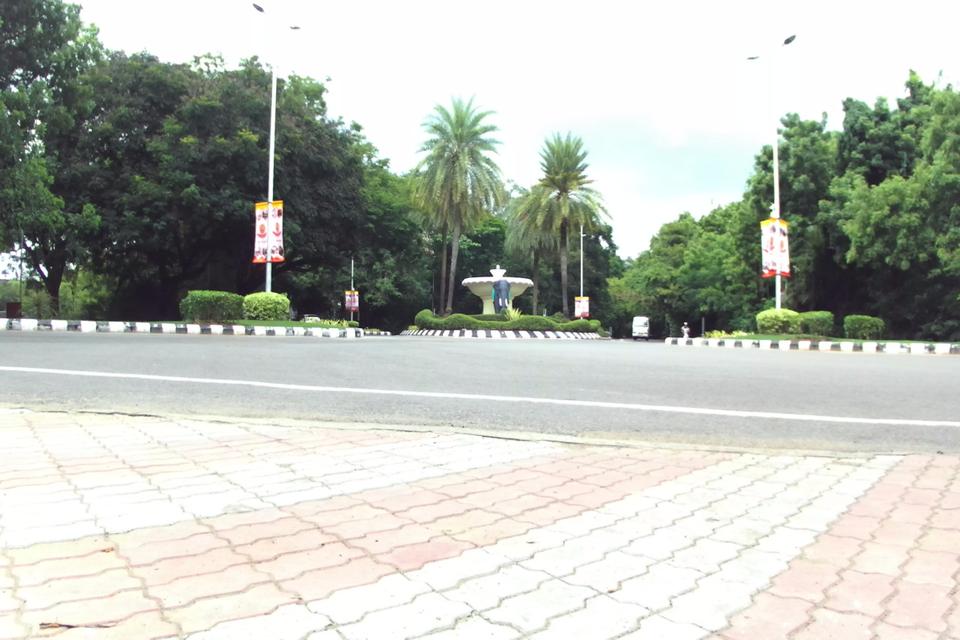}}}\hfill
{\subfloat[$GlassBuilding$]{\includegraphics[height=1.1in,width=1.60in]{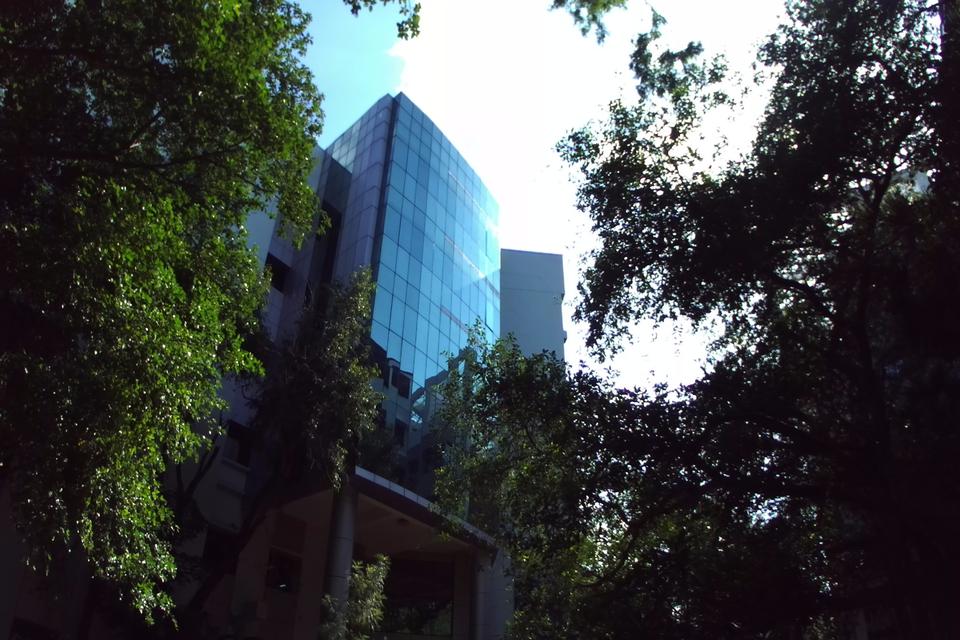}}}\hfill
{\subfloat[$Hall$]{\includegraphics[height=1.1in,width=1.60in]{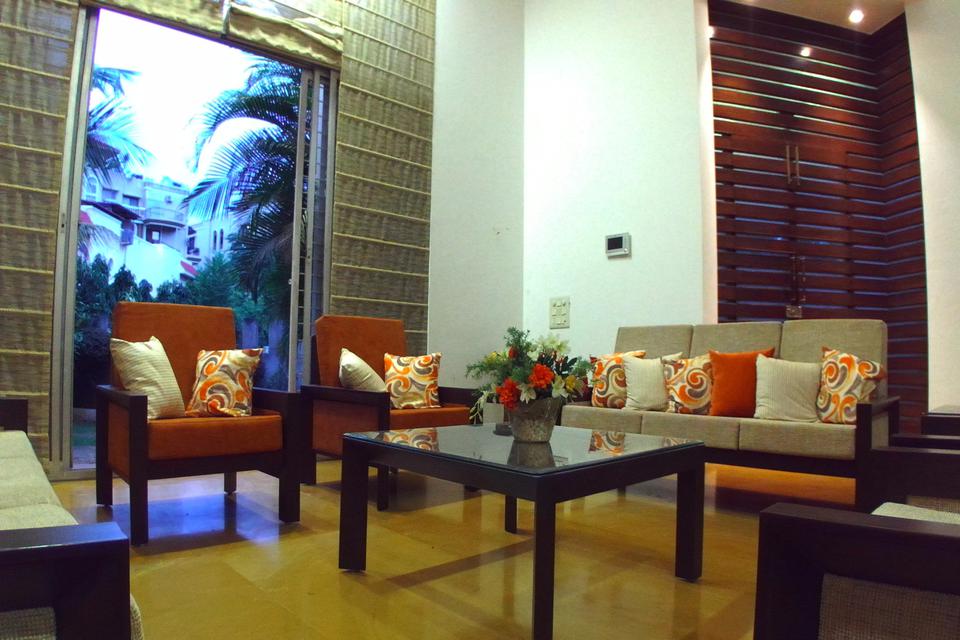}}}\hfill
{\subfloat[$HimalayaBuilding$]{\includegraphics[height=1.1in,width=1.60in]{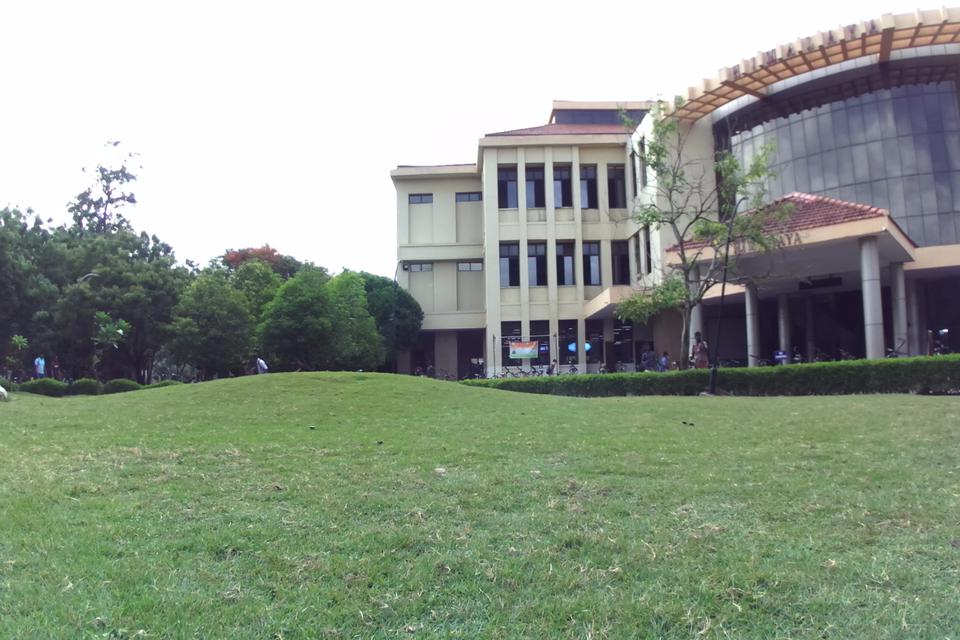}}}\hfill

{\subfloat[$House1$]{\includegraphics[height=1.1in,width=1.60in]{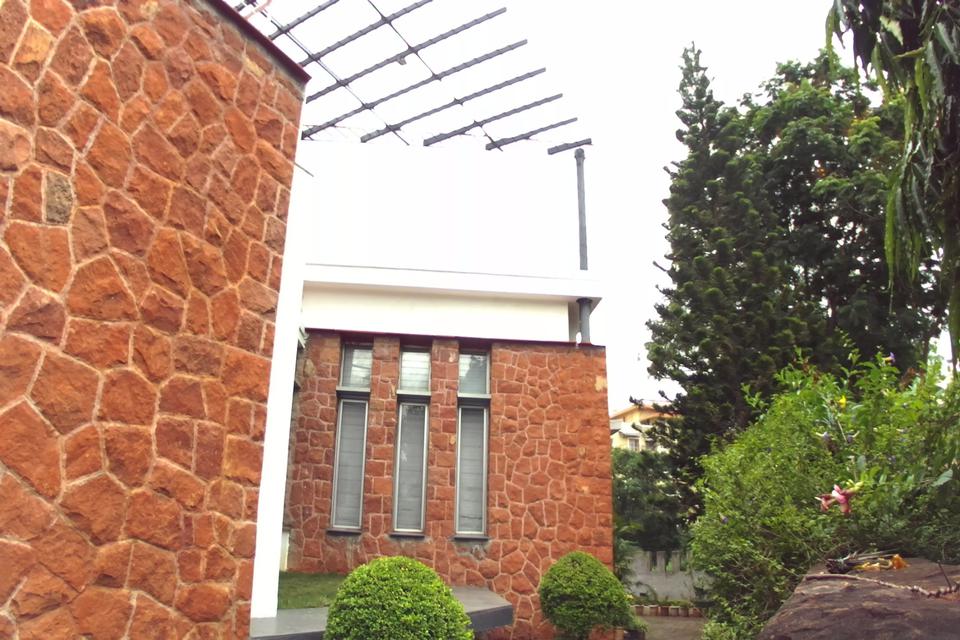}}}\hfill
{\subfloat[$House2$]{\includegraphics[height=1.1in,width=1.60in]{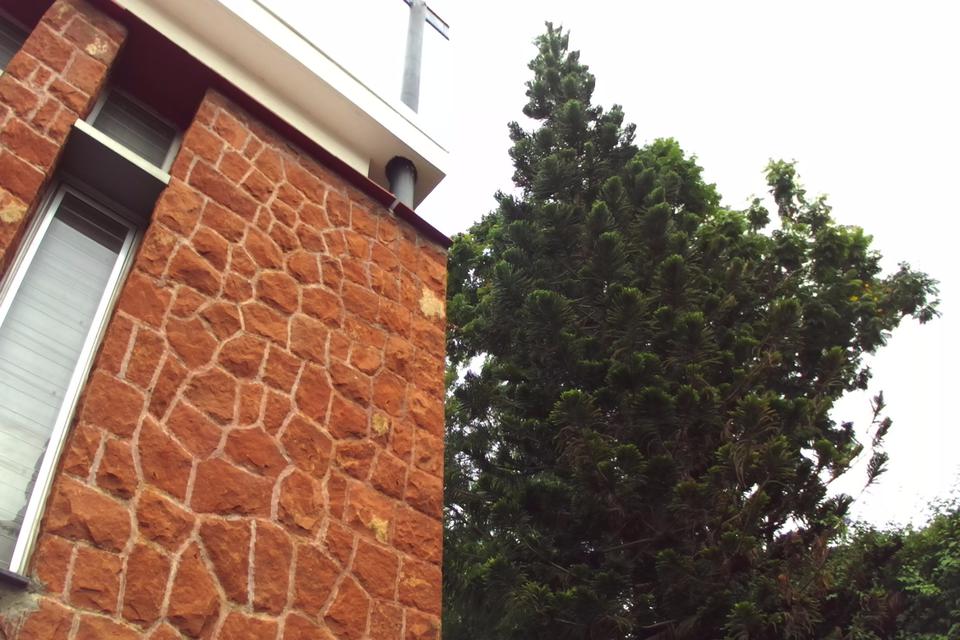}}}\hfill
{\subfloat[$Marsh$]{\includegraphics[height=1.1in,width=1.60in]{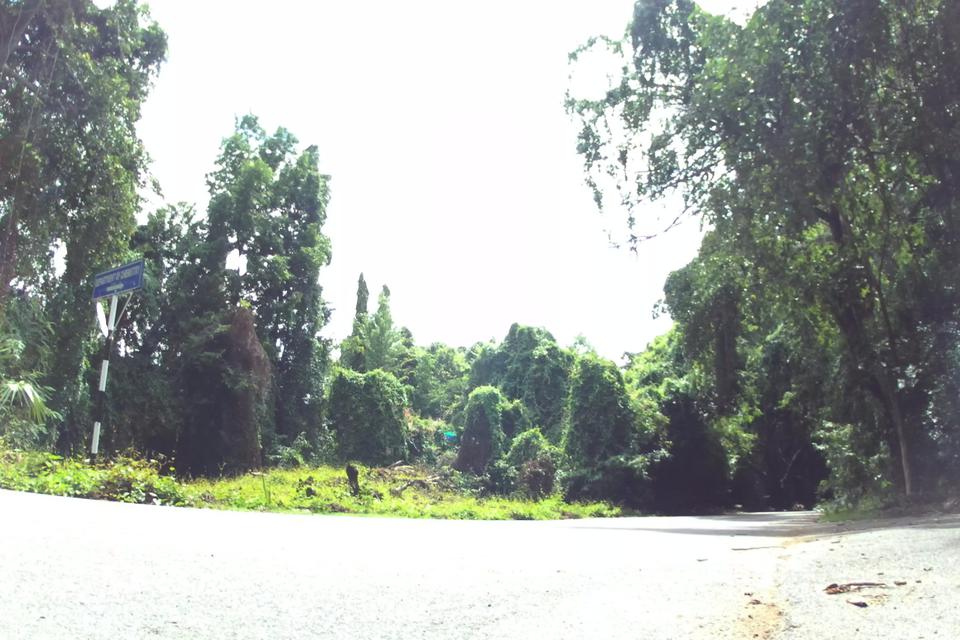}}}\hfill
{\subfloat[$NACBuilding$]{\includegraphics[height=1.1in,width=1.60in]{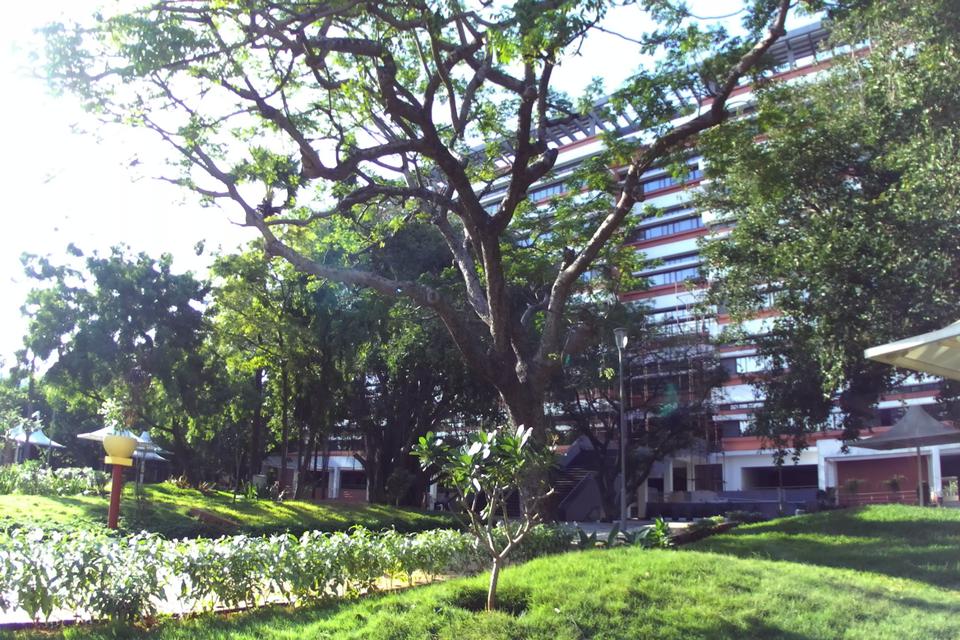}}}\hfill

{\subfloat[$Lotus$]{\includegraphics[height=1.1in,width=1.60in]{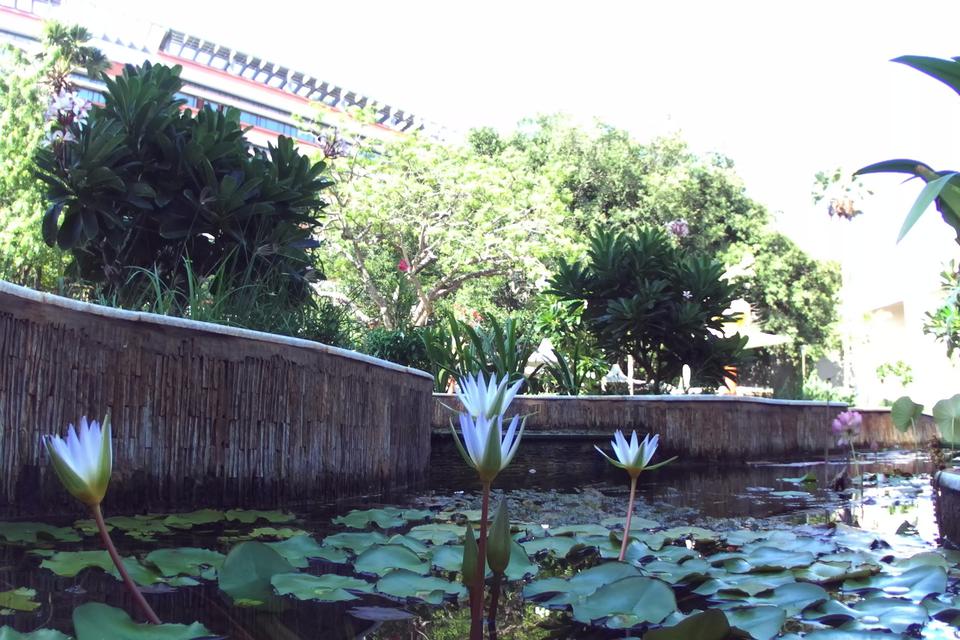}}}\hfill
{\subfloat[$Fountain$]{\includegraphics[height=1.1in,width=1.60in]{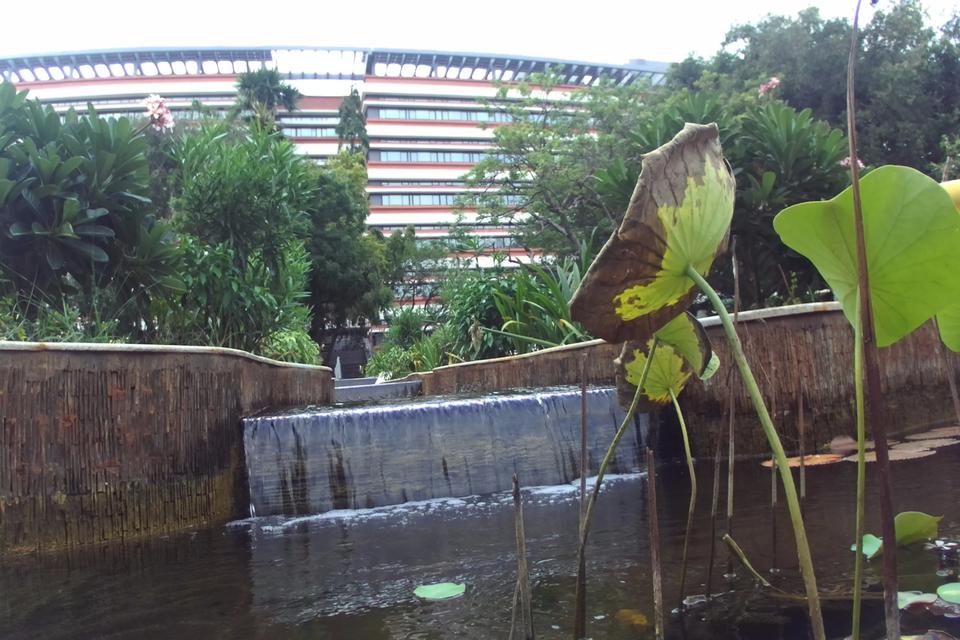}}}\hfill
{\subfloat[$NACSilhoutte$]{\includegraphics[height=1.1in,width=1.60in]{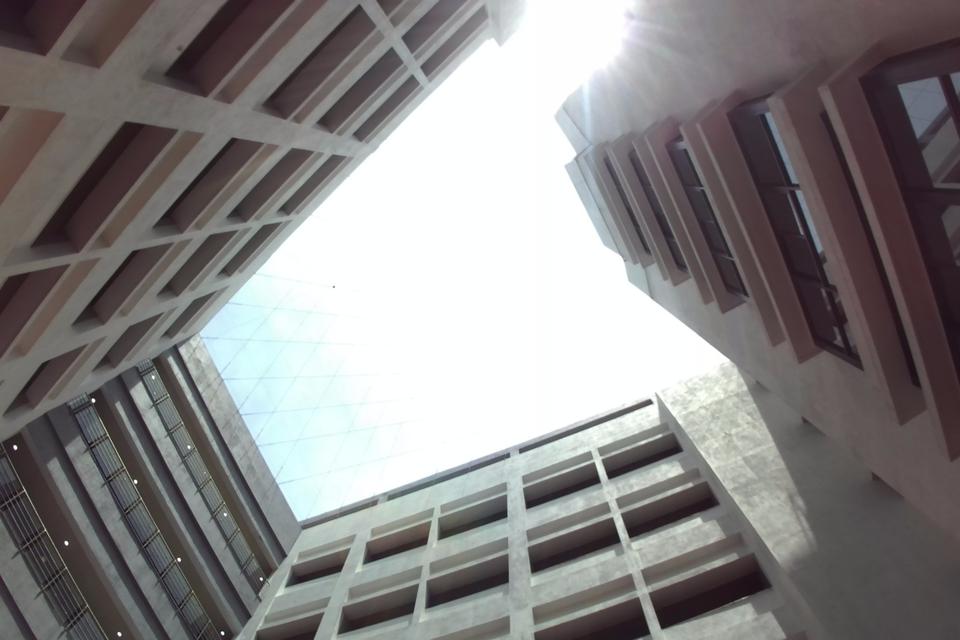}}}\hfill
{\subfloat[$Roadside1$]{\includegraphics[height=1.1in,width=1.60in]{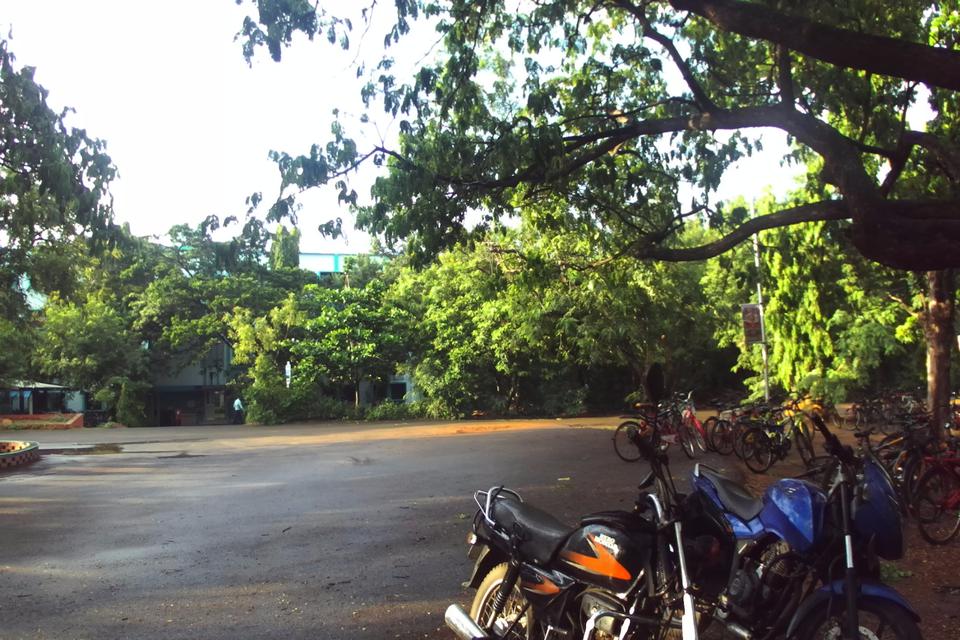}}}\hfill

{\subfloat[$Roadside3$]{\includegraphics[height=1.1in,width=1.60in]{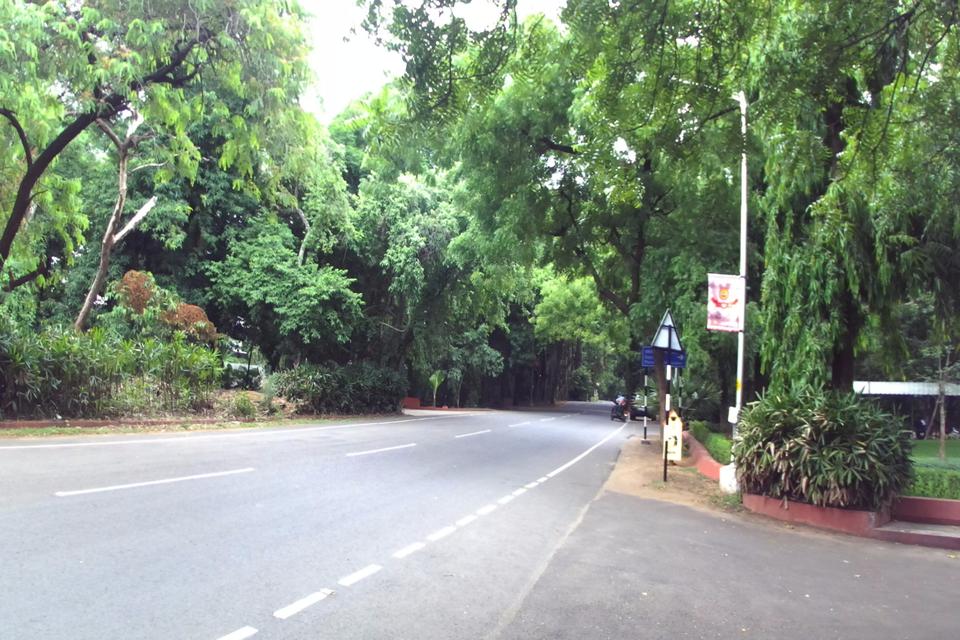}}}\hfill
{\subfloat[$Roadside4$]{\includegraphics[height=1.1in,width=1.60in]{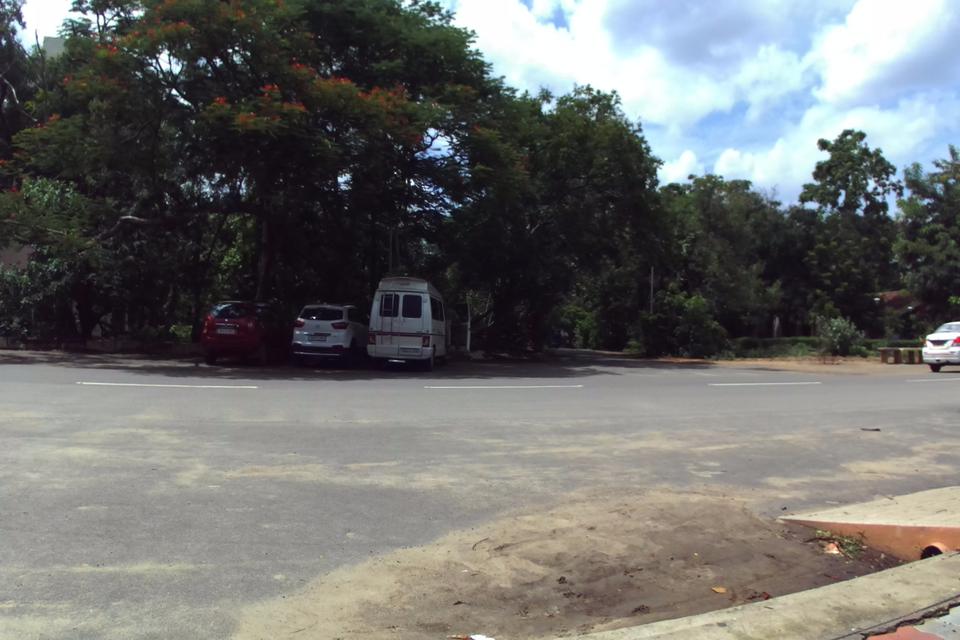}}}\hfill
{\subfloat[$SkylineBT1$]{\includegraphics[height=1.1in,width=1.60in]{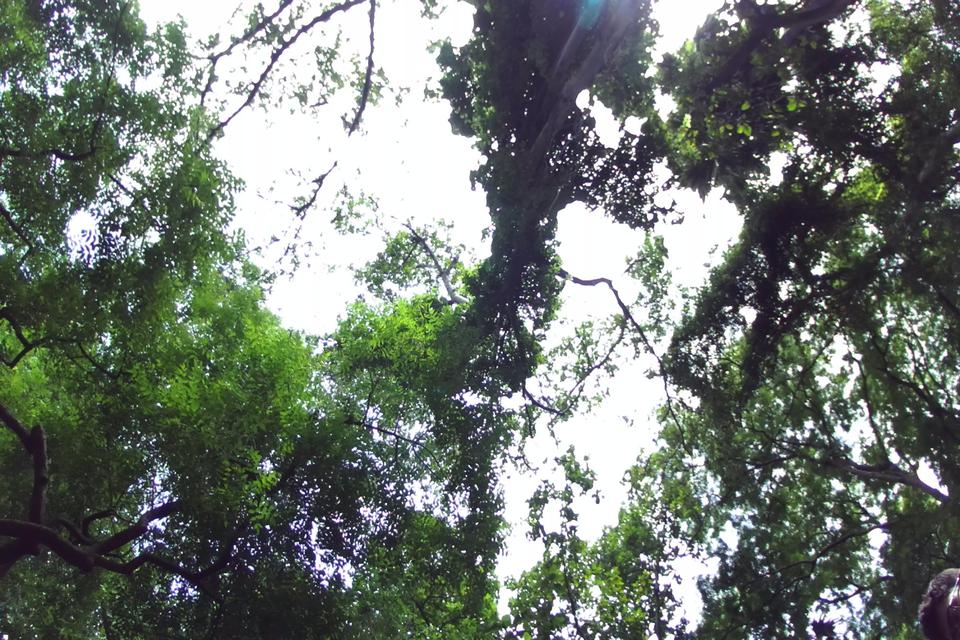}}}\hfill
{\subfloat[$SkylineBT2$]{\includegraphics[height=1.1in,width=1.60in]{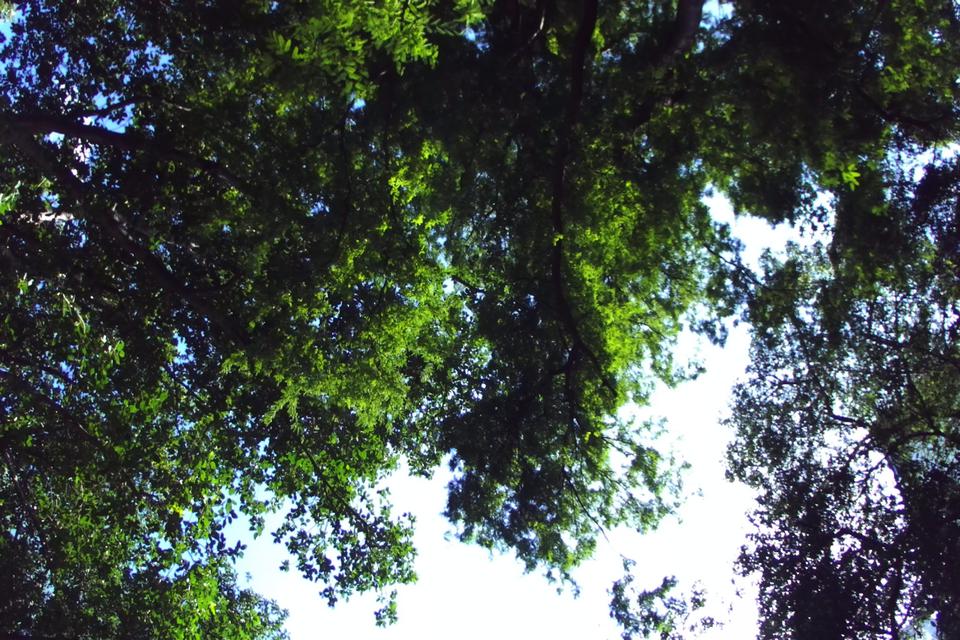}}}\hfill

\caption{Multi-exposure stereoscopic image database scenes. The images shown are left views captured with single camera exposure.}
\label{Image_databaseLEFTViews}
\end{figure*}

{\textbullet \hspace{1 mm} \textbf{Reflective Surfaces}: Features like car window panels, metal surfaces, glass buildings, wet roads, and water bodies contribute to the database's complexity. The Lambertian, glossy and specular surfaces in the outdoor scenes cause abrupt light intensity variations. Therefore, to get the various scene features, it is necessary to capture highly disparate exposures of the same scene. Such captures are more complicated than capturing plain skyscapes because the bright elements in the scenes are dispersed, irregular, and fragmented. Bright patches change with time in scenarios featuring a water surface or moving streams, resulting in no two consecutive multi-exposure images/video frames being similar. Moreover, underwater objects, in our case underwater flora, add a whole new complexity to the scene.}

{\textbullet \hspace{1 mm} \textbf{Outdoor \& Indoor Low-lighting}: Low lighting conditions lead to less saturation, low illumination, and less texture in outdoor scenarios thus creating a considerable domain shift compared to daytime outdoor scenes. The domain shift is the transition from daytime conditions (well-lit and uniform illumination) to night-time conditions (poor illumination/visibility and non-uniform illumination). Not many datasets exist, which include outdoor night/evening scenes. Our database incorporates such complex scenes as well. Indoor photography under artificial illumination is also investigated. In artificially light conditions, most HDR cameras fail to deliver good results. Our dataset uses the ZED stereo camera's outstanding low-light sensitivity and includes artificially illuminated interior scenes. Furthermore, the complexity of the indoor scene from the dataset is increased by interference caused by the artificial light source.}

{\textbullet \hspace{1 mm} \textbf{Complex Movements}: The dataset features scenes with many objects moving independently in a complex environment. It accommodates the non-rigid motion of human beings, animals, vehicles and other objects. To maintain precise frame-level alignment, the camera is kept stationary between subsequent shots. As the scenes are real, the objects in the image, such as forest trees swinging in the breeze, rustling leaves, birds flying, flowing water, moving vehicles and so on, move somewhat between different exposure captures. These types of movements are inherent while capturing natural scenes since it is beyond our control. Few images/videos are explicitly captured indoors to analyze the motion in the foreground while having a stable background. Additionally, to diversify the data, a few scenes are shot with certain camera movement to serve the purpose of real-time data. }

It is in fact, these characteristics which distinguishes our dataset, and  open up new research possibilities beyond traditional depth and tone estimation algorithms. The proposed dataset presents a new set of challenges for researchers, including the use of learning algorithms to create high-quality 3D HDR images and videos \cite{Im2019DPSNetED, Wu_2018_ECCV}.

\begin{table*}[t!]
  
  \centering
  \caption{ A multi-exposure image set from the proposed database. Each row consists of the captured left views of a scene taken from the ZED camera at multiple exposures (here at three different camera exposure levels). Scene detail and related intricacy are mentioned. The set consist of scenes: $Lotus$, $GlassBuilding$, $Canopy$, $Baba$ and $Hall$ (top to bottom).}
  \begin{tabular}{  | c  c  c  m{6cm} | }
    \hline
    \begin{minipage}{.18\textwidth}
      \includegraphics[width=\linewidth]{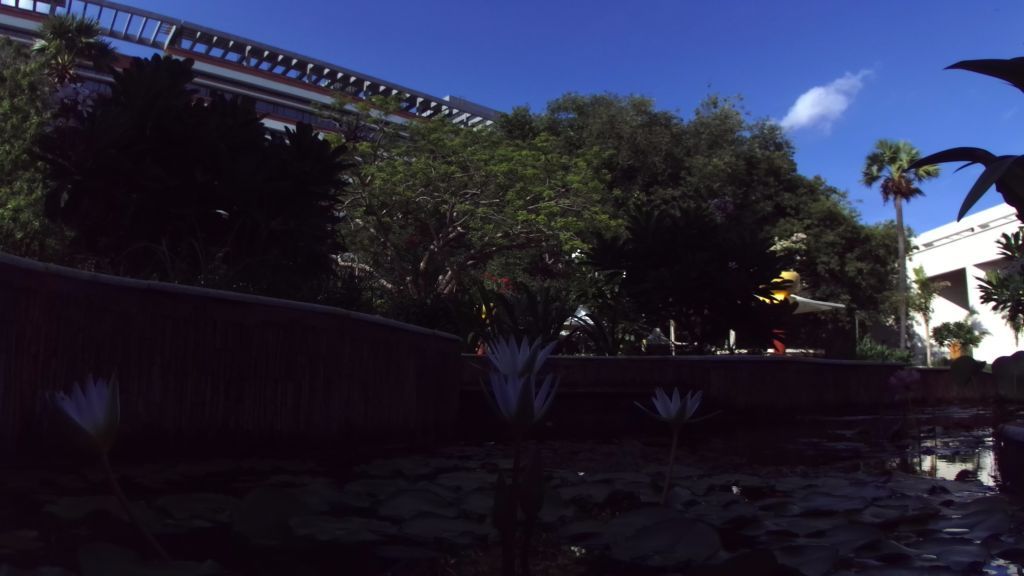}
    \end{minipage}
    &
    \begin{minipage}{.18\textwidth}
      \includegraphics[width=\linewidth]{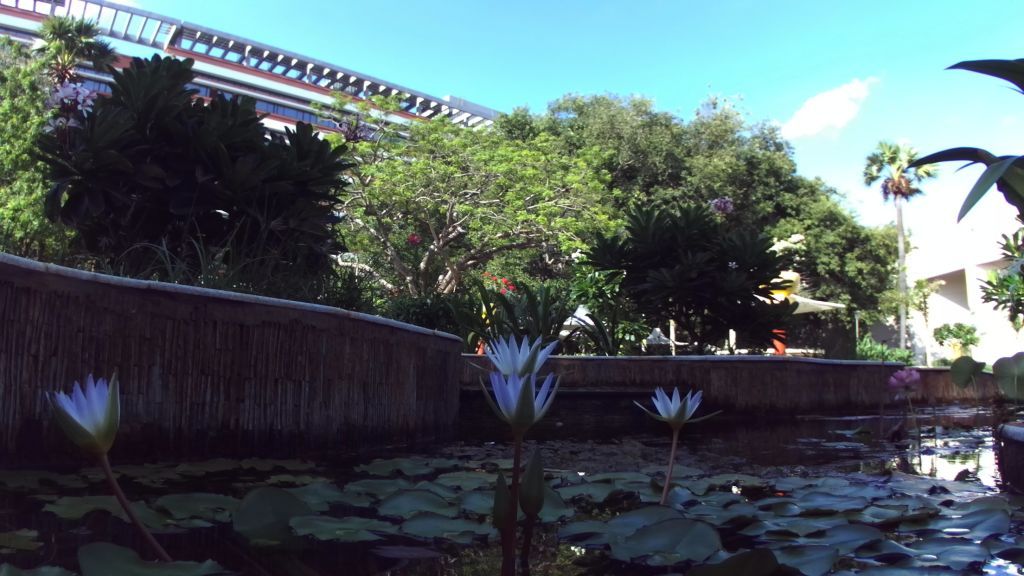}
    \end{minipage}
    &
    \begin{minipage}{.18\textwidth}
      \includegraphics[width=\linewidth]{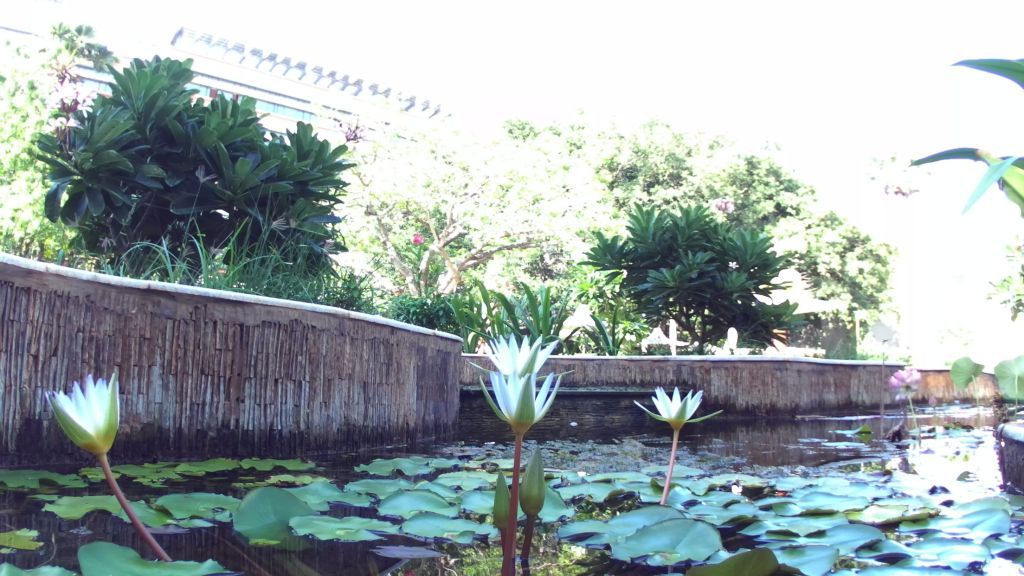}
    \end{minipage}
    &
    \vspace{0.05in}
    Scene captures wide dynamic range - layers of vegetation, sky, building, rich variations of green, slight object motion, high detail, complex depth structure, reflection from water. \vspace{0.05in}
    \\ \hline
    \begin{minipage}{.18\textwidth}
      \includegraphics[width=\linewidth]{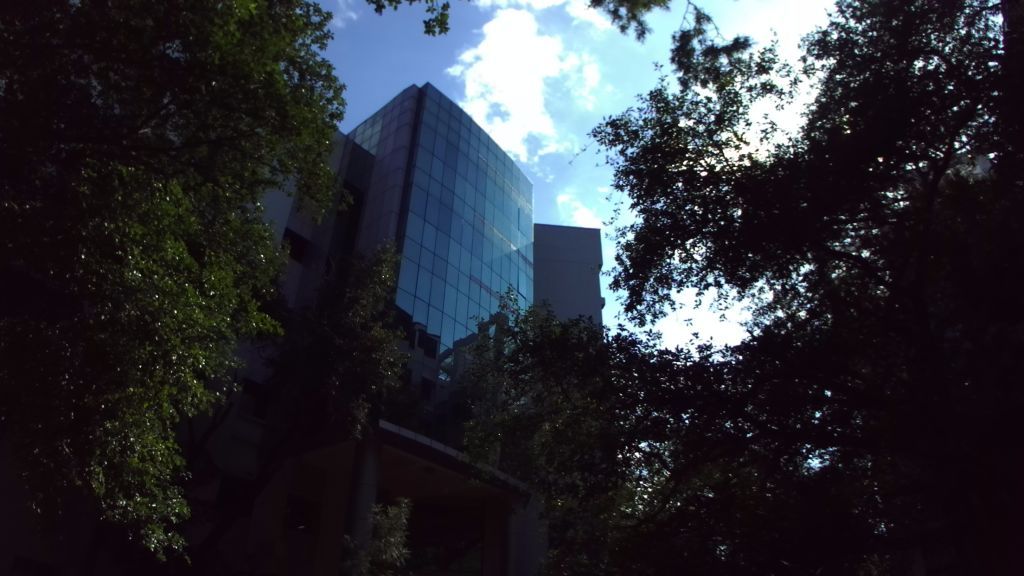}
    \end{minipage}
    &
    \begin{minipage}{.18\textwidth}
      \includegraphics[width=\linewidth]{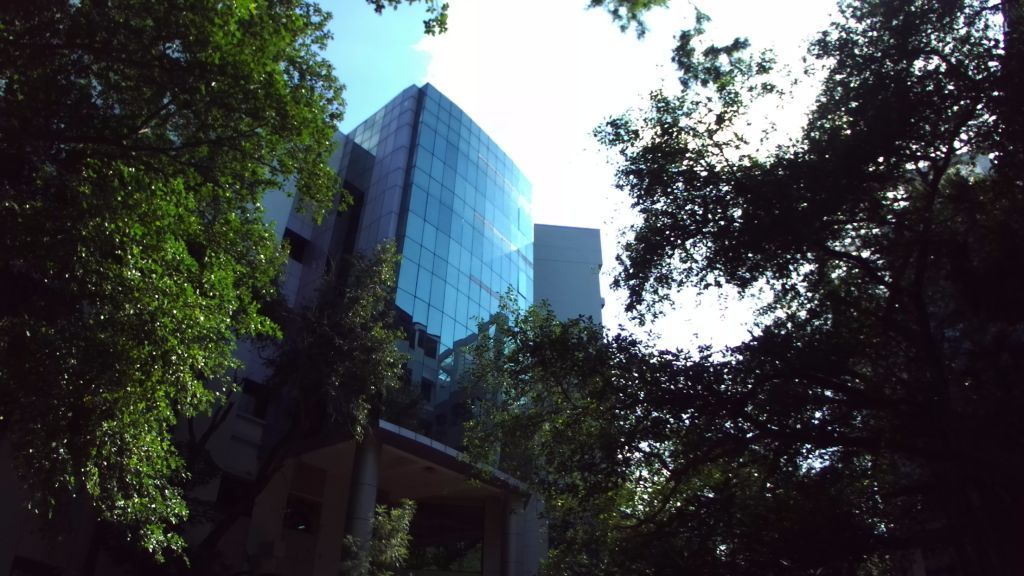}
    \end{minipage}
    &
    \begin{minipage}{.18\textwidth}
      \includegraphics[width=\linewidth]{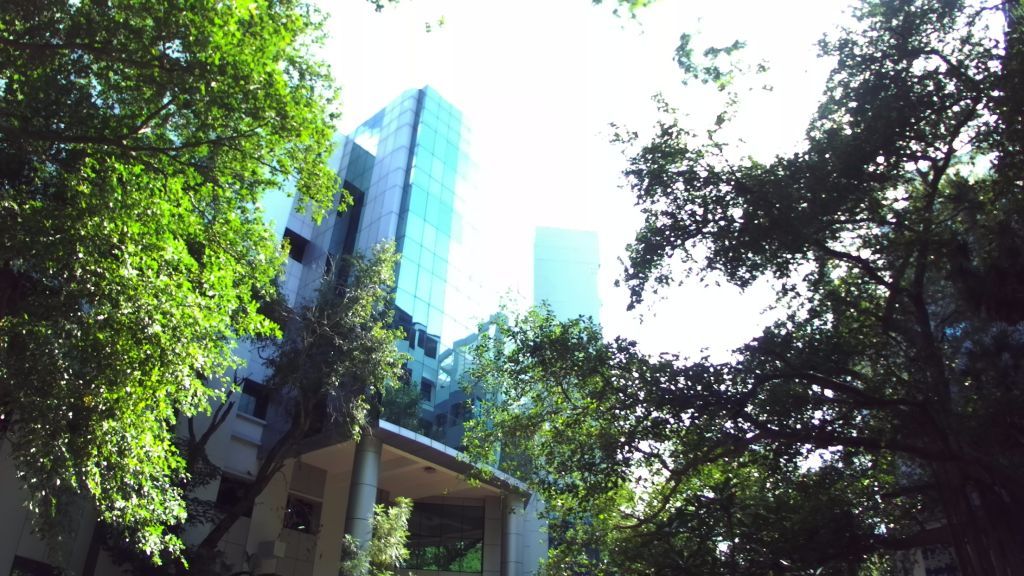}
    \end{minipage}
    &
    \vspace{0.05in}
    Scene depicts rich variation of features; reflection from glass surface, rich shades of green, slight motion of trees, cloud patterns, high detail, complex depth structure. \vspace{0.05in}
    \\ \hline
    
    \begin{minipage}{.18\textwidth}
      \includegraphics[width=\linewidth]{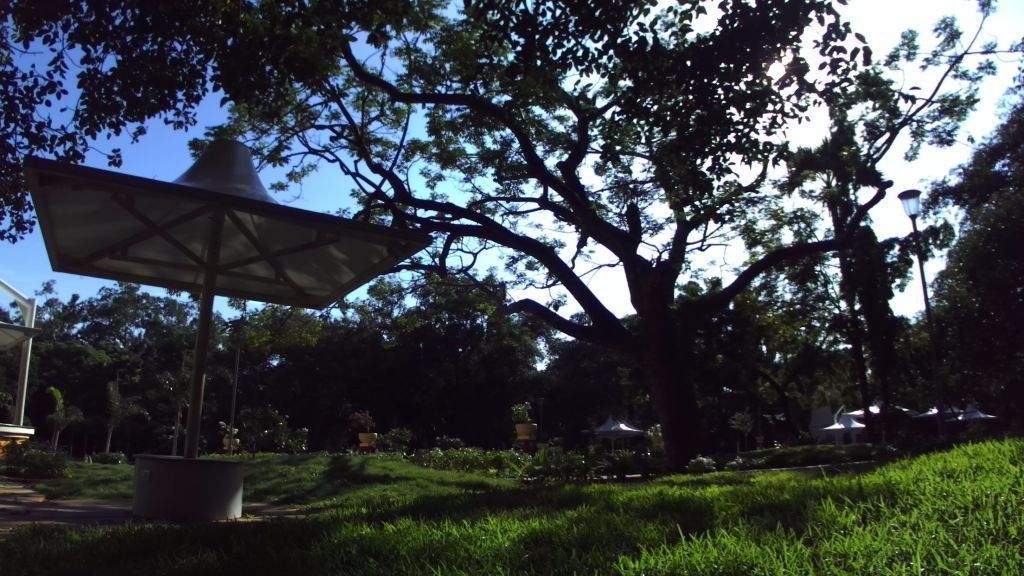}
    \end{minipage}
    &
    \begin{minipage}{.18\textwidth}
      \includegraphics[width=\linewidth]{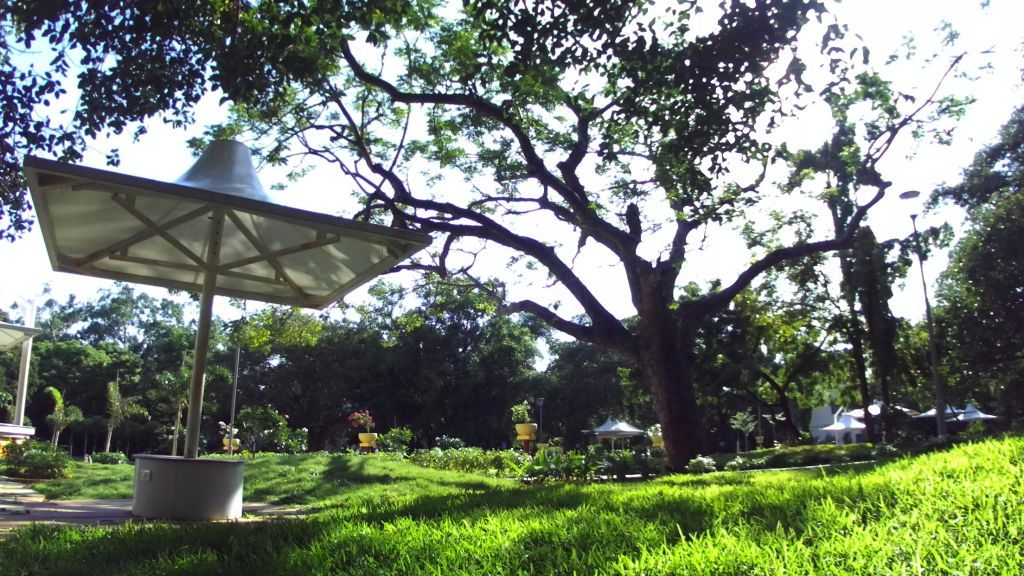}
    \end{minipage}
    &
    \begin{minipage}{.18\textwidth}
      \includegraphics[width=\linewidth]{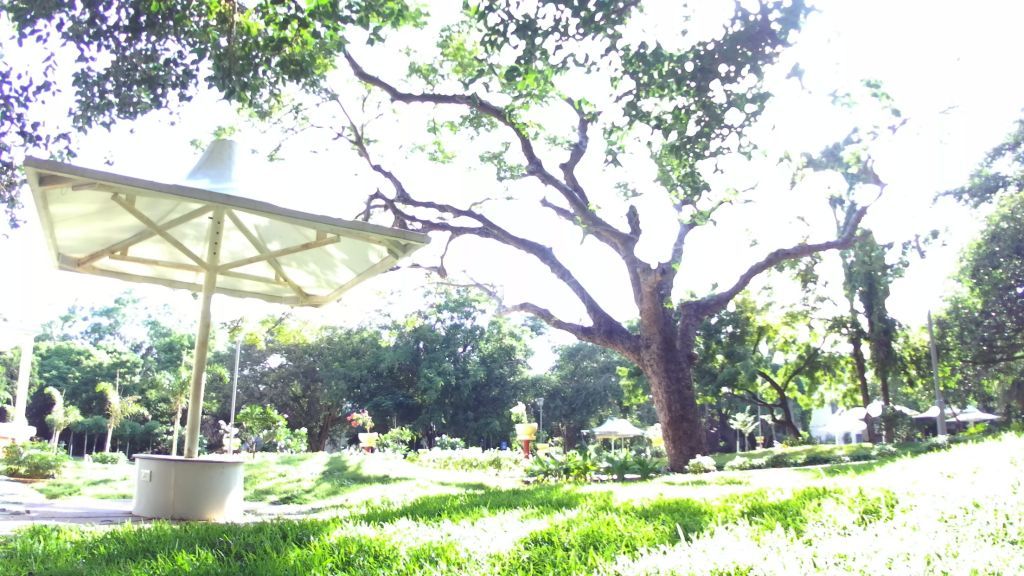}
    \end{minipage}
    &
    \vspace{0.05in}
    Branches and leaves silhouetted against the sky; rich shades of greens, moderate object motion, intricate details, complex depth structure, complex interplay of sunlight.\vspace{0.05in}
    \\ \hline
    
    \begin{minipage}{.18\textwidth}
      \includegraphics[width=\linewidth]{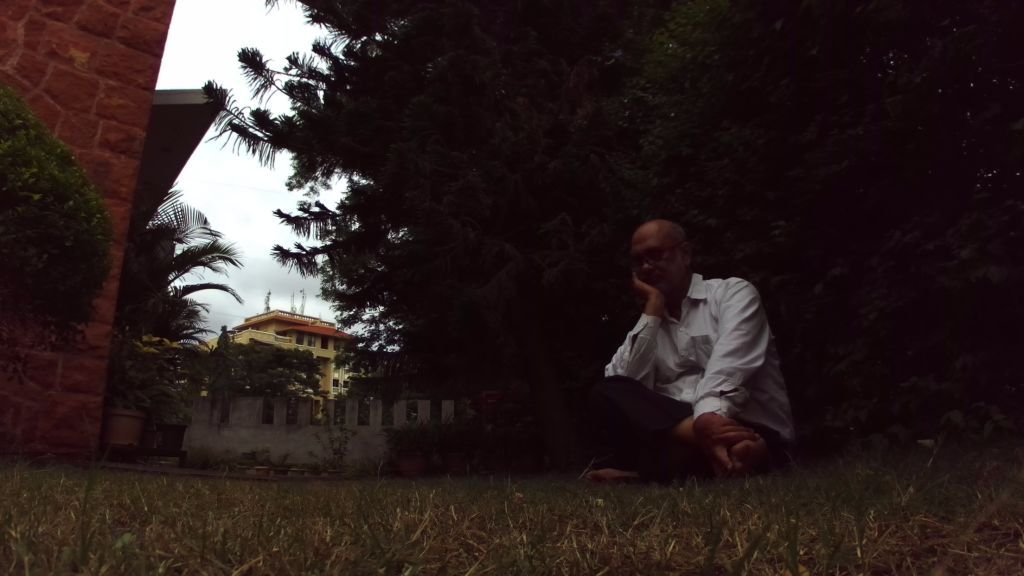}
    \end{minipage}
    &
    \begin{minipage}{.18\textwidth}
      \includegraphics[width=\linewidth]{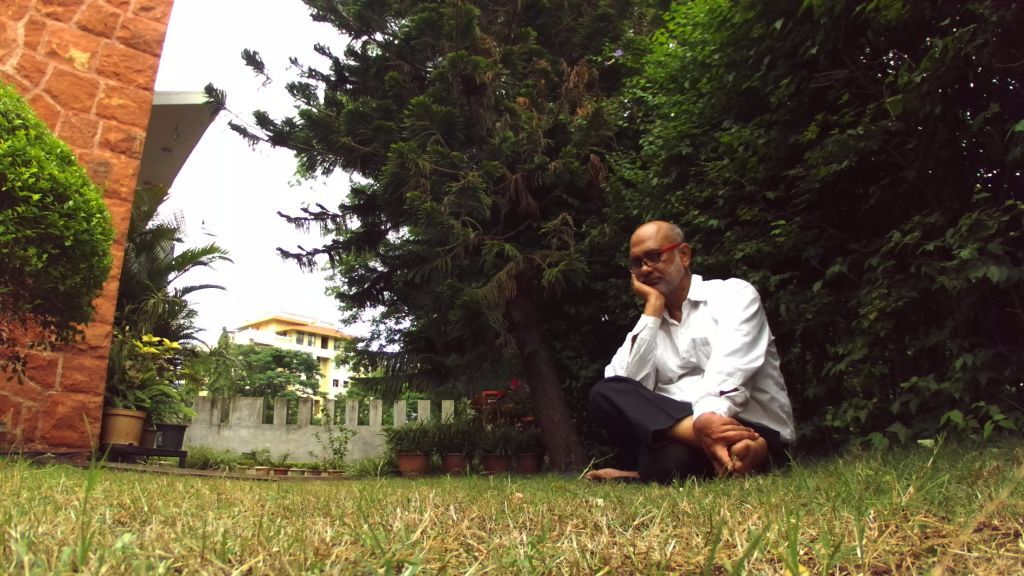}
    \end{minipage}
    &
    \begin{minipage}{.18\textwidth}
      \includegraphics[width=\linewidth]{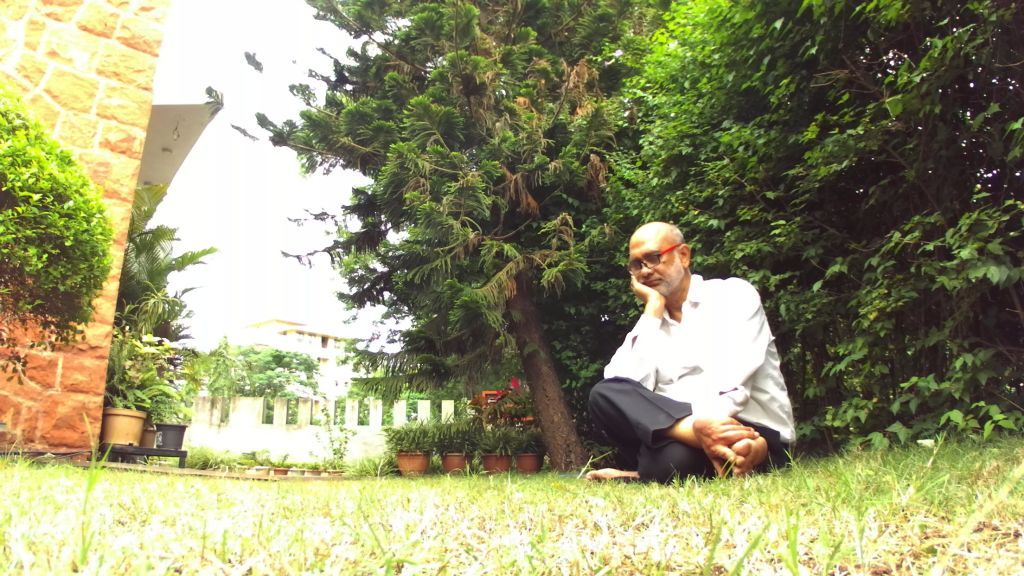}
    \end{minipage}
    &
    \vspace{0.05in}
    Distant buildings seen clearly at lower exposures, intricate details of plants seen at higher exposures. Wide dynamic range, complex depth structure, vegetation motion. \vspace{0.05in}
    \\ \hline
    
    \begin{minipage}{.18\textwidth}
      \includegraphics[width=\linewidth]{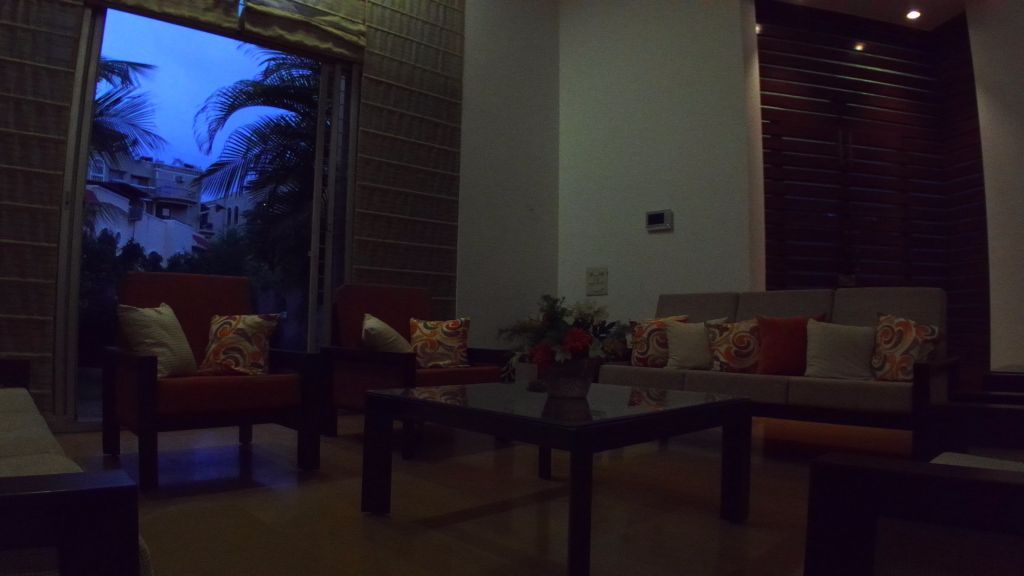}
    \end{minipage}
    &
    \begin{minipage}{.18\textwidth}
      \includegraphics[width=\linewidth]{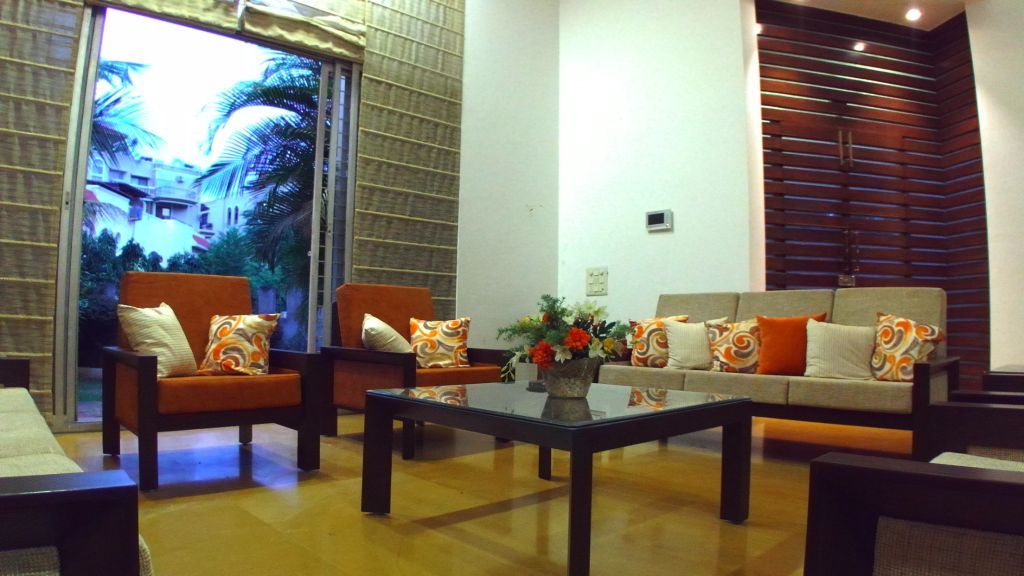}
    \end{minipage}
    &
    \begin{minipage}{.18\textwidth}
      \includegraphics[width=\linewidth]{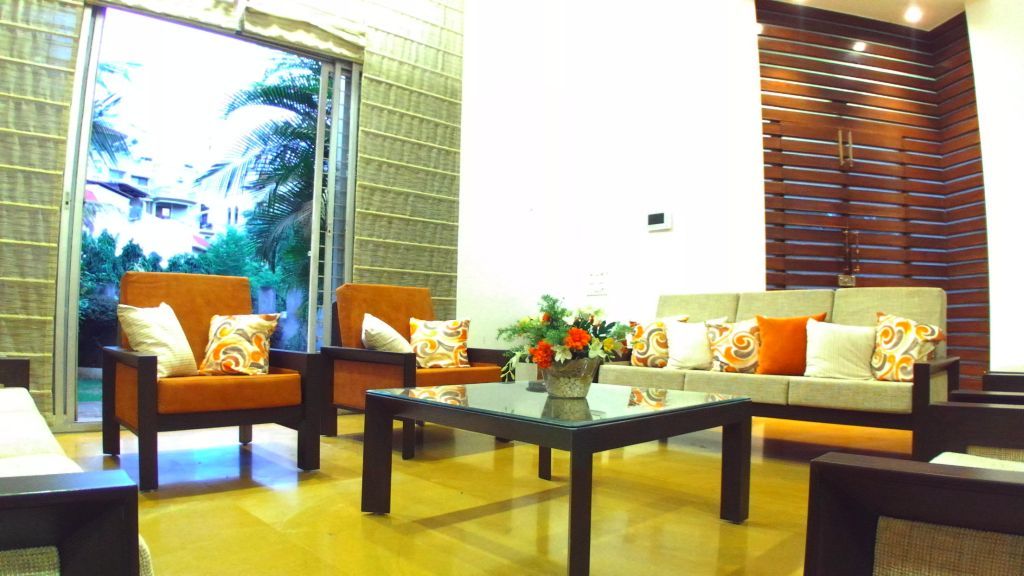}
    \end{minipage}
    &
    \vspace{0.05in}
    Two starkly different lighting conditions – artificially lit interior(brightly lit, non-uniformly illuminated) and outside dusk-time view (lowly lit, uniformly illuminated). Moderate details, moderate depth structure. \vspace{0.05in}
    \\ \hline
   
  \end{tabular}
  \label{table3:Imagesample}
\end{table*}

\subsection{Multi-exposure stereoscopic image dataset} 
Our image dataset includes 39 stereoscopic multi-exposure scenes. The resolution of each view is 2208$\times$1242 pixels (full HD). Forests, highways, houses, skylines, water fountains, indoor and low-lit scenes are all included in the dataset. Leaves, branches swinging in the wind ($Banyan1$, $BanyanAvenue$) introduce complexity due to intricate light interplay. Fig. \ref{Image_databaseLEFTViews} shows left view of 20 scenes (corresponding to a single exposure).

Scenes featuring objects with varying reflectance qualities such as glasses of buildings ($NACBuilding$, $Glass$ $Building$), car windows panels ($Roadside1, Roadside4$), highly reflective building walls ($House1$, $House2$), road ($Roadside3, Marsh$), water stream ($Lotus$, $Fountain$), metal/ plastic ($Canopy$) etc., are included in the proposed dataset. Specular reflection, a fundamental and universal reflection process, is viewpoint dependent. Even with slight changes in perspective, corresponding regions in stereo images are poorly correlated due to specular reflection-shift on the object surface \cite{Bhat98stereoand}. This impacts depth estimation of outdoor stereo scenes, thus limiting the 3D HDR research domain.

HDR reconstruction of images having clouds and brightly illuminated sky in the background ($NACsilhoutte$,  $GCLong$, $Pergola$, $HimalayaBuilding$, $Skyline BT1$, $Skyline BT2$) is challenging. As for high camera exposure, not much detail is obtained for the cloud pattern. Features related to the clouds can only be achieved at significantly low exposures for such scenarios. While outdoor scenes have natural light (well-lit and uniform illumination) and indoor scenes have artificial light (poorly-lit and non-uniform illumination), semi-outdoor scenes such as $Hall$ have both features provided by artificial and natural light. A brief description of a few scenes with images captured under varying exposure is given in Table \ref{table3:Imagesample}. 

\subsection{Multi-exposure stereoscopic video dataset} 

The proposed video dataset contains 18 stereoscopic video clips. Among these clips, 16 are captured with a still camera mounted on a rotatable tripod. Two video clips, i.e., $CoconutTree$ and $ForestDusk$, are captured with significant camera motion while the camera is handheld. The proposed scenes consist of slightly moving objects such as swaying trees and grass blades, objects with traceable motion such as Ferris wheel ($FerrisWheel$) and Pirate ship ride movement ($Rides$), and objects with non-traceable significant motion such as moving people and vehicles. Each video clip is captured at a frame rate of 30 frames per second and has a resolution of 1920 × 1080. Each clip has about 100$-$200 frames, making it long enough for experimentation and research. About 3$-$6 video clips are provided for each scene captured at different camera exposures. All of the videos are synced in time. To avoid frame misalignment, each frame is left undisturbed between successive shots. The motion of objects within the frame makes it difficult to duplicate the same motion/video sequence at successive exposures since many exposures are collected sequentially.

\begin{table*}[t!]
  \caption{Multi-exposure stereoscopic video database scene description, part (a). Each scene is represented by a stereo pair captured at a particular exposure level using ZED camera.}
  \centering
  \begin{tabular}{ | c | m{1.8cm} | c |  m{5.5cm} | }
    \hline
    
    \centering{\textbf{No.}} & \centering{\textbf{Name}} & \textbf{3D HDR Video scene} & \textbf{ Characteristics} \\ \hline
    1 & GC 2 &
    \begin{minipage}{.4\textwidth}
      \includegraphics[width=\linewidth]{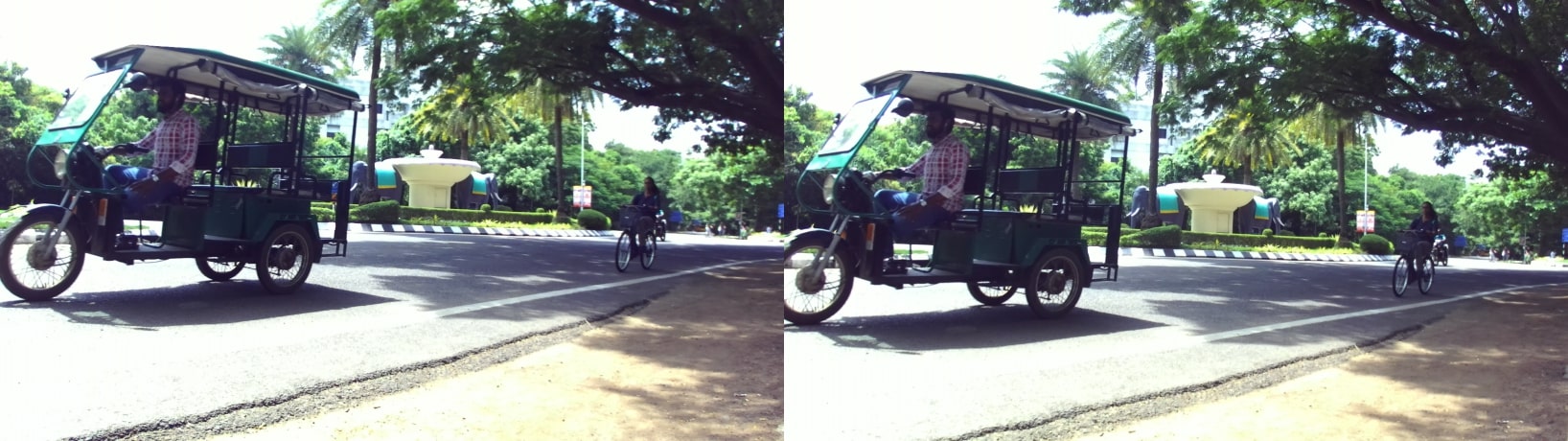}
    \end{minipage}
    &
    \vspace{0.05in}
    Road scene with moving vehicles; significant object motion, intricate background motion, no camera motion, high detail, complex depth structure, rich chromatic variation, natural light.\vspace{0.05in}
    \\ \hline
    
    2 & Garden Baba &
    \begin{minipage}{.4\textwidth}
      \includegraphics[width=\linewidth]{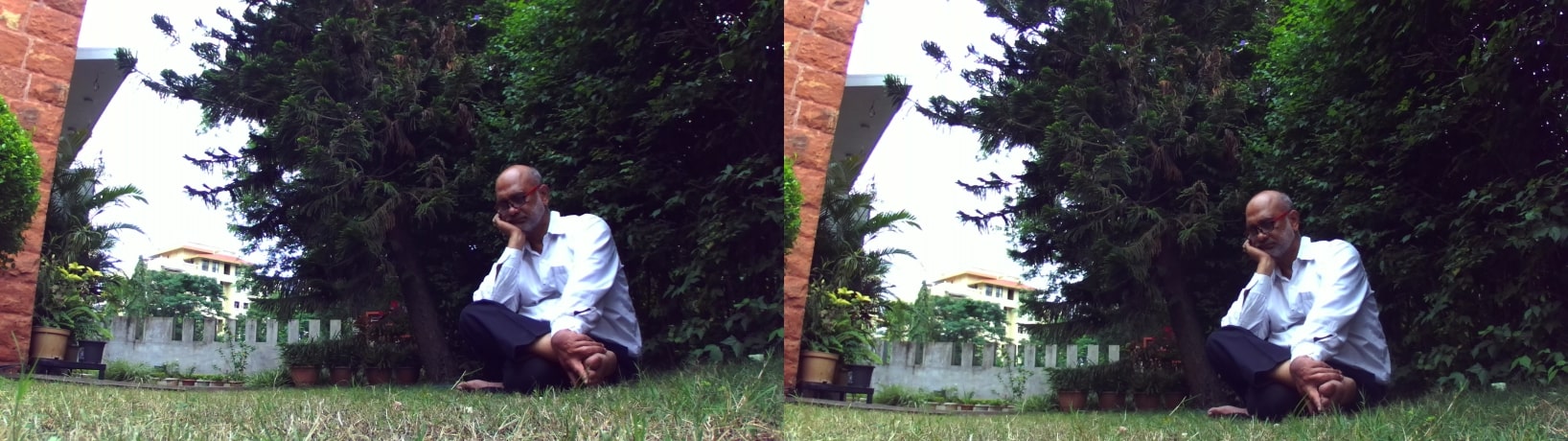}
    \end{minipage}
    &
    \vspace{0.05in}
    Person sitting in garden; rigorous movement of tree branches in wind; slight object motion (man), no camera motion, complex depth structure, rich chromatic variation, high details.\vspace{0.05in}
    \\ \hline
    
    3 & Lotus &
    \begin{minipage}{.4\textwidth}
      \includegraphics[width=\linewidth]{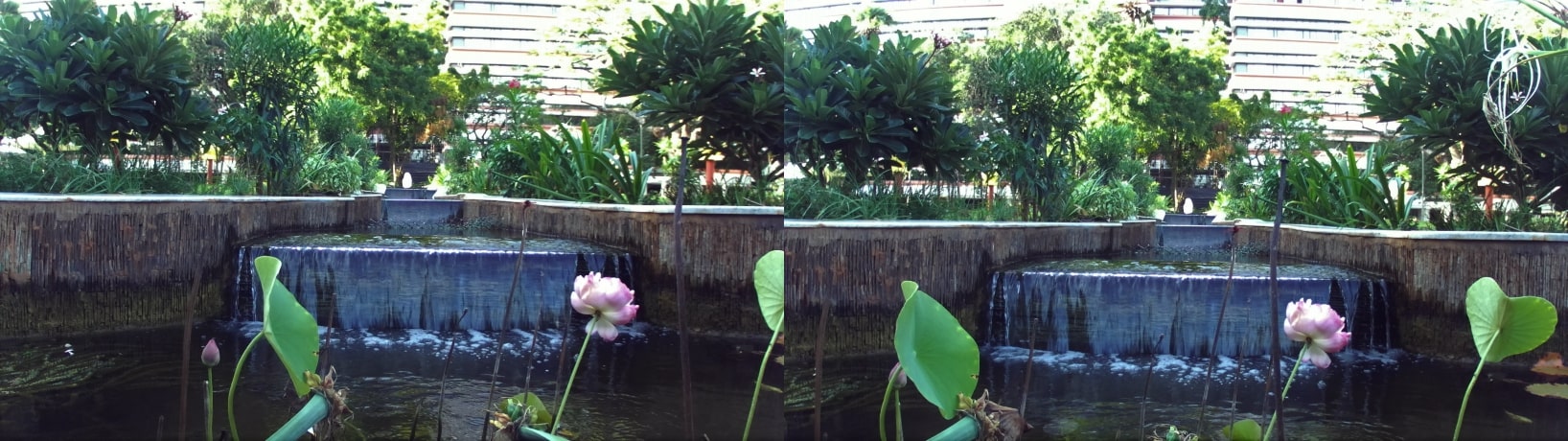}
    \end{minipage}
    &
    \vspace{0.05in}
    Lotus in pond; flowing water, reflections and transparency, slight object motion, intricate background motion, no camera motion, high detail, complex depth structure, natural light.\vspace{0.05in}
    \\ \hline
    4 & Ferris wheel &
    \begin{minipage}{.4\textwidth}
      \includegraphics[width=\linewidth]{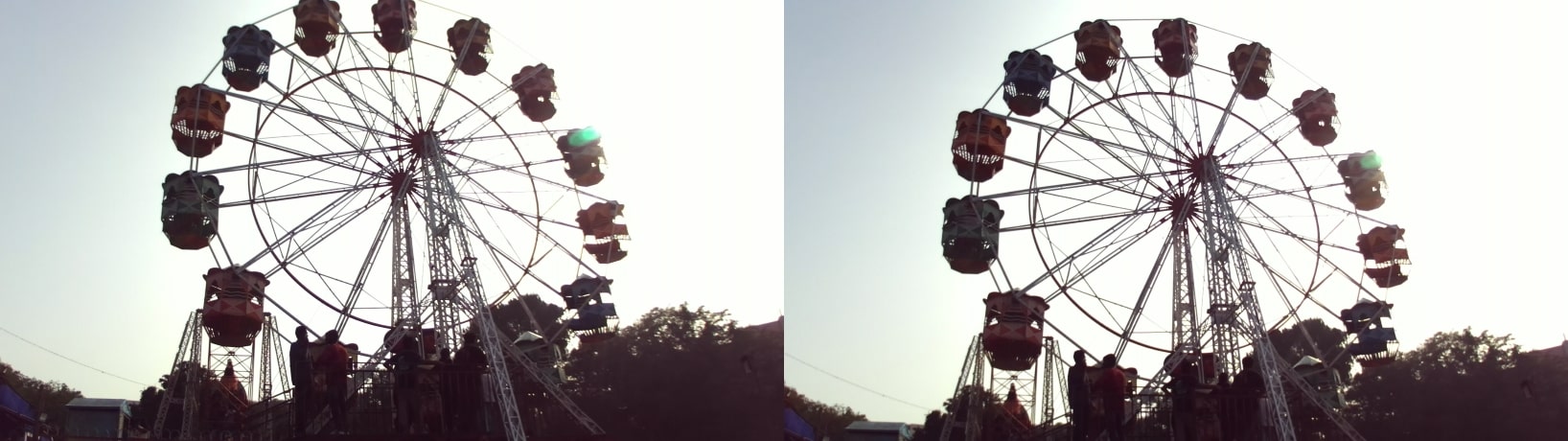}
    \end{minipage}
    &
  \vspace{0.05in} 
  Ferris wheel in motion; partially traceable object motion, , no camera motion, moderate detail, simple depth structure, complex light interplay due to direct sunlight, silhouette of objects \& trees.\vspace{0.05in}
    \\ \hline
    
    5 & Coconut tree &
    \begin{minipage}{.4\textwidth}
      \includegraphics[width=\linewidth]{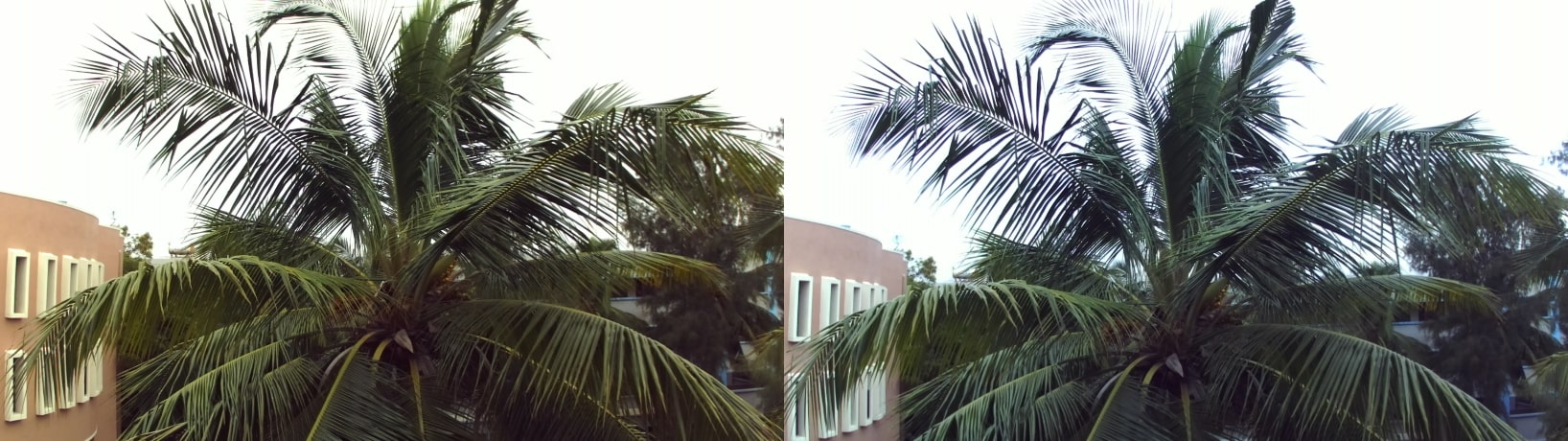}
    \end{minipage}
    &
    \vspace{0.05in}
    Moving video of coconut trees; stationary objects, hand held camera motion, high detail due to complex occlusion patterns of coconut leaves, complex depth structure, natural light.\vspace{0.05in}
    \\ \hline
    
     6 & GC 1 &
    \begin{minipage}{.4\textwidth}
      \includegraphics[width=\linewidth]{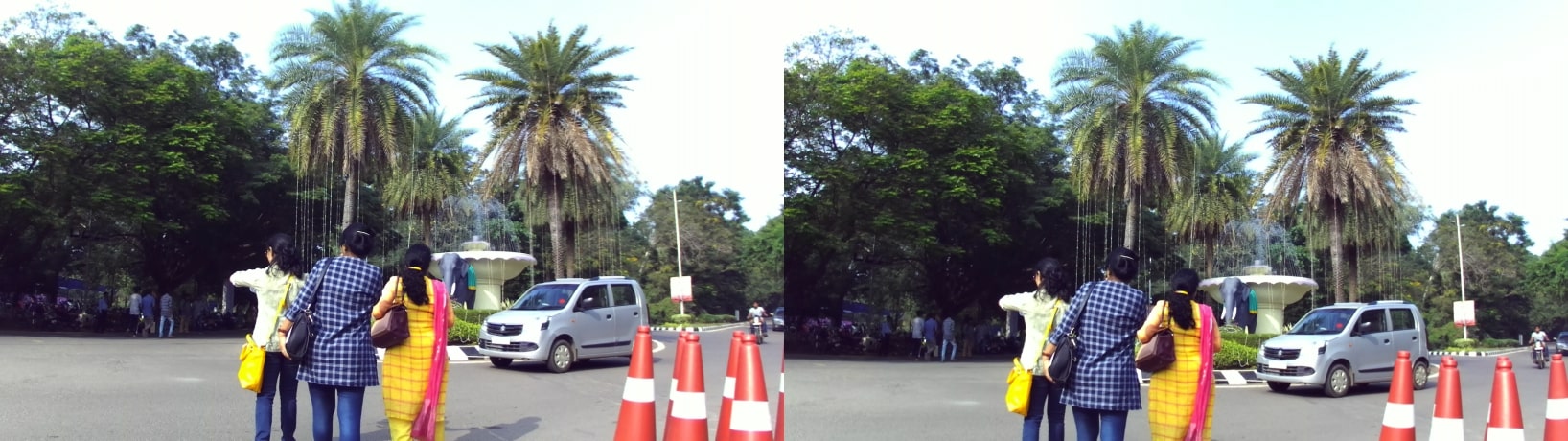}
    \end{minipage}
    &
    \vspace{0.05in}
    People crossing a junction while chatting;  Complex object motion, no camera motion, intricate background motion (vegetation), moderate detail, complex depth structure, natural light.\vspace{0.05in}
    \\ \hline
    
    7 & Singer &
    \begin{minipage}{.4\textwidth}
      \includegraphics[width=\linewidth]{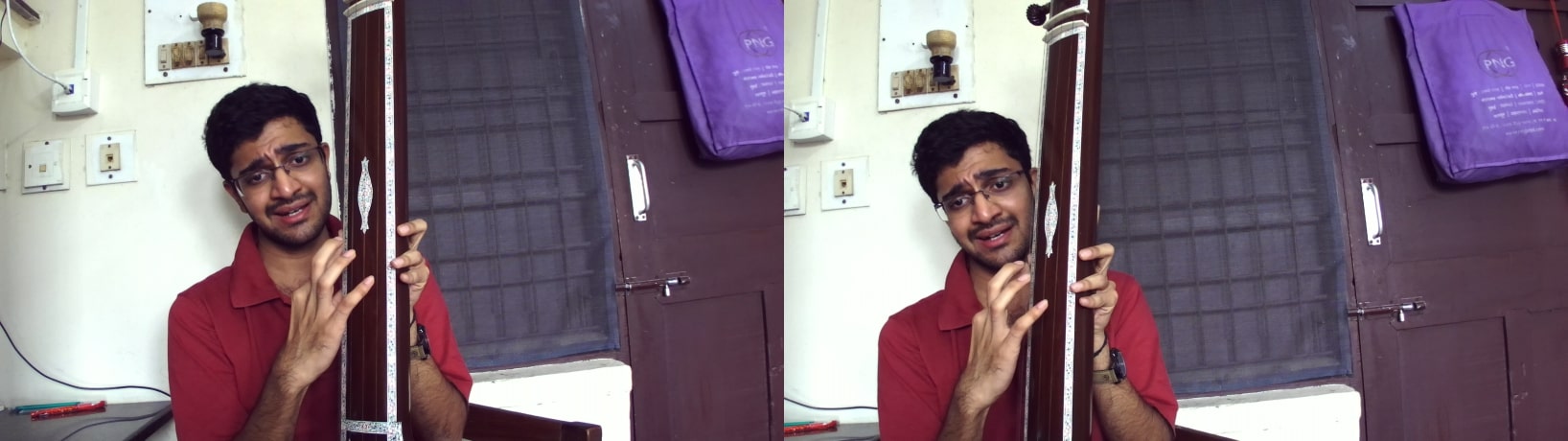}
    \end{minipage}
    &
    \vspace{0.05in}
    Person singing \& playing stringed instrument; complex finger and body motion, no background motion, no camera motion, moderate detail, medium depth structure, artificial light.\vspace{0.05in}
    \\ \hline
    
    8 & Gurunath &
    \begin{minipage}{.4\textwidth}
      \includegraphics[width=\linewidth]{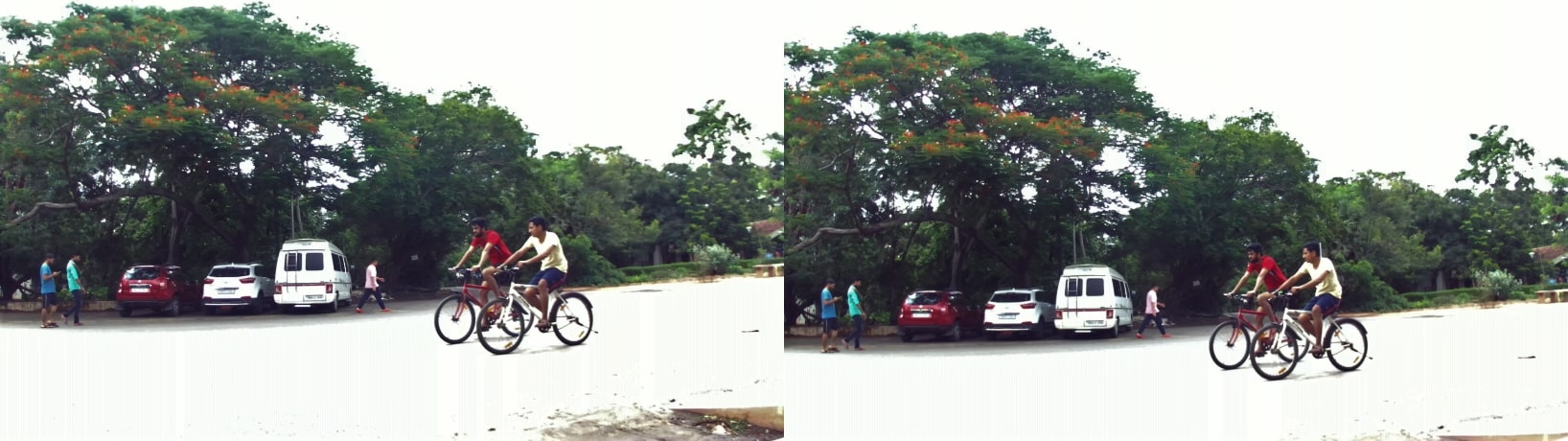}
    \end{minipage}
    &
    \vspace{0.05in}
    Cyclists on road; Complex object motion of passers by on cycle and on foot, no camera motion, intricate background motion (vegetation), complex depth structure, natural light.\vspace{0.05in}
    \\ \hline
    
     9 & Saarang &
    \begin{minipage}{.4\textwidth}
      \includegraphics[width=\linewidth]{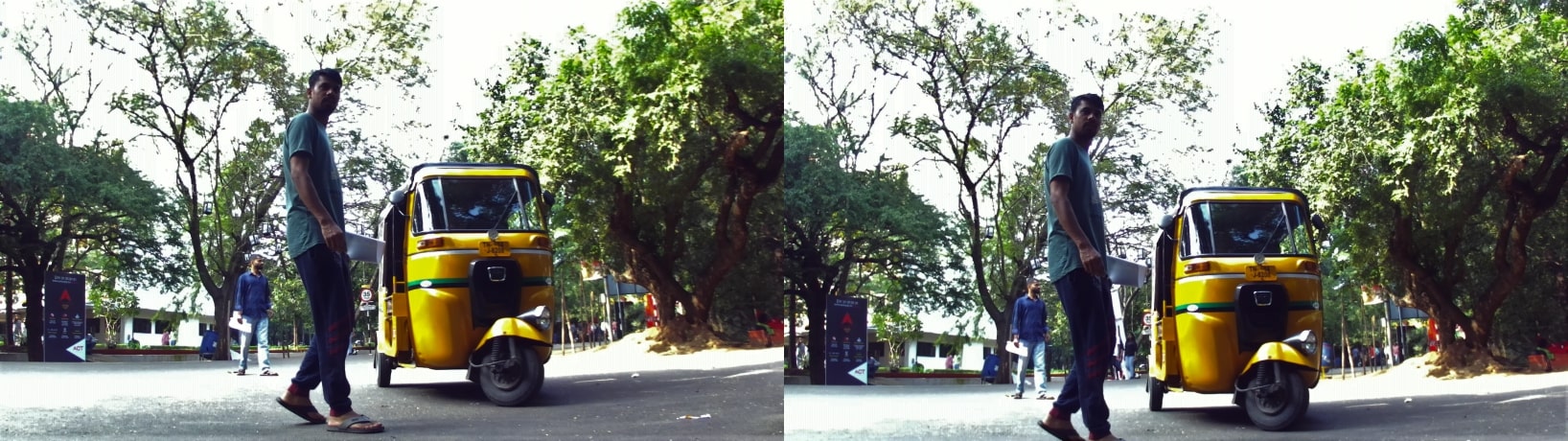}
    \end{minipage}
    &
    \vspace{0.05in}
    College festival; numerous vehicles and students, complex unpredictable motion, significant background motion (vegetation), no camera motion, high detail, natural light.\vspace{0.05in}
    \\ \hline
    
    
  \end{tabular}
  \label{Table_Video_A}
\end{table*}

\begin{table*}[t!]
  \caption{Multi-exposure stereoscopic video database scene description, part (b). Each scene is represented by a stereo pair captured at a particular exposure level using ZED camera.}
  \centering
  \begin{tabular}{ | m{0.35cm} | m{1.8cm} | c | m{5.5cm} | }
    \hline
    
    \centering{\textbf{No.}} & \centering{\textbf{Name}} & \textbf{3D HDR Video scene} & \textbf{Characteristics}
    \\ \hline
    10 & Cafe &
    \begin{minipage}{.4\textwidth}
      \includegraphics[width=\linewidth]{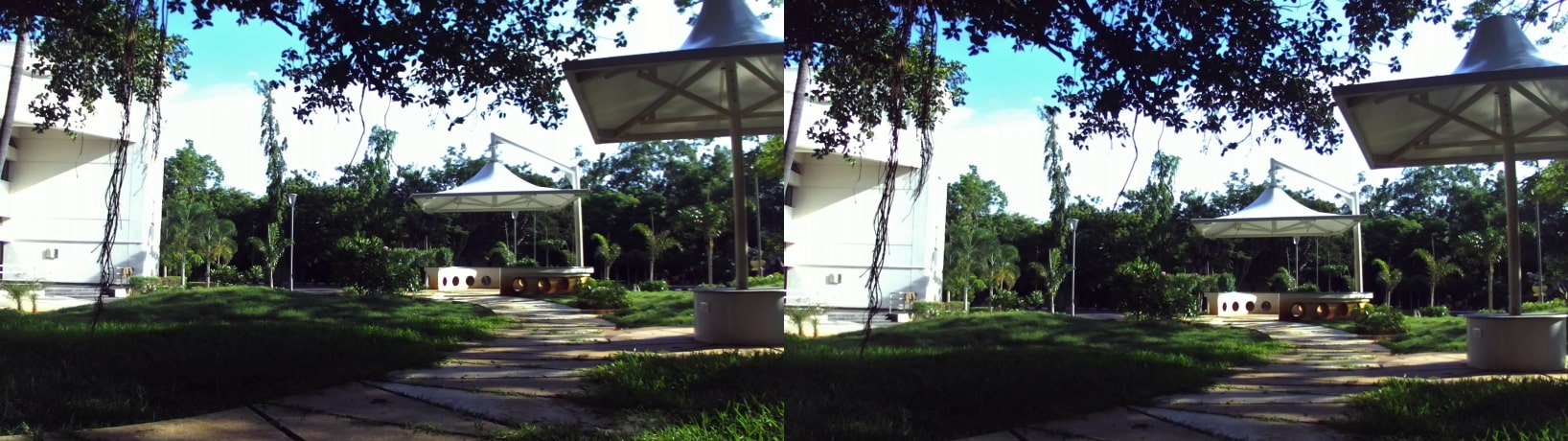}
    \end{minipage}
    &
    \vspace{0.05in}
    Evening view of canteen; foreground motion(people, cycle, animal), intricate background motion, no camera motion, high detail, complex depth structure, natural light.\vspace{0.05in}
    \\ \hline
    
    11 & Rides &
    \begin{minipage}{.4\textwidth}
      \includegraphics[width=\linewidth]{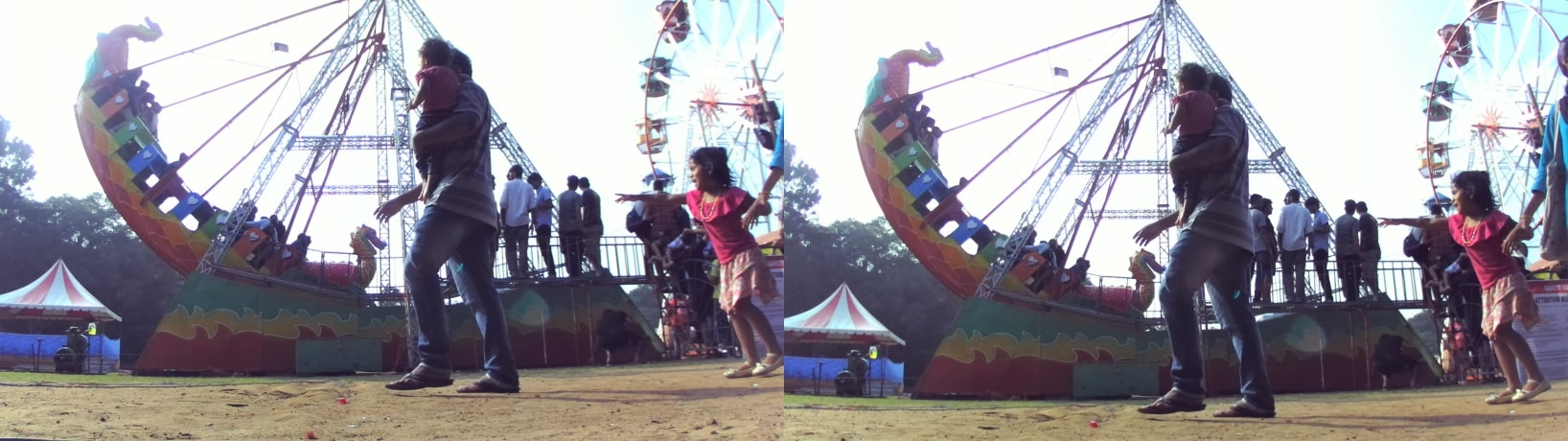}
    \end{minipage}
    &
    \vspace{0.05in}
    Ferris wheel \& Pirate ship ride in motion; people walking past (complex movements), no camera motion, moderate detail, medium depth structure, complex direct sunlight interplay.\vspace{0.05in}
    \\ \hline
    
    12 & Cafe skyline &
    \begin{minipage}{.4\textwidth}
      \includegraphics[width=\linewidth]{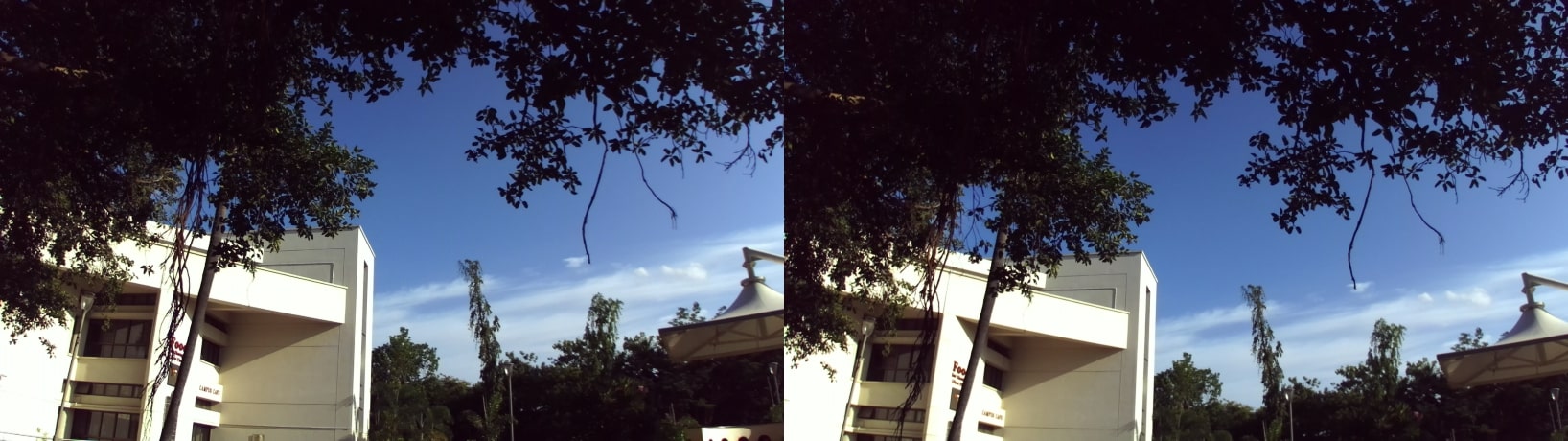}
    \end{minipage}
    &
    \vspace{0.05in}
    Banyan Roots \& tree branches swaying in wind; Complex object motion, no camera motion, sky \& clouds, high detail, medium depth structure, natural light, silhouette of leaves.\vspace{0.05in}
    \\ \hline
    13 & Avenue &
    \begin{minipage}{.4\textwidth}
      \includegraphics[width=\linewidth]{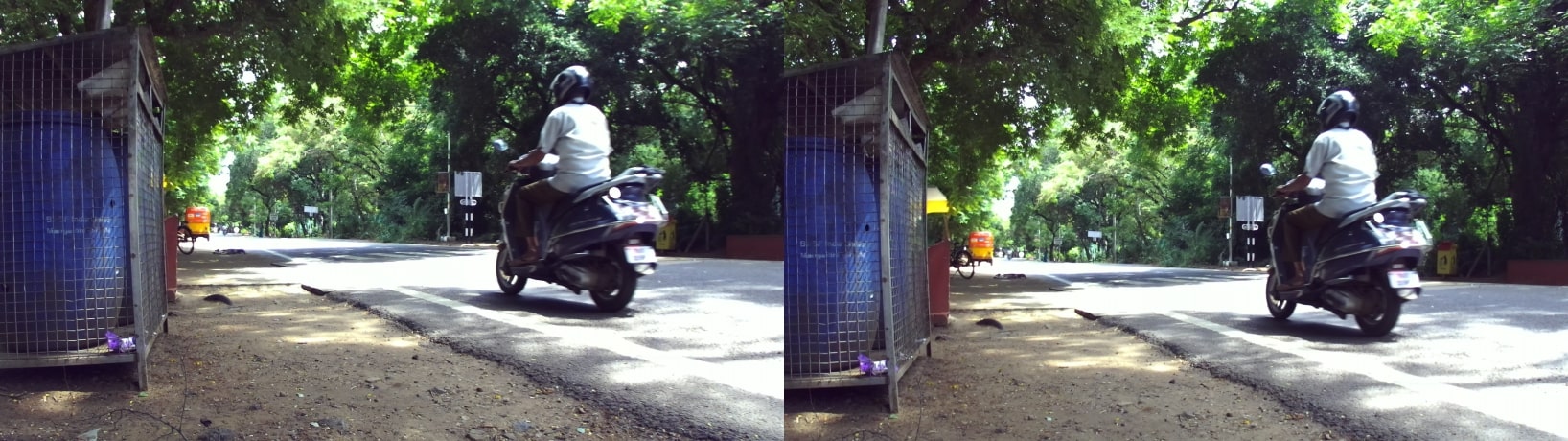}
    \end{minipage}
    &
  \vspace{0.05in} 
  Road through a forest; varied objects (motorcycles, buses and cars),moderate background motion (vegetation), no camera motion, high detail, complex moving depth structure, natural light.\vspace{0.05in}
    \\ \hline
    
    14 & Grass &
    \begin{minipage}{.4\textwidth}
      \includegraphics[width=\linewidth]{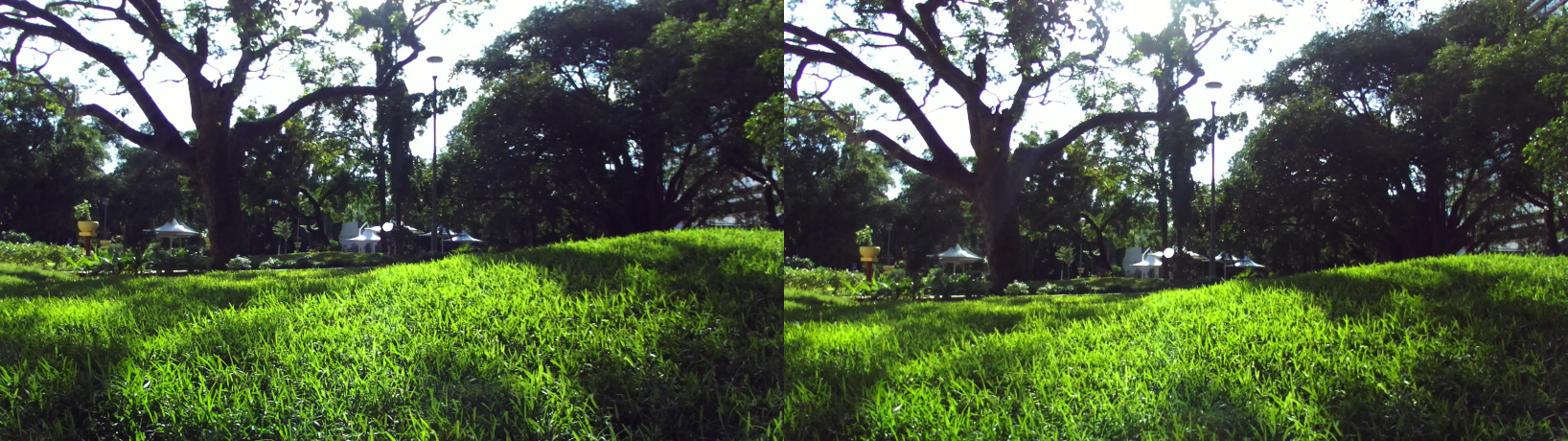}
    \end{minipage}
    &
    \vspace{0.05in}
    Trees and blades of grass moving in strong wind (complex motion), changing patterns of lights and shadows, no camera motion, complex depth structure, high detail, direct sunlight.\vspace{0.05in}
    \\ \hline
    
    15 & Classroom &
    \begin{minipage}{.4\textwidth}
      \includegraphics[width=\linewidth]{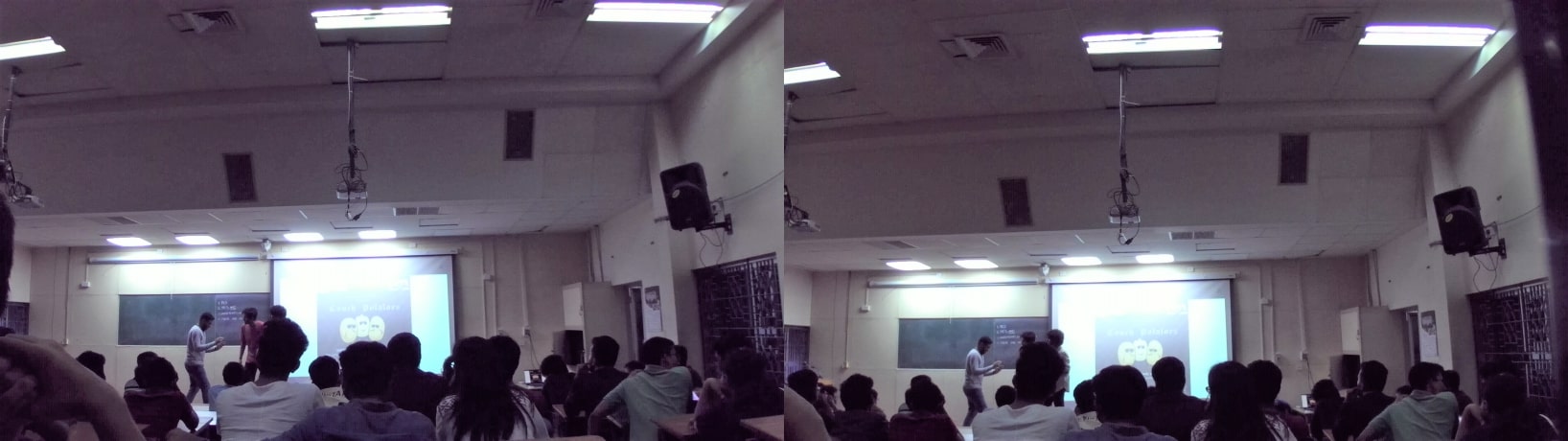}
    \end{minipage}
    &
    \vspace{0.05in}
    Classroom scene; group presentation, complex unpredictable motion of students, no camera motion, simple detail, moderate depth structure, interference due to artificial light source.\vspace{0.05in}
    \\ \hline
   
  16 & Forest dusk &
    \begin{minipage}{.4\textwidth}
      \includegraphics[width=\linewidth]{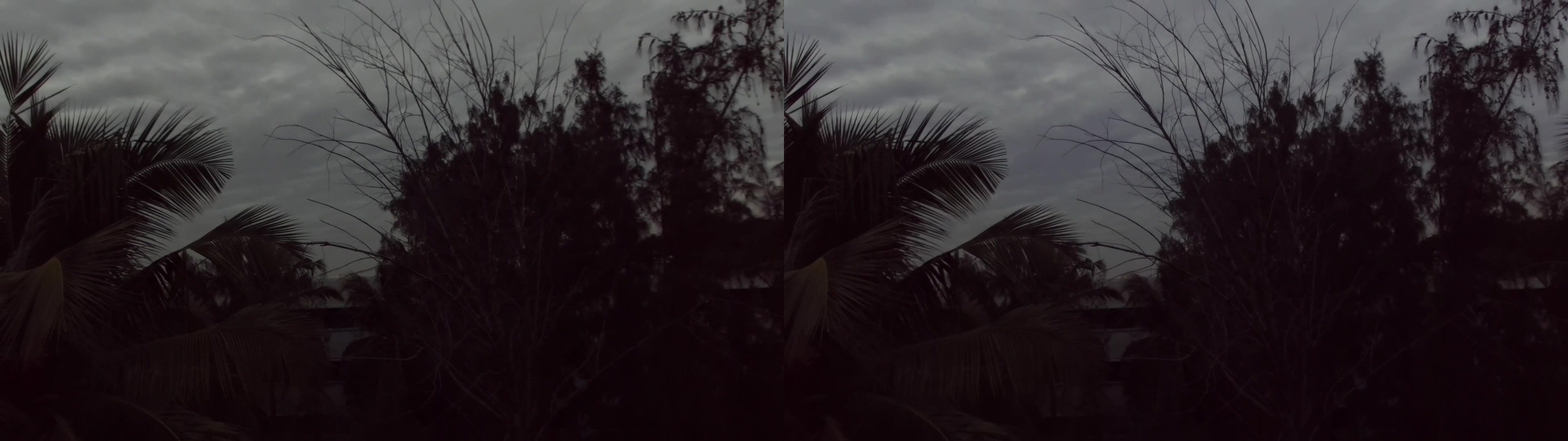}
    \end{minipage}
    &
    \vspace{0.05in}
    Plants in low natural light; hand held camera motion, low saturation, cloud patterns, vegetation occlusion patterns, moderate depth structure, silhouette of vegetation, static background.\vspace{0.05in}
    \\ \hline
    
    17 & House verandah &
    \begin{minipage}{.4\textwidth}
      \includegraphics[width=\linewidth]{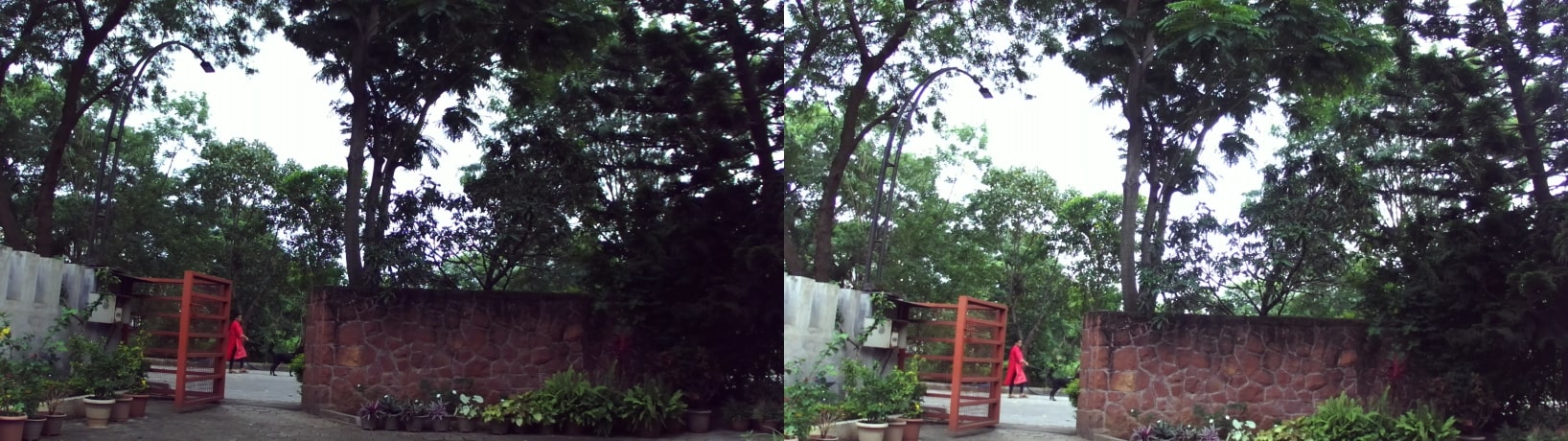}
    \end{minipage}
    &
    \vspace{0.05in}
    Roadside view from house; rigorous movement of vegetation, people moving on road, high details, complex depth structure, no camera motion, rich chromatic variation, bright sky.\vspace{0.05in}
    \\ \hline
    
    18 & NAC Building &
    \begin{minipage}{.4\textwidth}
      \includegraphics[width=\linewidth]{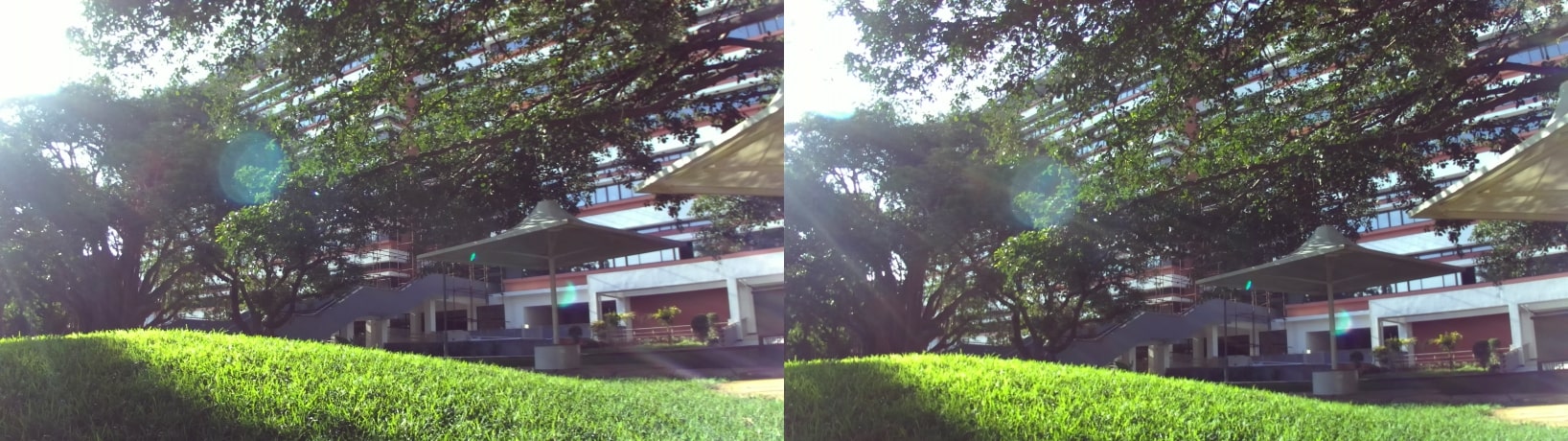}
    \end{minipage}
    &
    \vspace{0.05in}
    Building \& Garden; Lens flare, sunbeam, complex light interplay, reflection, no camera motion, high details, complex depth structure, rigorous movement of vegetation in the wind.\vspace{0.05in}
    \\ \hline

  \end{tabular}
  \label{Table_Video_B}
\end{table*}

\begin{table*}[t!]

\centering
  \caption{Left and right views of some selected scenes along with the obtained depth maps.}
  \begin{tabular}{|m{1cm} | m{4.1cm} m{4.1cm} m{4.1cm} | }
    \hline
    \vspace{1mm}
    \centerline{\textbf{Sr. no.}} &
    \vspace{1mm}
    \centerline{\textbf{Left view}} &
    \vspace{1mm}
    \centerline{\textbf{Right view}} &
    \vspace{1mm}
    \centerline{\textbf{Depth map}} 
    \\ \hline
    1 &
    \begin{minipage}{.23\textwidth}
    \vspace{0.05in}    
      \includegraphics[width=\linewidth]{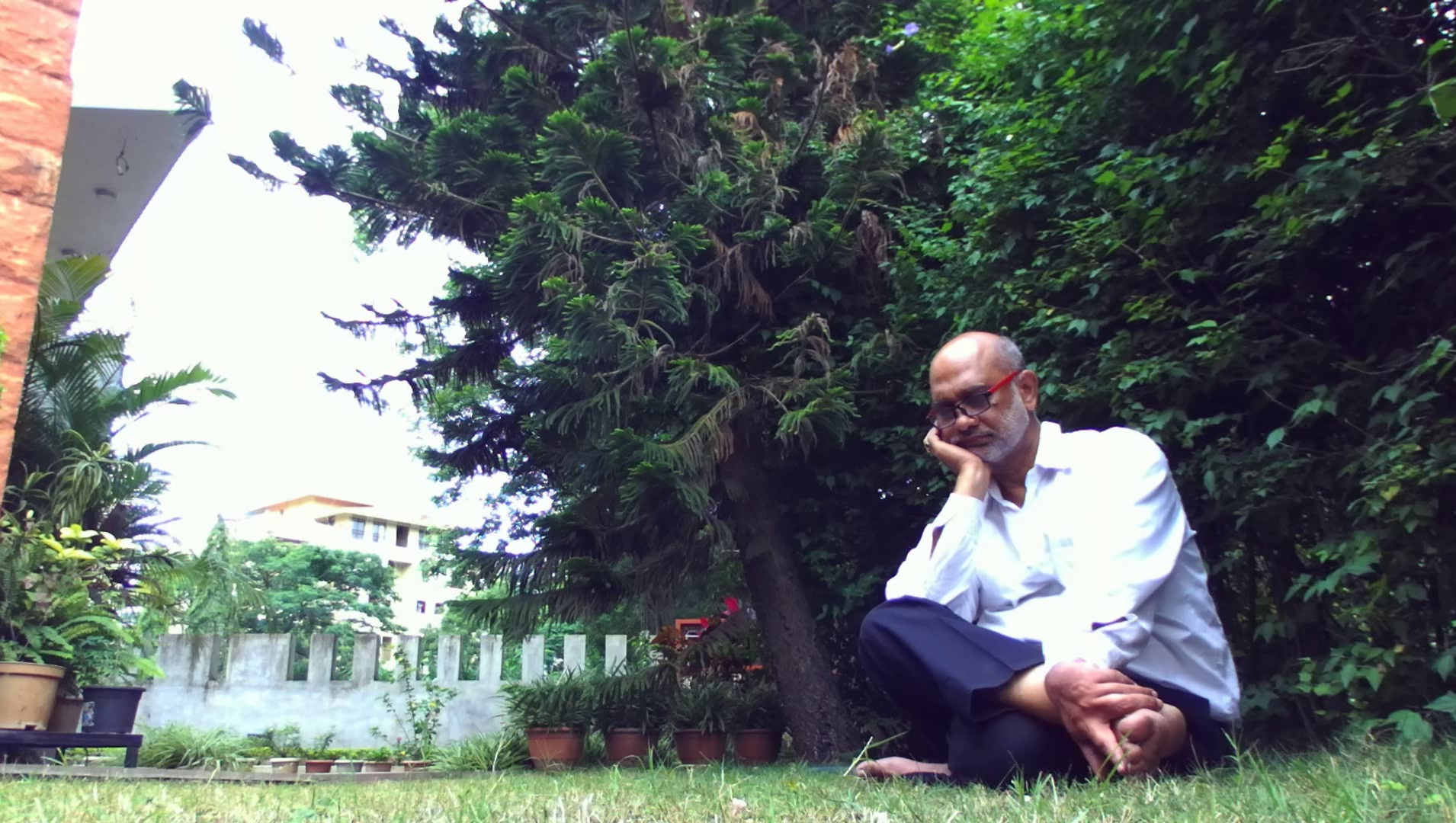}
      \vspace{0.02in}
    \end{minipage} &
    \begin{minipage}{.23\textwidth}
    \vspace{0.05in}   
      \includegraphics[width=\linewidth]{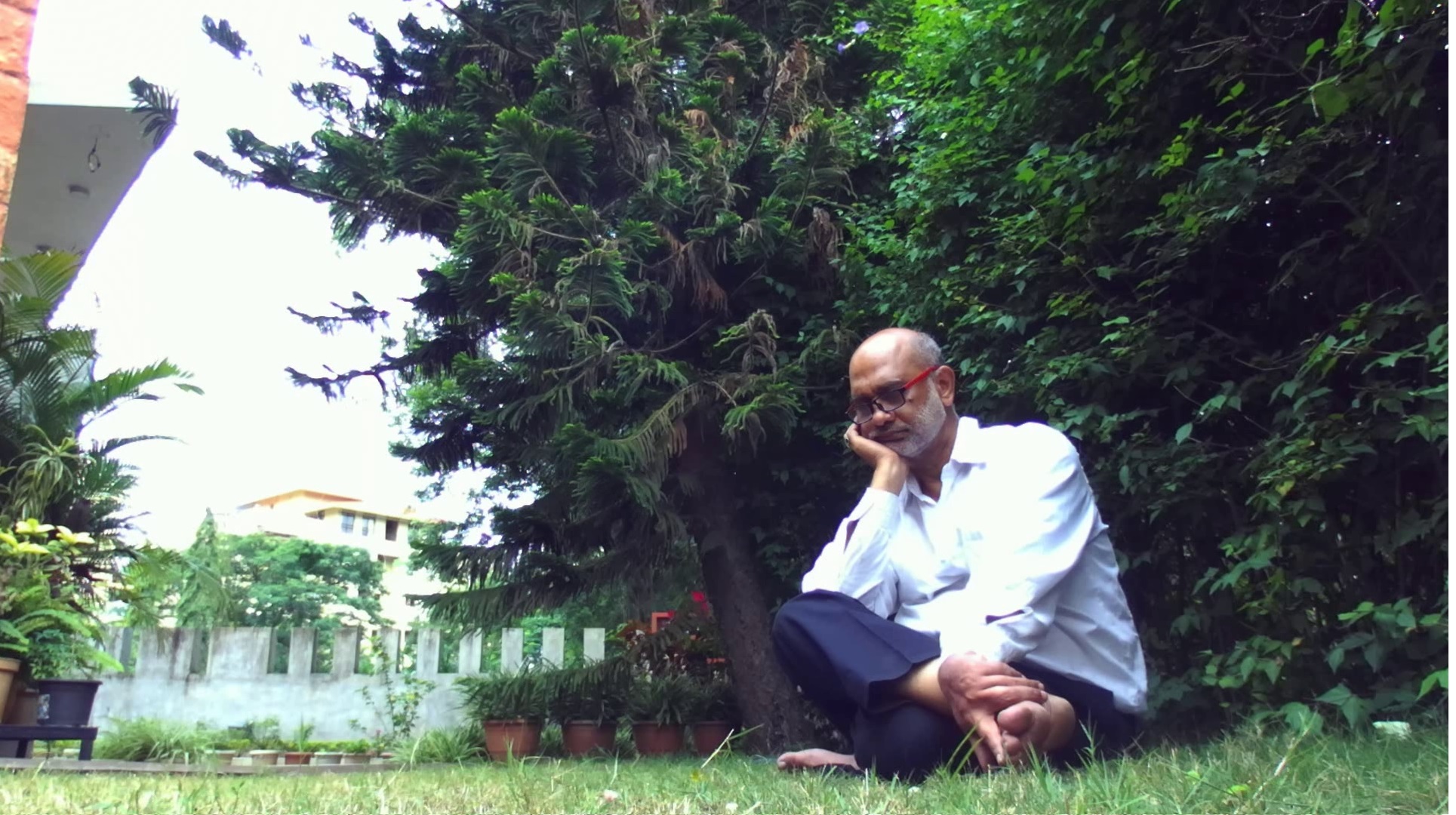}
      \vspace{0.02in}
    \end{minipage} & 
    \begin{minipage}{.23\textwidth}
    \vspace{0.05in}   
      \includegraphics[width=\linewidth]{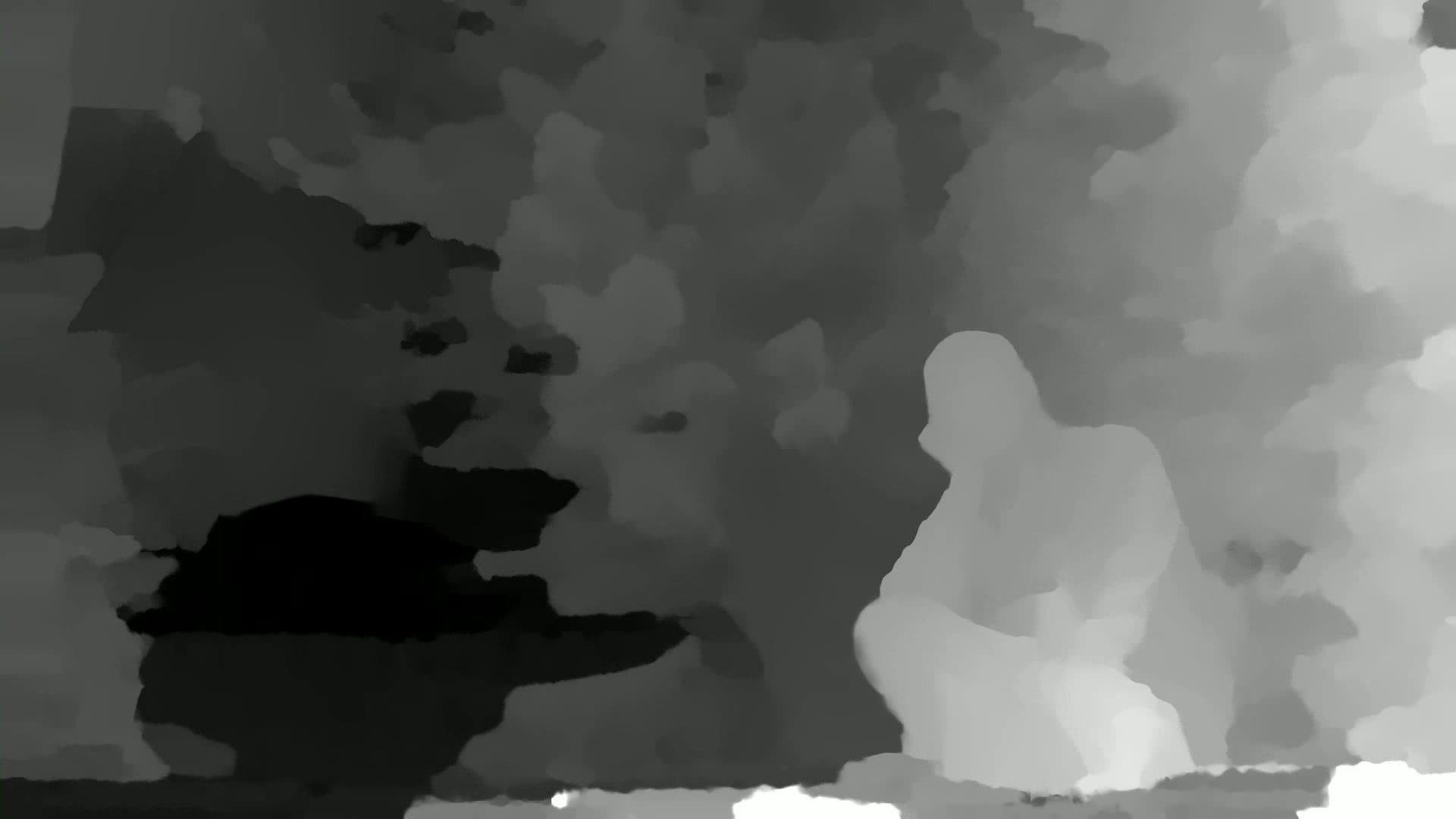}
      \vspace{0.02in}
    \end{minipage}\\ 
    
    2 &
    \begin{minipage}{.23\textwidth}
      \includegraphics[width=\linewidth]{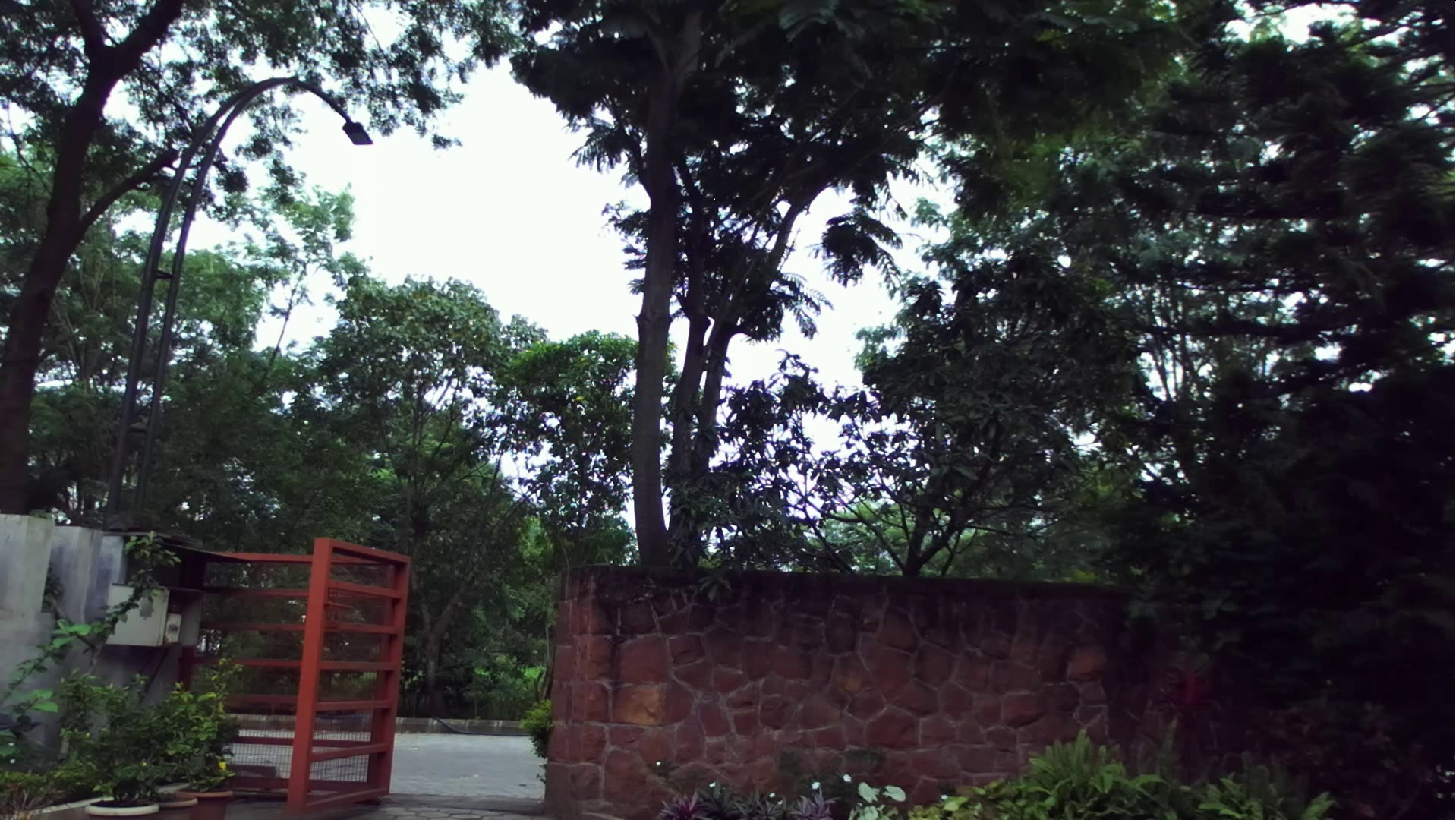}
      \vspace{0.02in}   
    \end{minipage} &
    \begin{minipage}{.23\textwidth}
      \includegraphics[width=\linewidth]{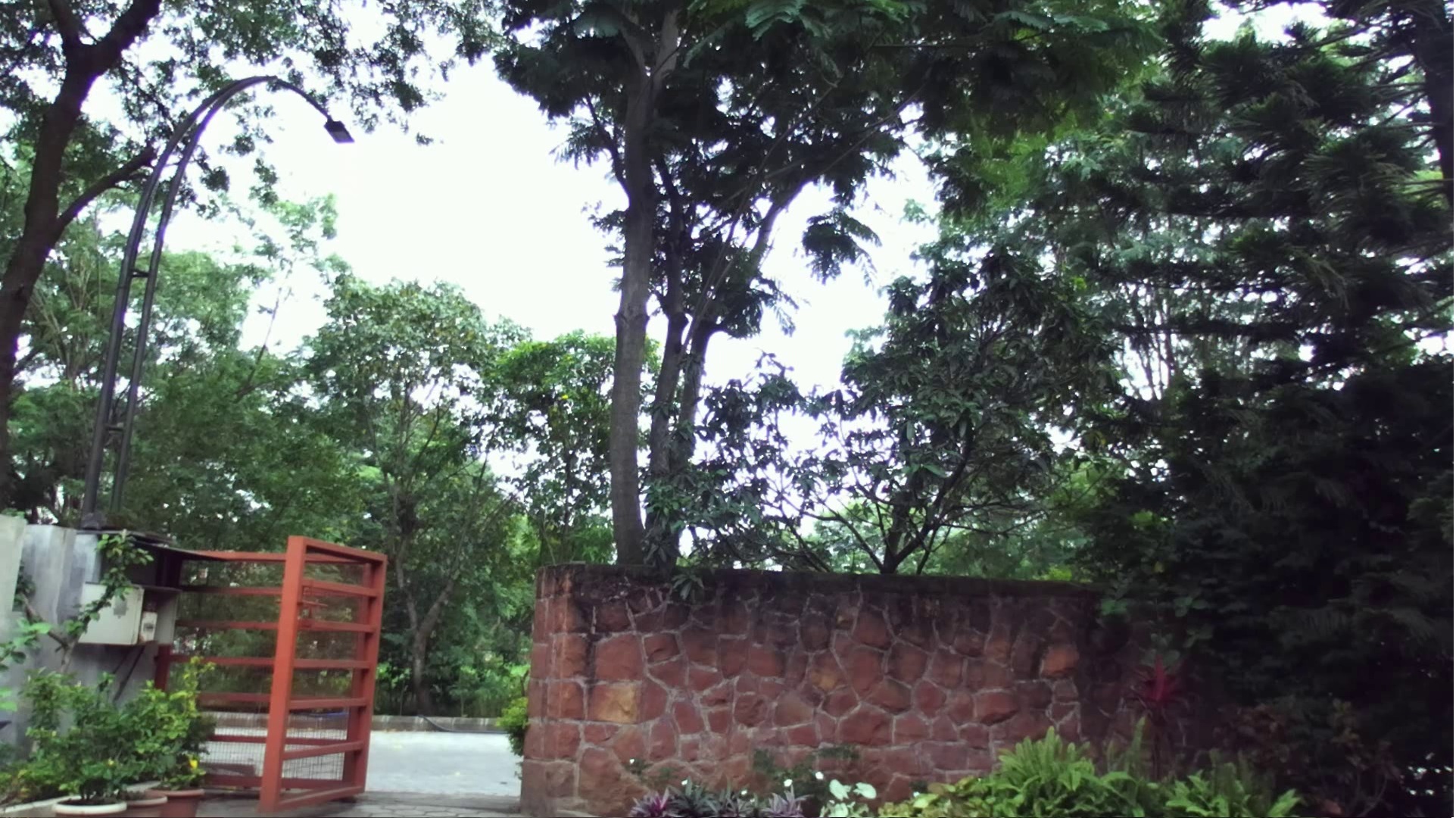}
      \vspace{0.02in}   
    \end{minipage} & 
    \begin{minipage}{.23\textwidth}
      \includegraphics[width=\linewidth]{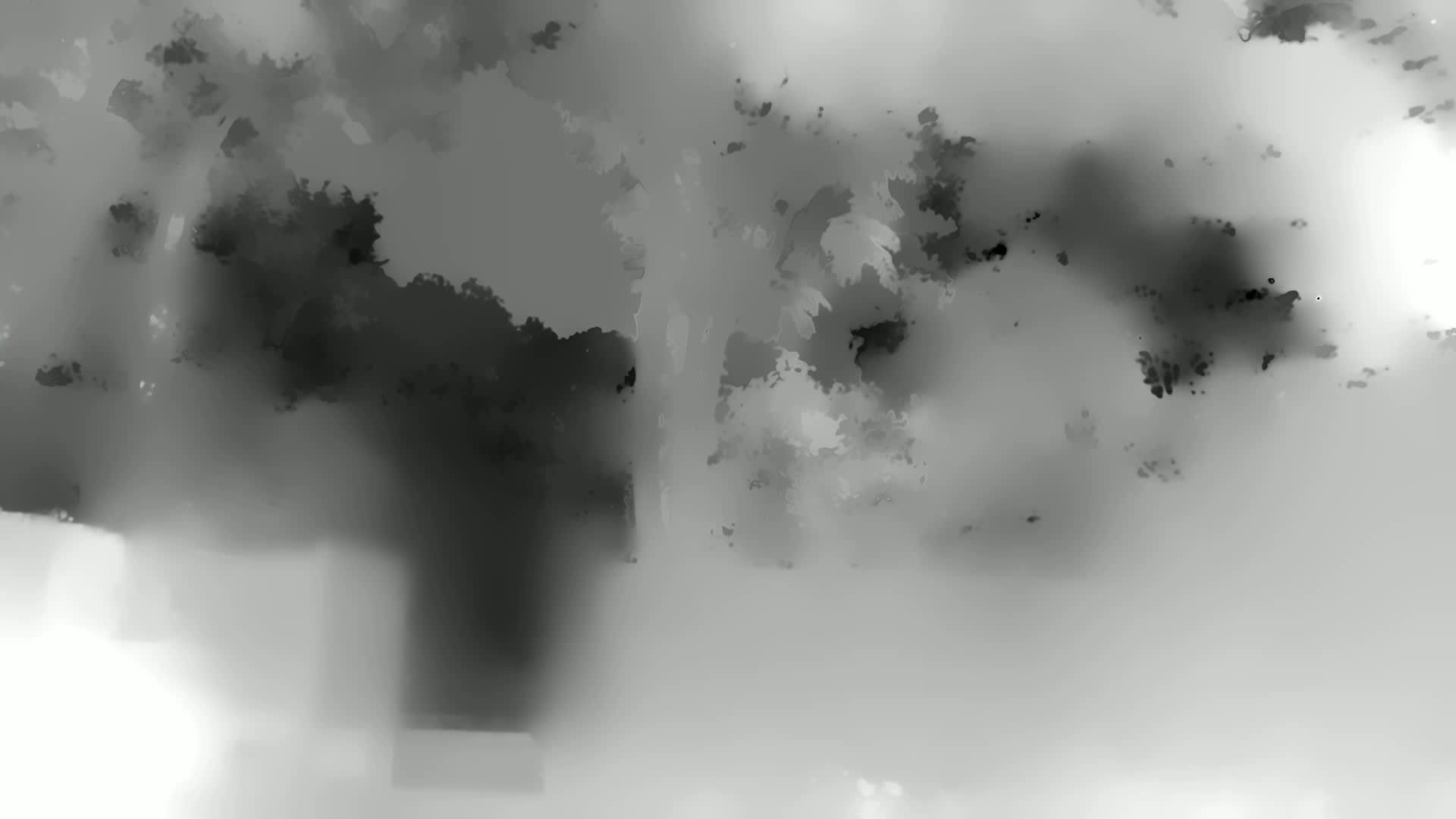}
      \vspace{0.02in}   
    \end{minipage}\\ 
    3 &
    \begin{minipage}{.23\textwidth}
      \includegraphics[width=\linewidth]{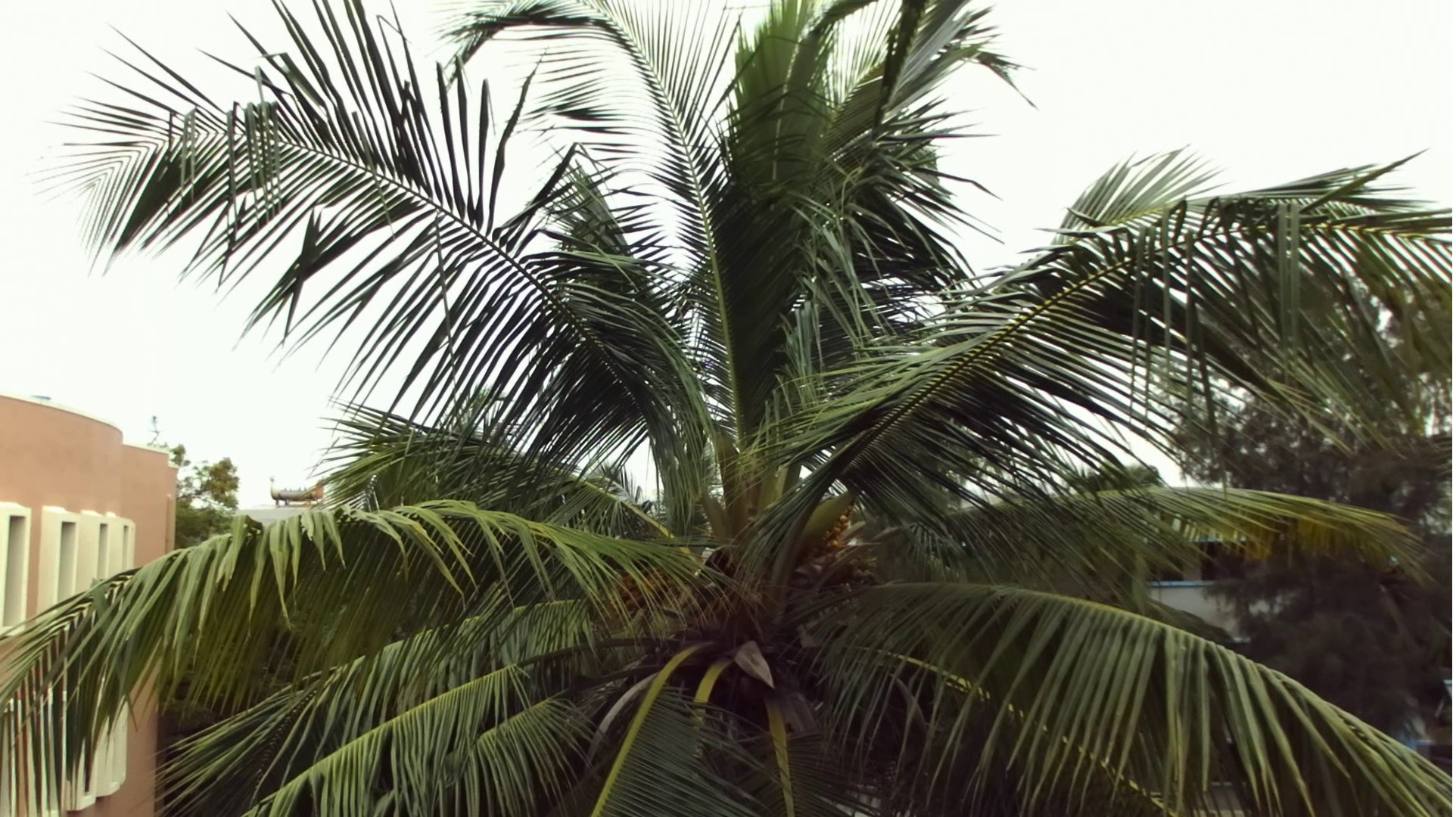}
      \vspace{0.02in}   
    \end{minipage} &
    \begin{minipage}{.23\textwidth}
      \includegraphics[width=\linewidth]{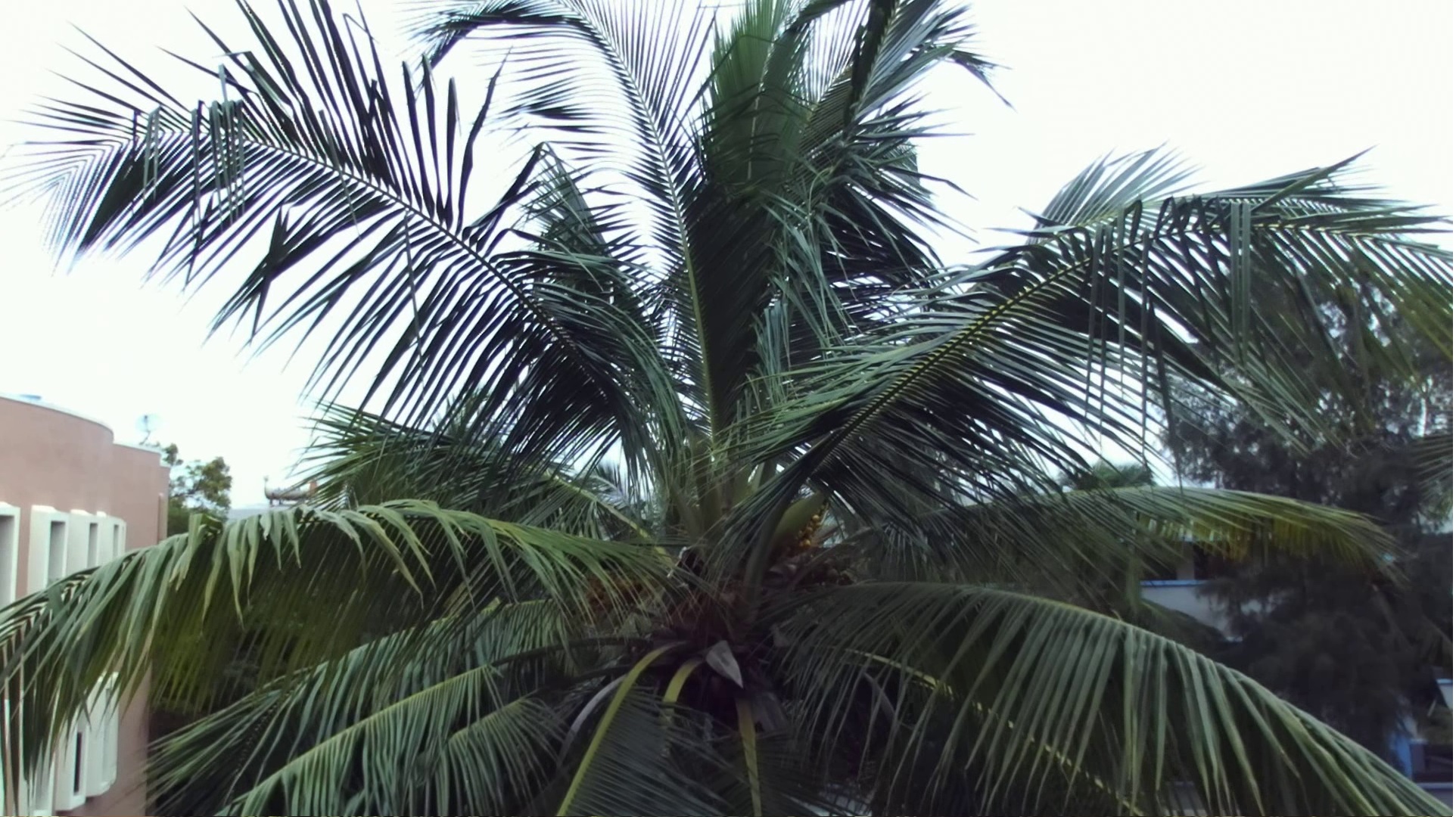}
      \vspace{0.02in}   
    \end{minipage} & 
    \begin{minipage}{.23\textwidth}
      \includegraphics[width=\linewidth]{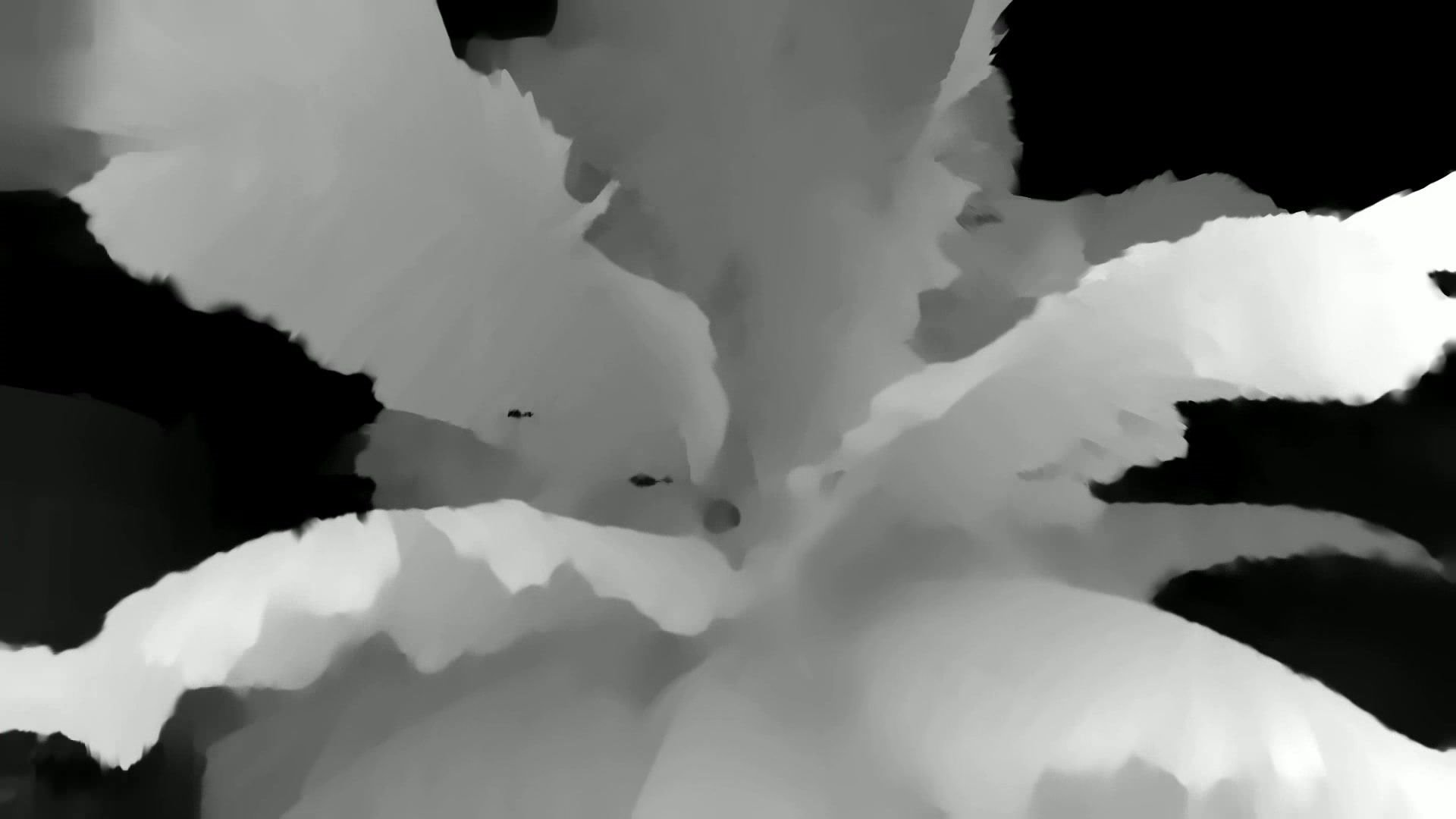}
      \vspace{0.02in}   
    \end{minipage}\\ 
    4 &
    \begin{minipage}{.23\textwidth}
      \includegraphics[width=\linewidth]{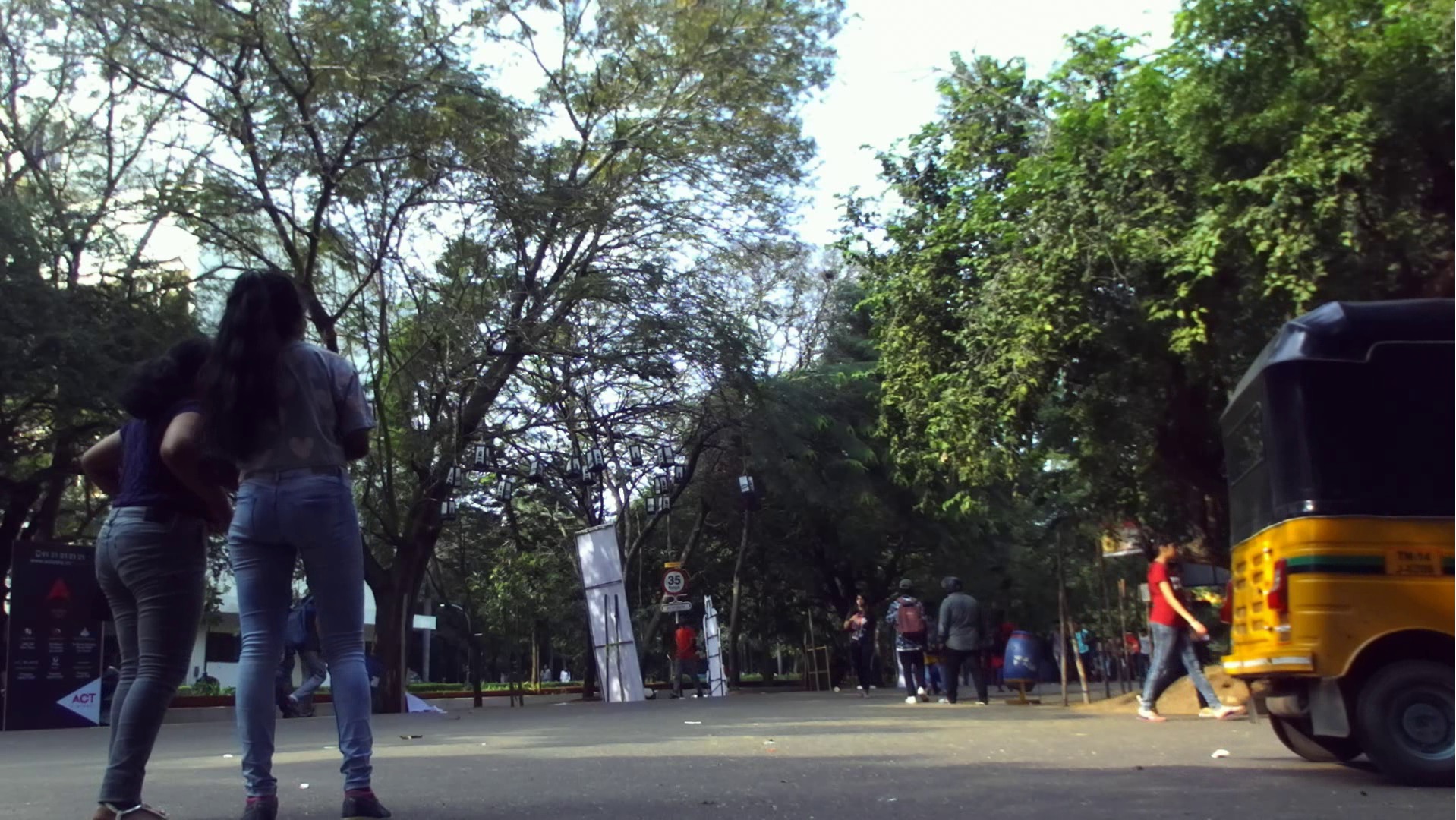}
      \vspace{0.02in}   
    \end{minipage} &
    \begin{minipage}{.23\textwidth}
      \includegraphics[width=\linewidth]{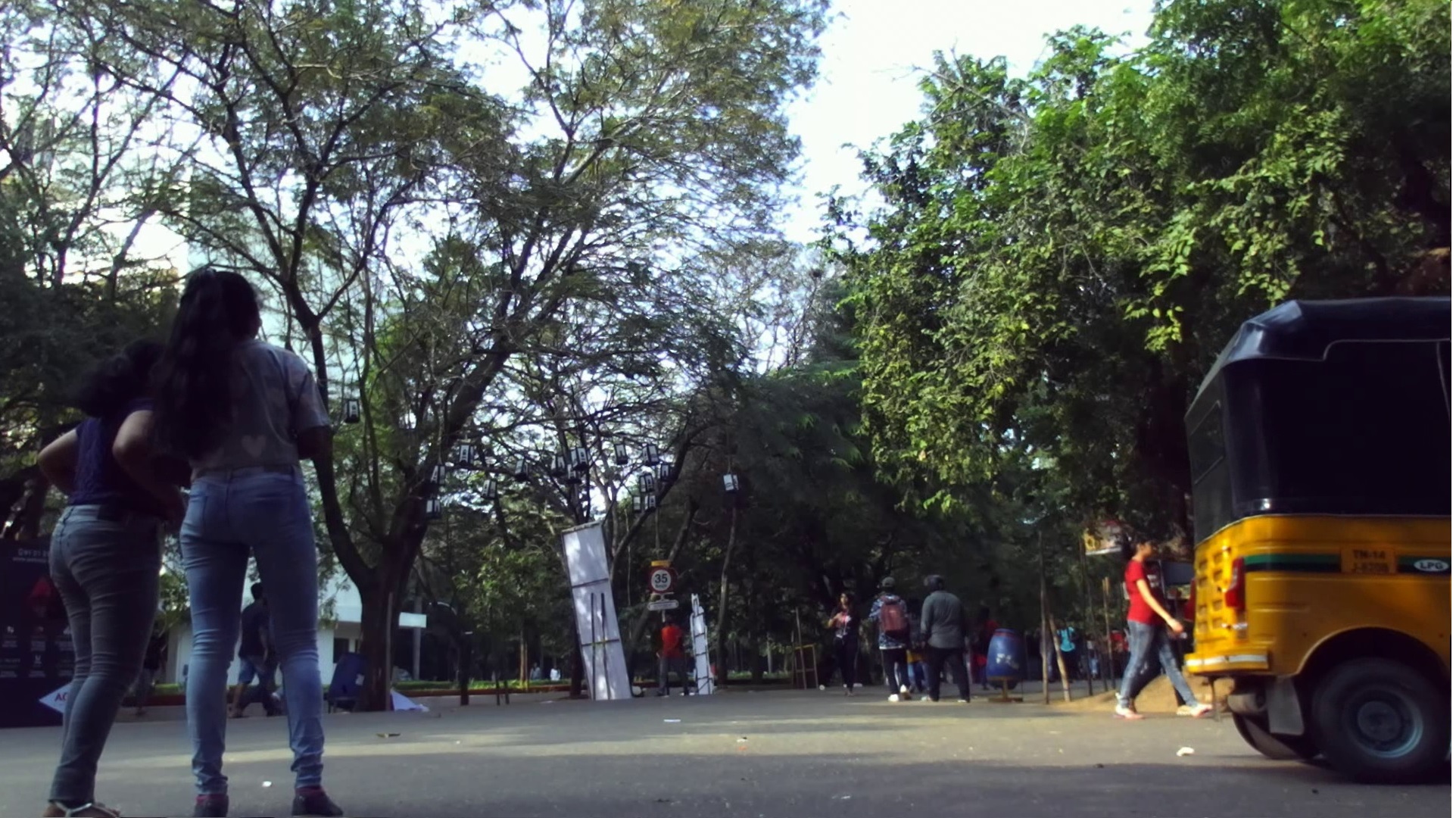}
      \vspace{0.02in}   
    \end{minipage} & 
    \begin{minipage}{.23\textwidth}
      \includegraphics[width=\linewidth]{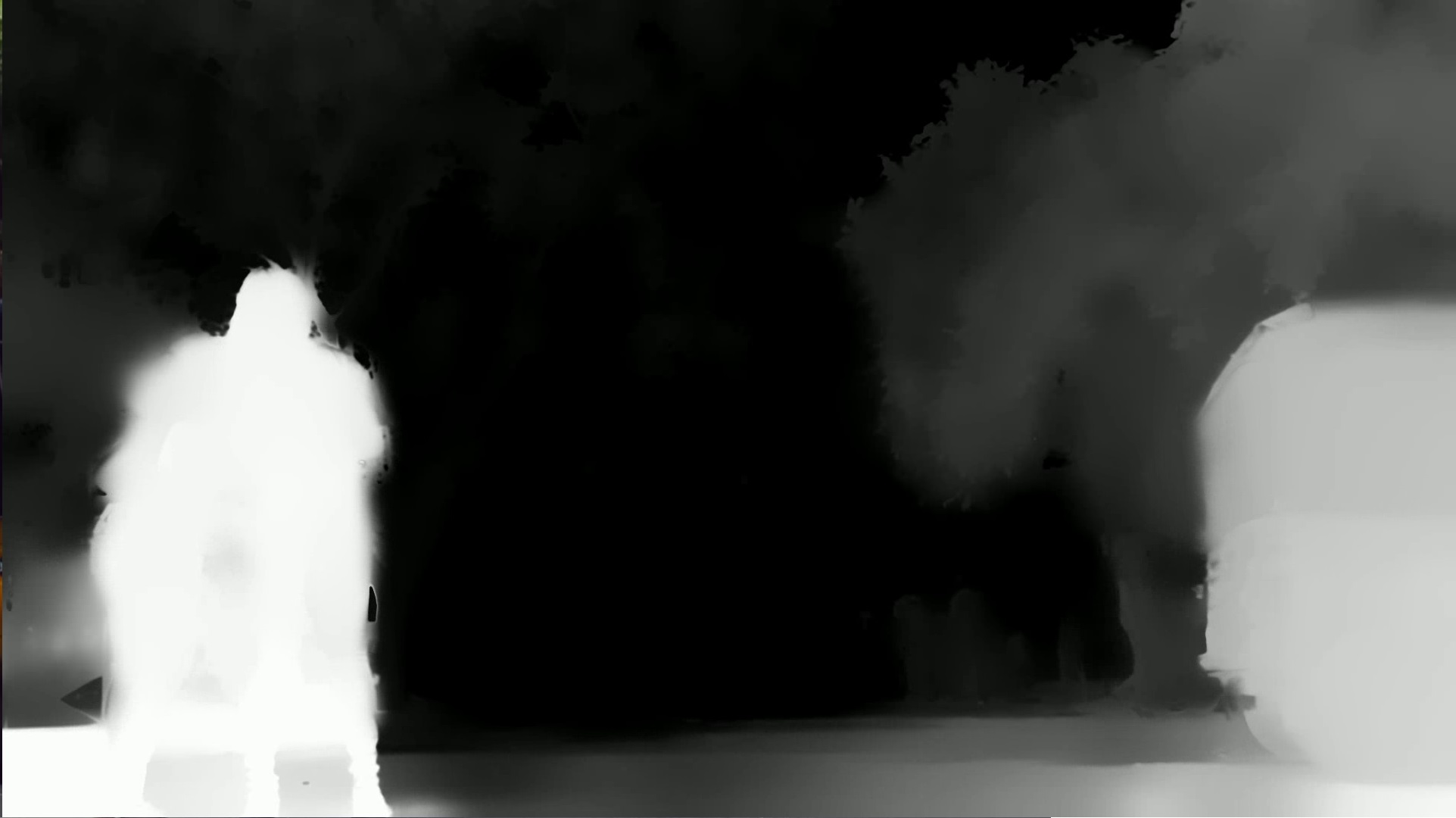}
      \vspace{0.02in}   
    \end{minipage}\\ 
    5 &
    \begin{minipage}{.23\textwidth}
      \includegraphics[width=\linewidth]{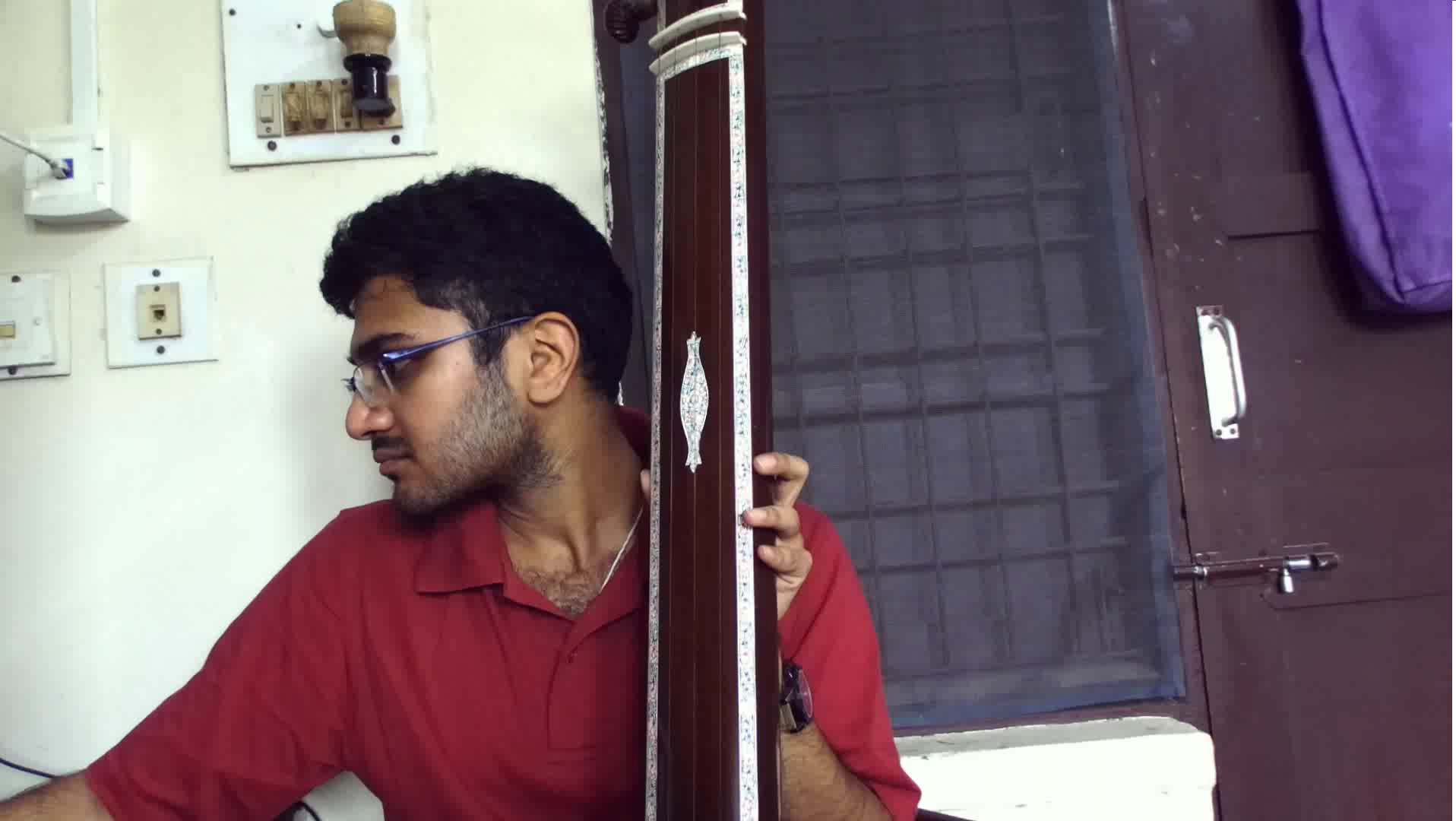}
      \vspace{0.02in}   
    \end{minipage} &
    \begin{minipage}{.23\textwidth}
      \includegraphics[width=\linewidth]{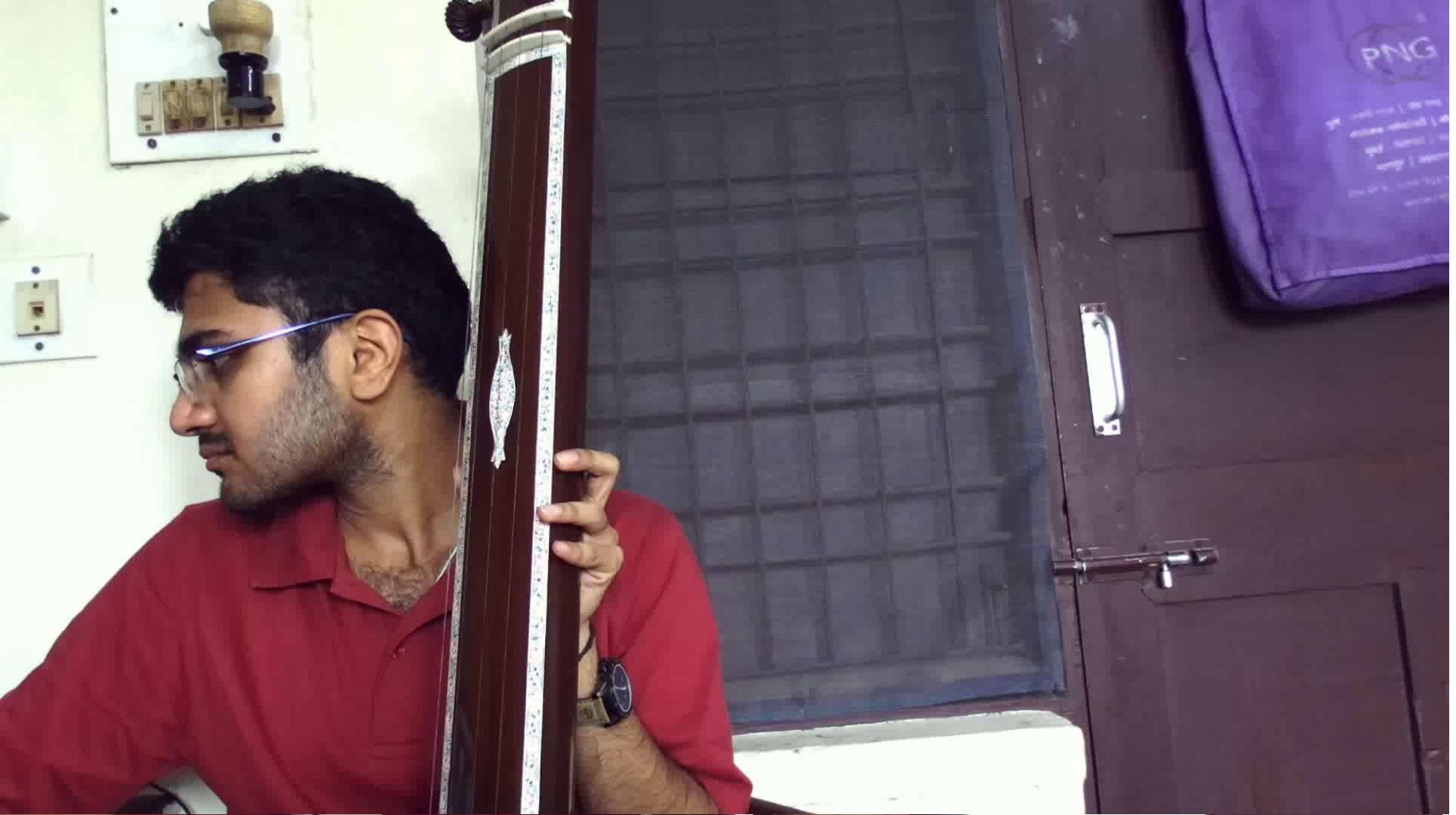}
      \vspace{0.02in}   
    \end{minipage} & 
    \begin{minipage}{.23\textwidth}
      \includegraphics[width=\linewidth]{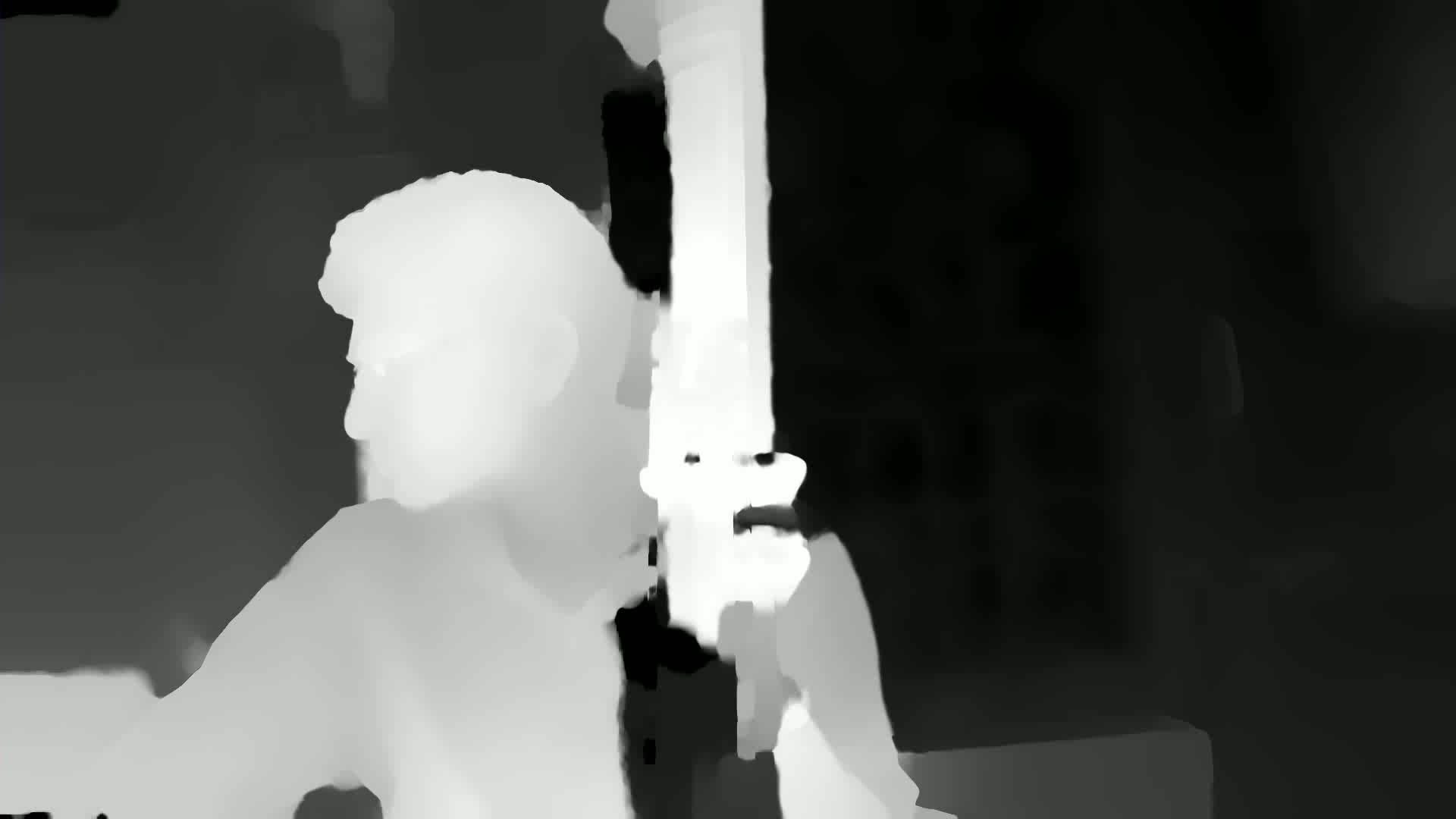}
      \vspace{0.02in}   
    \end{minipage}\\
    \hline
    \end{tabular}
    \label{tab:depth_map}
\end{table*}

$Classroom$ and $Singer$ are the two videos captured indoors. $Classroom$ shot with an artificial light source within the frame shows flickering introduced by the source. While sunlight is a steady light source, artificial light has a pulsating characteristic. However, the human eye cannot notice this flickering, a camera is capable of detecting these high pulse frequencies. Although $Singer$ does not have the interference of light in the video as the artificial light source is not directly in the frame, the intricate movements of the instrument playing hands possesses some crucial features to be analyzed.

With several other artifacts, lens flare can be observed in the $NAC Building$. Lens flare is a photographic effect that occurs when an intense light enters the camera lens, strikes the sensor, and is dispersed. Lens flare causes limited access to scene features of the flare-affected region, thus severely affecting depth estimation and HDR video reconstruction. Apart from the complexity introduced by flowing water in $Lotus$, the complexity also lies in the analysis of underwater objects, i.e., underwater flora. Scattering and attenuation of light as it propagates through the water leads to unclear appearance of underwater objects. 

Estimating the depth of moving objects from videos, where both the camera and the objects in the frame are moving, is a challenge that is open and underconstrained. $CoconutTree$ and $ForestDusk$ consist of a scenario where the handheld camera moves, and there are slightly swaying trees in the background. The camera motion is kept random to exploit all the features caused due to misalignment of the successive frames. Realistic constraints such as shadows as provided in $GC1$, $FerrisWheel$, $Rides$, $GC2$, $Cafe$, $Gurunath$, $Sarang$, $Avenue$; together with sunbeams as provided in $NACBuilding$, profoundly affects the present HDR Reconstruction and depth estimation methods. It's obvious that abrupt illumination or saturation shift disrupts the texture and colour consistency in the scene, resulting in shadows being identified as new raised surfaces or foregrounds. The most prominent features of the video dataset scenes are described in Table \ref{Table_Video_A} and \ref{Table_Video_B}. The left and right views of some selected scenes along with the obtained depth maps is shown in Table \ref{tab:depth_map}. 

To understand the scene's depth, texture, and colour pattern dynamics from the multi-exposure stereo recordings, new deep learning-based methods are required for such type of video dataset with intricate object motion. These algorithms must be able to detect object motion between multi-exposure shots and make the necessary corrections. Given the tremendous spatial complexity of the scenes, it's critical to have reliable algorithms that produce high-quality, exact 3D HDR videos. As a result, the research community faces a new challenge in developing intelligent algorithms to manage our dataset. 
\section{Progress, challenges, use cases and potential research opportunities in immersive technologies}
\label{sec:ChallengesImmersive}
In this section, we present the  advancements, application cases, issues, and possible research opportunities in the domain of HDR image/video reconstruction, depth estimation for 3D content generation, tone mapping and dynamic range
compression.

\subsection{Progress in HDR image/video reconstruction} 

In traditional HDR imaging, special HDR cameras are used to capture HDR images \cite{Nayana2015HighDR}\cite{Tursun2015DEghosting}. However, these cameras happen to be too expensive for most users. As a result, one typical method for HDR imaging is to use reconstruction techniques to create an HDR image from images acquired by a low-dynamic-range (LDR) camera.

\subsubsection{HDR Image Reconstruction using Multi-Exposure LDR}

One of the most prevalent approach for reconstructing HDR images is to fuse numerous LDR images \cite{Paul2008RecoveringHDRRM, Mann95onbeing, Granados2010LinearDCamera}. In this approach, one image in the stack of multi-exposed LDR images (generally the image with median exposure) is taken as the reference image. The remaining images are utilized to compensate for the lost features caused by over-/under-exposure of some local regions in the reference image. However, low-quality HDR images are obtained if the LDR images are not aligned properly, which is a practical scenario for most scenes arising because of foreground object motion. This leads to ghosting and blur artifacts. The foreground object's substantial movement causes misalignment between LDR images. Also, because of occlusion induced by moving objects, some content information is lost in the over-/under-exposed zones.

For a minor object movement, a few methods \cite{Jacobs2008AutoHDRGeneration, Grosch06fastand, Yan2017HDRSparse, Sri2012GhostDetection, Tursun2015DEghosting, Kalantari2017DHDRIofDynScenes} have been developed to detect apparent motion in multiple LDR images followed by removal of these regions in the fusion. It's clear that these methods lead to major loss of information for large object motions. To overcome these issues, several classic approaches use optical flow \cite{ Yan2019RobustArtFreeHDR, ZBW11} to align the LDR pictures in the pre-processing step before fusing them. Nonetheless, for optical flow based alignment, motion in over-saturated and under-saturated regions causes visible distortions in reconstruction results. Better alignment methods include feature alignment through correlation guidance \cite{s20185102,Pu_2020_ACCV}, image translation based feature alignment \cite{GreenRoshK2019DeepML}, static multi-exposure fusion based alignment \cite{Prabhakar2017DeepFuseAD, ma2019mefnet, Xu2020MEFGAN, xu2020aaai}, etc.

Lin et al. \cite{Lin2021DL4HDRI} showed that feature alignment through correlation guidance is flexible and effective. However, it is sensitive to over-saturated areas, which frequently results in the loss of textual details due to feature exclusion and needs high computational cost. From a practical standpoint,  real-time HDR imaging is preferred for real-life applications. Real-time HDR imaging research is lacking due to limited availability of real-time multi-exposure LDR data. 

\subsubsection{HDR Image Reconstruction using Single-Exposure LDR}

Reconstruction of an HDR image from a single exposure, unlike multi-exposure LDR images, does not have limitations due to misalignment. Conventional methods for HDR image reconstruction from single exposure LDR image employ inverse tone mapping operators (iTMOs) to increase the dynamic range of LDR images \cite{Rafael2014HQReverseTM, Huo2014ITMRetinalResponse, Francesco2006ITM}. Although such methods give an impression of enhanced dynamic range by improving highlights, upon inspection, it turns out that little information has been reconstructed in the saturated regions. Recent methods proposed to adopt CNNs for single image HDR reconstruction. HDR image generation for real complex scenes using single-exposure LDR is quite challenging. It has to perform tasks such as suppression of highlights in over-exposed areas (light source and reflection area), noise elimination in under-exposed areas area (strong noise area and under-brightness area) and several other task arising due to scene specific complexity. Our captured dataset seeks to provide the research community with complex natural images as training and validation data to enhance the single exposure based HDR depth estimation methods.

\subsubsection{Stereo-Based HDR Image Reconstruction}

Lin et al. \cite{Lin2009Stereoscopic} proposed a basic stereo based HDR reconstruction framework. The framework pipeline includes scale-invariant feature transform (SIFT) matching for finding the corresponding features in stereo pairs for modeling the camera response function (CRF). This is followed by obtaining disparity image from stereo matching. To deal with the stereo mismatch, the camera response curve is used to transform the image with the viewpoint for HDR image synthesis into an image with the same exposure as the other image from the stereo pair. Subsequently, the ghost removal method is used. With such basic framework the HDR image reconstruction quality depends upon the performance of each processing step.

Deep neural networks have recently been introduced in stereo-based HDR imaging. Chen et al. \cite{Chen2019SHDRIGAN} proposed a representative framework for stereo-based HDR imaging consisting of exposure view transfer and image fusion steps. The State-of-the-Art method \cite{Chen2020SHDRI} substituted the conventional pipeline \cite{Lin2009Stereoscopic} with DNNs. Though, introduction of DNNs increases the computational cost. Multiple modular DNN methods require large dataset for better training and fine-tuning for the reconstruction of HDR images in an end-to-end manner. Also, unavailability of large multi-exposure stereo dataset of real-world scenes hinder the scope of DNNs in real-world scenarios.

\subsubsection{HDR video reconstruction}

Deep learning methods learn HDR visual content from input LDR video or LDR frame sequence. Optical flow methods are commonly used to align consecutive frames, but when dealing with multiple exposures, the optical flow algorithm produces inferior HDR video. Upholding temporal consistency in consecutive frames of multi-exposure sequences is a relatively challenging task that requires more attention from the research community. 

Capturing an LDR sequence with single alternating exposures and reconstructing the missing content at each frame is a viable approach to make an HDR video. Kalantari et al. \cite{Kalantari2019DeepHV} proposed a two-step procedure where the initial step aligns the successive frames with the current frame using an optical flow network. Next, the final HDR frames are produced from the aligned images using a merge network. By contrast, Xu et al. \cite{Xu2019DeepVideoITM} performed gamma correction to convert the LDR resources back to real scenes, which were then used to create the HDR video. Moreover, they presented a novel approach for developing a deep learning-based video inverse tone mapping algorithm that aims to eliminate flickering issues induced by temporal inconsistencies.

Another approach for making HDR video is by using multiple alternating exposures sequences as input. Chen et al.\cite{chen2021hdr} introduced a coarse-to-fine framework, where initially coarse HDR video reconstruction is executed through optical flow alignment and fusion in the image space followed by refinement of the coarse predictions in the feature space. Jiang et al. \cite{Jiang2021HDRVR}, on the other hand developed an HDR reconstruction method that uses tri-exposure quad-bayer sensors. The lack of high-quality datasets restrict the growth of HDR video reconstruction. Despite the fact that certain synthetic datasets \cite{Jiang2021HDRVR, chen2021hdr} have been created for HDR video reconstruction domain, there remains a significant domain gap with real-world data.

\subsection{Consistent tone mapping challenges \& dynamic range compression} 

Tone mapping attempts to transform one set of colours into another, approximating HDR visual information on screens with restricted dynamic range. However, tone-mapping quality is subjective, and tone-mapping style choice varies from application to application depending upon user requirements. Broadly, traditional tone mapping operators (TMOs) are categorized as: Global tone mapping methods \cite{Ferwerda1996AMO, Larson1997TRO4HDR, Reinhard2005DynRangeRed, Tumblin1993ToneRep} apply a tone mapping curve to each and every pixel, whereas local tone mapping methods \cite{Durand2002FastBiFilteringHDRiD, Yang2018LocalTMAlgo, Rafal2006ContrastProcessing, Farbman2008EdgeMultiScTone, Liang2018HybridLayerTM} apply variations on each pixel while taking properties of the neighboring pixels into consideration.

Despite the fact that, traditional tone mapping methods produce good results, hyperparameter tuning is usually needed to get the best visual result. Deep learning-based approaches \cite{Patel2018GANTMHDR, Yang_2018_CVPR, Aakansha2020DeepTMO4HDRi} have recently been presented that do not require parameter tuning and significantly cut computation time by employing powerful GPUs. There is a need to formulate an adaptive tone mapping operator which can quickly alter itself to wide variability in real-world HDR scene in order to reproduce the best subjective quality output without any perceptual damage to it's content on a low resolution display and as well as a high-resolution display.

Tone-mapping is considerably more challenging in the case of 3D-HDR video because not only is temporal coherency required, but also inter-view coherency is essential. Also, it is essential to maintain tone difference between the two views lower than the minimum degree of tone difference arising due to binocular suppression.

\subsection{Depth Estimation Challenges} 

Stereo-based depth estimation methods use stereo matching algorithms to derive depth information from stereo image pairs. Most stereo matching algorithms are categorized as global and local methods, and aim to reduce the matching ambiguity factors arising from saturation region, texture region, and illumination variation. Existing monocular and stereo-based depth estimation methods are trained and tested using LDR or standard dynamic range(SDR) images/frame sequences. Present stereo matching methods provide inaccurate depth estimates in under- and over-exposed regions of LDR images. Also, depth estimation of a scene from an HDR image or multi-exposure image stack using current stereo matching approaches remains an ill-posed problem in specific scenarios. 

Certain features such as natural lighting, visibility transition arising due to viewpoint variation, scale variations, non-Lambertian reflections or partially transparent surfaces, illumination variations, the impact of low-textured regions, discontinuities in natural structures and high details, make consistent depth estimation from multi-exposure stereo views of the natural scene even more difficult. Also, stereo algorithms are more prone to subpixel calibration errors and their performance is dependent on the scene complexity. Further, depth estimation for multi-exposure stereo images for natural complex scenes are prone to inconsistencies arising due to temporal mismatch caused by motion associated to minute objects. As a result, there is still potential for more research into depth estimation from stereo HDR footage. 

Video-based depth estimation algorithms use successive frame sequences which are closely correlated and spatially near in 3D space. The aim is to predict depth such that the obtained depth has matching values along the temporal correspondences. Many video-based depth estimation algorithms have been proposed, with the central concept being exploiting the temporal information. According to various monocular sequence based depth prediction techniques \cite{Tananaev2018TemporallyCD, zhang2019temporal, Vaishakh2020RecurrentDEMonoVideo} and stereo sequence based depth estimation techniques \cite{Zhong2018OpenWorldSV, Zhan_2018_CVPR}, modelling the temporal associations or imposing optimization regulations among frames can enhance the depth predictions. Because of moving objects, camera posture and complex features of the scene, HDR video depth estimation has been a difficult challenge. As a result, the HDR video depth estimation approach is apt to erroneously textured regions, duplicated patterns and occlusions. 

\subsection{3D HDR encoding challenges}

As opposed to LDR and SDR images and videos, maintaining floating-point accuracy for processing HDR images and movies necessitates much greater storage and transmission costs. As a result, many HDR image/video compression algorithms that converts floating point images/frames to file formats suited for LDR video codecs have been developed. Following that, the data are encoded into compressed HDR video streams.  

Based on allowance for backward compatibility, HDR compression techniques are generally classified into two categories. The first category encodes HDR image or video without taking backward compatibility into account \cite{Mukherjee2019HDRVC, Rafal2004PerceptionHDRVEncoding}. The second category takes backward compatibility into account \cite{khan2010HDREncodingScheme, Rusanovskyy2016HDRcodingBackwardComp, Mantiuk2006HDRMPEGCompression}. HDR compression methods with backward compatibility typically comprises two layers: the base and enhancement layers. The base layer consists of tone-mapped LDR image/frame, further encoded using a conventional 8-bit codec. The second layer includes tone-mapping residual information that HDR application may utilize.

Multi-view images/frames are essential for creating 3D images/videos but are challenging to encode. Some existing coding methods for multi-view images/videos include Multi-View Coding (MVC)\cite{Chen2008MVC} and Multi-View High Efficient Video Coding(MV-HEVC) \cite{Hannuksela2015MVHEVC}. Chiang et al.\cite{Chiang2017HDRCodingMulti-view} effectively encode the multi-exposure multi-view images. This method transmit single LDR image per view, with the exposure of the multi-view images ordered in an interlaced method.After disparity map degeneration, HDR is constructed utilising the information from the neighbouring view at the decoder. This is followed by view wrapping. Due to the efficient compression of the multi-exposure multi-view LDR images, HDR images are reconstructed at the receiver side with less bandwidth use. Video compression standards for 3D and HDR videos are being developed by MPEG/ITU, however much less research has been done in encoding of 3D-HDR videos. Thus, standards for 3D-HDR video have yet to be developed.

\section{Conclusion} 
\label{sec:Conc}

We present a new diversified dataset consisting of multi−exposure stereo images/videos that provide a natural glance of the IIT Madras campus. The dataset stands out because of the complexity of diverse scenes it incorporates. Attributes of the dataset include Illumination variation, complex object movements, intricate texture variation of the background due to wind, Lambertian and specular reflection, cloud patterns, complicated depth structure, rich color dynamics, etc. The proposed dataset aims to provide various research possibilities to create a backward-compatible end-to-end content production methodology for 3D-HDR video. Additionally, challenges like depth estimation, encoding/decoding, tone mapping, dynamic range compression, visual attention modelling, quality assessment, and optimal stereoscopic/ autostereoscopic display presentation can be investigated with such a complex dataset. Our goal is to analyze the dataset particularly to show the potential for novel 3D-HDR technologies such as multimedia-centric Mobile 3DTV apps. In the future, we will concentrate on generating HDR stereo content with better rendering techniques. Further, the focus will be on developing a flexible yet scalable 3D-HDR error-resilient video encoding with error concealment awareness capabilities adjusted for reliable transmission of 3D-HDR content over the existing broadcasting pipeline.

\section*{Acknowledgement}
The scientific efforts leading to the results reported in this paper have been carried out under the supervision of Dr. Mansi Sharma, INSPIRE Hosted Faculty, IIT Madras. This work has been supported, in part, by the Department of Science and Technology, Government of India project “Tools and Processes for Multi-view 3D Display Technologies”, DST/INSPIRE/04/2017/001853.

\bibliographystyle{plain}
\bibliography{root}

\end{document}